\documentclass[Afour,sageh,times]{sagej}
\usepackage{moreverb,url}
\usepackage[colorlinks,bookmarksopen,bookmarksnumbered,citecolor=blue,linkcolor=blue,urlcolor=blue]{hyperref}

\usepackage{amsmath,amsfonts}
\usepackage{array}
\usepackage[caption=false,font=normalsize,labelfont=sf,textfont=sf]{subfig}
\usepackage{textcomp}
\usepackage{stfloats}
\usepackage{url}
\usepackage{verbatim}
\usepackage{graphicx}
\usepackage[latin1]{inputenc}
\usepackage{colortbl}
\usepackage{soul}
\usepackage{multirow}
\usepackage{pifont}
\usepackage{color}
\usepackage{alltt}
\usepackage{enumerate}
\usepackage{siunitx}
\usepackage{breakurl}
\usepackage{epstopdf}
\usepackage{pbox}
\usepackage{mathtools}
\usepackage{empheq}
\usepackage{amsmath}
\usepackage{bm}
\usepackage{tabularx}
\usepackage{amssymb}
\usepackage{mathrsfs}
\usepackage{algorithm}
\usepackage{algpseudocode}
\usepackage{booktabs}
\usepackage{titlesec}
\usepackage{makecell}
\titleformat{\subsubsection}{\normalfont\itshape}{\thesubsubsection}{1em}{}
\titlespacing*{\subsubsection}{0pt}{1ex plus 1ex minus .1ex}{1ex plus .1ex}

\newcommand{\vectT}{\textbf{\(\bm{T}\)}}

\newcommand{\vecte}{\textit{\(\bm{e}\)}}

\newcommand{\vectAA}{\textbf{\(\bm{A}\)}}
\newcommand{\vectXX}{\textbf{\(\bm{X}\)}}
\newcommand{\vectYY}{\textbf{\(\bm{Y}\)}}
\newcommand{\vectBB}{\textbf{\(\bm{B}\)}}

\newcommand\BibTeX{{\rmfamily B\kern-.05em \textsc{i\kern-.025em b}\kern-.08em
		T\kern-.1667em\lower.7ex\hbox{E}\kern-.125emX}}

\def\volumeyear{2025}

\begin{document}
\runninghead{Chen et al.}

\title{Optimal Uncertainty-Aware Calibration for the AX=YB Problem}

\author{Yanjia Chen, Xiangfei Li, Huan Zhao, Yiyuan Hong, Guanxiao Xia, \\Jiexin Zhang and Han Ding}

\affiliation{All authors are with the State Key Laboratory of Intelligent Manufacturing Equipment and Technology and the School of Mechanical Science and Engineering, Huazhong University of Science and Technology, Wuhan, 430074, China.}

\corrauth{Huan Zhao, State Key Laboratory of Intelligent Manufacturing Equipment and Technology and School of Mechanical Science and Engineering, Huazhong University of Science and Technology, No. 1037 Luoyu Road, Hong shan, Wuhan, Hubei, 430074, China.}

\email{huanzhao@hust.edu.cn.}

%%%%%%%%%%%%%%%%%%%%%%%%%%%%%%%%%%%%%%%%%%%%%%%%%%%%%%%%%%%%%%%%%%%%%%%%%%%%%%%%%%%%%%%%%%%%%%%%%%%%%%%%%%%%%%%%%%%%%%%%%%%
\begin{abstract}

\textit{\textrm{This article proposes a general optimization framework for solving hand-eye calibration problem. Unlike traditional methods, an iterative algorithm based on Lie algebra that achieves approximately global optimal solutions is developed. During the optimization process, the method strictly preserves the structural constraints of the calibration parameters and enables synchronized updates between calibration parameters. Recognizing that data used in real-word hand-eye calibration often contain uncertainty, especially in over-loading and large workspace industrial robot scenarios, which can significantly degrade accuracy, and accurately modeling such uncertainty is inherently difficult, this article avoids explicit uncertainty modeling. Instead, an uncertainty metric to evaluate the relative uncertainty between data sources is introduced and used to dynamically refine the iterative process. To further enhance convergence efficiency, an effective initial solution generation method that improves overall stability and accuracy is designed. Numerical simulations and real-world experiments validate the effectiveness of the proposed approach, and in synthetic datasets, the proposed approach improves the estimation accuracy by at least 67\% under high-uncertainty conditions compared with the existing methods.}} 

\end{abstract}

% Note that keywords are not normally used for peer-review papers.

\maketitle

\section{1. Introduction}
\label{sec:1}

With the increasing deployment of industrial robots (IRs), their roles are evolving beyond traditional manufacturing tasks like welding \cite{w1}, polishing \cite{w2}, and painting \cite{w3} to more advanced applications such as grinding aircraft engine blades \cite{w4}, assembling electronic devices \cite{w5}, and constructing aerospace equipment \cite{w6}. As the standards for task performance continue to elevate, the requirements for the precision of robotic actions are becoming more stringent.

Achieving high precision operations is one of the key research directions in robotics. However, due to the structural configuration of IRs, accomplishing such precision tasks requires leveraging sensor data to reduce the uncertainty in both position and orientation. Currently, widely adopted approaches include robot calibration \cite{w7} and endowing robots with pose perception capabilities \cite{w8}, i.e utilizing vision-based guidance to assist robotic operations. Robot calibration typically requires unifying the robot's coordinate system with that of an external vision measurement system
, similarly vision guided methods also require aligning the visual sensor's coordinate system with the robot's coordinates. 

Depending on the installation of the vision sensor, two common configurations are used. In the first scenario, the vision measurement system is mounted directly on the robot \cite{w9}. In the second, it is fixed externally in the robot's environment \cite{w10}. These two configurations exhibit strong analogies in their geometric and computational models. The present study is conducted under the second scenario but is readily extendable to the first.

To achieve high precision performance, a critical task is to accurately transform the information acquired by the vision measurement system into the robot's coordinate frame. This process is commonly referred to as hand-eye calibration (HEC), formulated as $\boldsymbol{AX=XB}$, or hand-eye and robot-world calibration, formulated as $\boldsymbol{AX=YB}$. It is worth noting, that the $\boldsymbol{AX=YB}$ problem can be transformed into an $\boldsymbol{AX=XB}$ problem, allowing $\boldsymbol{X}$ to be estimated first, followed by $\boldsymbol{Y}$. Therefore, the $\boldsymbol{AX=YB}$ formulation is more general compared to the $\boldsymbol{AX=XB}$ problem. In this paper, the final solution framework is designed for the $\boldsymbol{AX=YB}$ problem but is also compatible with the $\boldsymbol{AX=XB}$ problem. For simplicity, the two problems are collectively referred to as the HEC problem.

The core objective of HEC problem is to estimate the hand-eye calibration parameters (HECPs) $\boldsymbol{X}$ and $\boldsymbol{Y}$, representing the poses of the camera with respect to the robot end-effector, and the robot end-effector with respect to the calibration target, respectively, and it is typically represented using homogeneous transformation matrices (HTMs). A straightforward numerical solution would ideally fulfill this task. However, due to the influence of uncertainty factors, such an approach based on a numerical solution often propagates errors from the calibration parameters into the overall system, which is contrary to the objective of high-precision operations. 

These uncertainty factors refer to uncontrollable errors, including human operational errors, robotic motion errors, and measurement inaccuracies. Among them, human-induced errors may result in misalignment or incorrect correspondence of sensor data \cite{w11}. The causes of robotic motion errors are inherently complex, involving factors such as installation clearance, non-linear deformations, mechanical backlash, and variations in dynamic parameters \cite{w12}. Similarly, the sources of measurement errors are equally intricate, encompassing the principles of measurement, inherent sensor inaccuracies, and environmental influences \cite{w13}. In summary, it may be impossible to achieve a perfect explicit model of the above-mentioned uncertainties.

At present, many studies on the estimation of calibration parameters primarily consider the impact of measurement noise on solution accuracy, and almost universally model this noise as Gaussian \cite{w14} or Gaussian-based distributions \cite{w15}, as such it is essentially a form of random noise. This type of uncertainty is referred to in this paper as Aleatoric Uncertainty (AU). In contrast, only a limited number of studies take into account the influence of robotic precision such as \cite{w16}, which is attributed to system level or structural uncertainties and this form of uncertainty is referred to herein as Epistemic Uncertainty (EU), this uncertainty, if not considered, deteriorates the result of the hand-eye calibration when the robot's motion accuracy is poor.

It is unfortunate that current approaches to solution analysis still predominantly model the uncertainty in HEC problems as independent Gaussian distributions. From a certain perspective, this is an oversimplification and may be inappropriate, particularly in the case of the robot having poor motion accuracy. Among the two types of uncertainties that affect the estimation accuracy of HECPs, unlike AU, the sources of EU exhibit dependent rather than independently stochastic behavior. In general, it can be systematically mapped from the pose of IRs, as a result, the uncertainties for HEC in Cartesian space are not necessarily Gaussian distributed, and in many cases, they exhibit statistical dependence. 

It is important to clarify that the term robot motion accuracy mentioned here specifically refers to the pose accuracy of the robot (RPA), which must be distinguished from the pose repeatability of the robot (RPR). RPA measures the robot's ability to reach a specified pose in space, while RPR quantifies its ability to return consistently to the same pose. In non-teaching-based operational scenarios, task accuracy relies more heavily on RPA. Unfortunately, RPA is often significantly lower than RPR, particularly in large workspaces or when handling heavy payloads at the end effector. In such cases, the resulting errors can reach the millimeter scale \cite{w17}, posing a serious challenge for high-precision tasks. To improve RPA, numerous studies have focused on precision calibration of IRs which can enhance RPA by an order of magnitude \cite{w18}. However, such calibration efforts are highly dependent on accurate robot kinematic modeling and the use of external measurement systems \cite{w19}. Moreover, most of these methods such as in \cite{w20} and \cite{w21} also rely heavily on the estimation of HECPs, which further underscores the interconnected nature of a high level of accuracy in estimating HECPs and robot accuracy improvement.

As previously noted, virtually all conventional methods neglect the influence of EU, which also significantly impacts the estimation of HECPs. Meanwhile, modern applications for IRs frequently involve scenarios with heavy payloads or large workspaces, where systems inherently exhibit substantial EU. During HEC process, the input data sources \{\({\boldsymbol{A_{i}}}\)\} and \{\({\boldsymbol{B_{i}}}\)\} are subject to compounded uncertainties that conventional methods fail to account for, ultimately leading to suboptimal calibration accuracy. In theory, constructing a perfect uncertainty model for robot calibration systems is extremely challenging. However, a crucial and insightful point is that regardless of the complexity of the uncertainty model for HECPs estimation fundamentally depends on the data sources \{\({\boldsymbol{A_{i}}}\)\} and \{\({\boldsymbol{B_{i}}}\)\}. For the $\boldsymbol{AX=YB}$ problem, assuming \{\({\boldsymbol{A_{i}}}\)\} is derived from the robot's teach pendant and \{\({\boldsymbol{B_{i}}}\)\} from an external sensor, it follows that the effects of both EU and AU ultimately manifest only through distortions in \{\({\boldsymbol{A_{i}}}\)\} and \{\({\boldsymbol{B_{i}}}\)\}. Theoretically, if one could obtain ideal \{\({\boldsymbol{A_{i}}}\)\} and \{\({\boldsymbol{B_{i}}}\)\}, or even non-ideal \{\({\boldsymbol{A_{i}}}\)\} and \{\({\boldsymbol{B_{i}}}\)\} affected by the uncertainties exerting similar influence, it would still be feasible to derive accurate analytical solutions for the HECPs without resorting to complex post-processing. Unfortunately, in real-world scenarios, such idealized assumptions rarely hold, making the direct acquisition of consistent and accurately modeled \{\({\boldsymbol{A_{i}}}\)\} and \{\({\boldsymbol{B_{i}}}\)\} a fundamental challenge.

\begin{figure*}[!htbp]% 
	\centering
	\includegraphics[width=0.95\textwidth]{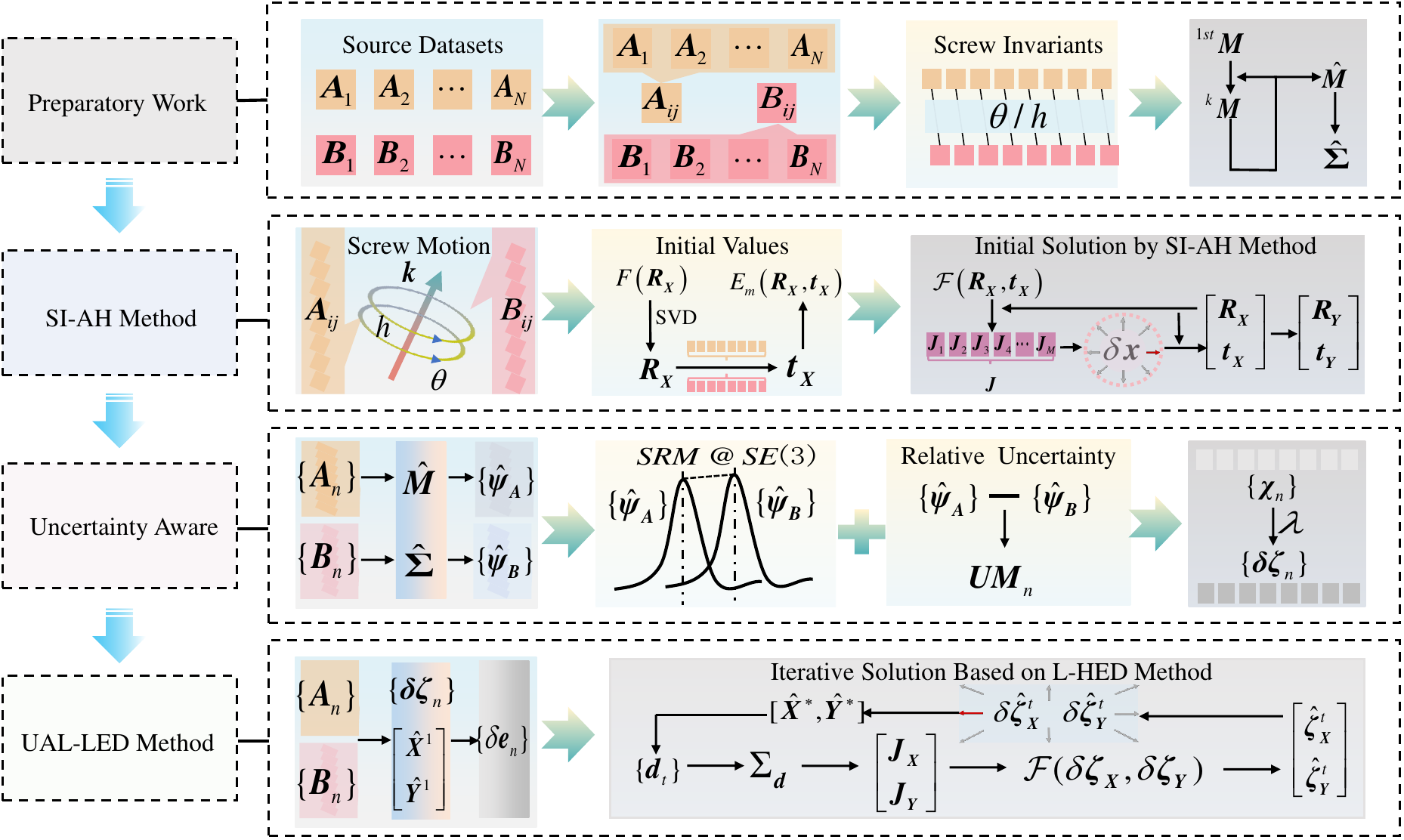} 
	\caption{\ \ \leftskip=0pt \rightskip=0pt plus 0cm The framework for the proposed methods.}
	\label{FIG_1}
\end{figure*}

In summary, due to the presence of AU and EU in real-world HEC problem, conventional solution methods often suffer from convergence to local optima or reduced global accuracy under high uncertainty. Furthermore it is a challenge of explicitly modeling AU and EU in HEC. Besides, uncertainty also impairs the precision of source data selection, thereby limiting the effectiveness of traditional filtering techniques. The framework proposed in this paper which shown in Figure \ref{FIG_1} presents a general estimation HECPs method, with the fundamental aim of addressing the limitations of existing approaches under highly uncertain data sources, thereby enhancing overall applicability. 

Based on the equivalence and transformability between the  $\boldsymbol{AX=XB}$ and $\boldsymbol{AX=YB}$ formulations, the framework first exploits calibration invariants under the $\boldsymbol{AX=XB}$ configuration to filter human-induced operational errors and remove evidently erroneous data. The scale-invariant analytical heuristic method (SI-AH) is then constructed in this form to enable joint iteration over both the rotational and translational components. This iterative result serves as an initial estimate for the subsequent global iterative solving of the $\boldsymbol{AX=YB}$ problem. To achieve high-precision estimation of calibration parameters $\boldsymbol{X}$ and $\boldsymbol{Y}$, a synchronized iterative model for the $\boldsymbol{AX=YB}$ problem is developed based on convex optimization and Lie group theory. The Lie algebra-based heuristic escape descent (L-HED) method is introduced to enable globally synchronized iteration of the calibration matrices $\boldsymbol{X}$ and $\boldsymbol{Y}$. Ultimately, the L-HED method is refined through the integration of the constructed relative uncertainty metric, resulting in an uncertainty-aware global optimization approach, termed Uncertainty-Aware Lie Algebra-Based Heuristic Escape Descent (UAL-HED). This method effectively addresses the challenges posed by highly uncertain data sources in HEC problem.

The main contribution of the article, in terms of theory, the L-HED method was proposed, an effective global optimization technique that integrates local iterative procedures based on Lie algebra with heuristic-driven mechanisms designed to escape local minima, specifically tailored to adhere to Euclidean group constraints. Considering the absence of calibration invariants in the $\boldsymbol{AX=YB}$ equation, a novel class Sharpe Ratio Metric under the Euclidean group (\textit{SRM@SE(3)}) for $\boldsymbol{AX=YB}$ problems was introduced in order to accurately captures the relative uncertainty between datasets \{\({\boldsymbol{A_{i}}}\)\} and \{\({\boldsymbol{B_{i}}}\)\}, thus circumventing explicit uncertainty modeling challenges. Incorporating \textit{SRM@SE(3)} into the L-HED method results in the UAL-HED method, which effectively addresses optimization problems involving highly uncertain data sources, bridging the gap left by traditional methods in accurately handling such data. Moreover, to accommodate the $\boldsymbol{AX=XB}$ problem, the SI-AH method which provides robust initial estimates beneficial to the subsequent iterative optimization processes was proposed. 

In terms of experimental validation, extensive numerical simulation and real-world experiments substantiate the effectiveness of the proposed methods and metrics. Furthermore, numerical simulations comprehensively evaluate the performance of classical numerical techniques, an advanced iterative method, and the approach in this paper under varying noise conditions, clearly identifying scenarios where each method excels. Given the bi-invariant measure properties of the Euclidean group, the paper also discusses the form of heuristic escape indicators appropriate for different problem contexts. To address the challenge posed by the absence of true calibration matrices $\boldsymbol{X}$ and $\boldsymbol{Y}$ in real-world scenarios, various residual formulations in simulations was investigated, analyzing their effectiveness in accurately reflecting calibration errors, and identifying optimal formulations for assessing the practical effectiveness of the proposed methods in real-word.

The remainder of this paper is organized as follows: Section 2\ref{sec:1.5} reviews related work on HECPs estimation and different of methods for improving estimation accuracy. Section 3\ref{sec:2} presents preliminary work, including a review of the mathematical formulas necessary for theoretical derivation and a preliminary analysis of the HEC problem. Section 4\ref{sec:3} details our methods, including the derivation of the L-HED approach, construction of the \textit{SRM@SE(3)} to achieve relative uncertainty modeling of source data, and the SI-AH solution method. Section 5\ref{sec:4} evaluates and discusses the effectiveness of the proposed methods and metrics through experimental results. Finally, Section 6\ref{sec:5} concludes the article. 

\section{2. Related work}
\label{sec:1.5}
The motivation behind HEC problem is to resolve the transformation relationship between a robot and its corresponding measurement sensor \cite{w22}. To accurately estimate the calibration matrix, various researchers have employed different representation methods since the problem's inception, such as HTMs \cite{w23}, Quaternions \cite{w24}, Lie-algebra \cite{w25}, Axis-angle \cite{w26}, Euler-angles \cite{w50}, and Kronecker product \cite{w27}, etc. The solution process can also be categorized into synchronous solving of rotation and translation components of the calibration matrix \cite{w28}, or decoupled solving \cite{w29}. Similarly, HECPs $\boldsymbol{X}$ and $\boldsymbol{Y}$ can be solved either in a decoupled or synchronous manner \cite{w30}, \cite{w31}. However, decoupled methods are prone to cumulative and propagated errors. A notable advantage of performing synchronous solving is its ability to prevent the accumulation of errors during the final estimation stage, particularly in scenarios where the source data exhibits high uncertainty. In addition, some researchers have focused on selecting optimal source data to achieve higher calibration accuracy \cite{w32}, \cite{w33}, and ensuring accurate data correspondence such as \cite{w34}, \cite{w35}, \cite{w36}, \cite{w37}.

Regardless of the formulation or the effectiveness of data selection strategies, the inevitable challenge in real-world scenarios lies in the inability to obtain ideal source data. Consequently, the HEC problem is ultimately transformed into a least squares problem (\cite{w38}), a linear programming problem (\cite{w39}), or a nonlinear optimization problem (\cite{w51}). Given that linear methods are typically formulated to minimize algebraic errors(\cite{w53}), their achievable accuracy is inherently bounded, leading to sensitivity of their performance to the level of uncertainty present in the data. To address this limitation, solutions produced by linear solvers are frequently refined through subsequent nonlinear optimization to attain higher precision. The primary purpose of nonlinear methods is to minimize either algebraic error or set-based error through the use of nonlinear optimization techniques. However, traditional methods are still susceptible to the influence of uncertain disturbances, which adversely affect solution accuracy. Hence, obtaining a globally optimal solution has become a central objective within this research domain.objective in this field of research.

A globally optimal method was proposed in \cite{w40}, which solves the calibration equations using quaternions or screw motion constraints. Another approach, based on the minimization of an epipolar constraint-driven objective function via branch-and-bound, achieves a global optimum under the $L^1$-norm in \cite{w41}. A two-stage stochastic geometric optimization algorithm was introduced to search for global stochastic minima in \cite{w42}. In \cite{w43}, the authors applied the Cayley transform to decompose and unify the $\boldsymbol{AX=XB}$ and $\boldsymbol{AX=YB}$ problems, constructing an optimization equation from which an algebraic polynomial system is formulated. A constrained local minimum set is derived using Wu's elimination method, from which the best value is selected as the global optimal solution. A general least-squares-based approach guided by reprojection was proposed in \cite{w44}, enabling calibration with an arbitrary number of cameras. To address the issue of dependency on the minimum singular value of the regression matrix during simultaneous calibration, \cite{w45} transforms the optimization target into a linear matrix inequality, converting the problem into a convex optimization task and solving it iteratively using semidefinite programming. In \cite{w46}, a globally optimal point cloud registration is achieved through a coarse alignment algorithm based on four-point congruent sets and a refinement algorithm utilizing progressive adaptive variance minimization. Similarly, an iterative correction method of an initial calibration matrix was developed to achieve a globally optimal solution in \cite{w47}. Beyond these methods, neural networks have also been employed for estimating and solving the calibration matrix problem \cite{w48}, and generative adversarial networks have been utilized to perform HEC without requiring data correspondence, enabling estimation of the calibration matrix \cite{w49}.

As is well known, the source data used in HEC inherently contain errors. However, handling such uncertainty remains a mostly overlooked issue in \cite{w8}. The study in \cite{w14} addresses this by modeling source data errors as various Gaussian distributions, and derives the calibration matrix through the construction of a maximum likelihood estimation framework that satisfies the maximum likelihood condition. The work in \cite{w15} leverages the valuable information contained within the measurement uncertainties. It proposes a Gauss-Helmert-based model for simultaneous HEC sensors and scale estimation, demonstrating strong performance under high noise conditions. The framework proposed in \cite{w16} not only yields highly accurate estimates of hand-eye poses, but also provides reliable information regarding the uncertainty of the robot. This enables the correction of robot poses, facilitating simple and cost-effective robot calibration.

\section{3. Preliminaries}
\label{sec:2}
\subsection{3.1 Mathematical basis of matrix Lie groups}\label{sec:2.1}

\begin{figure}[!htbp]\centering
	\includegraphics[width=0.37\textwidth]{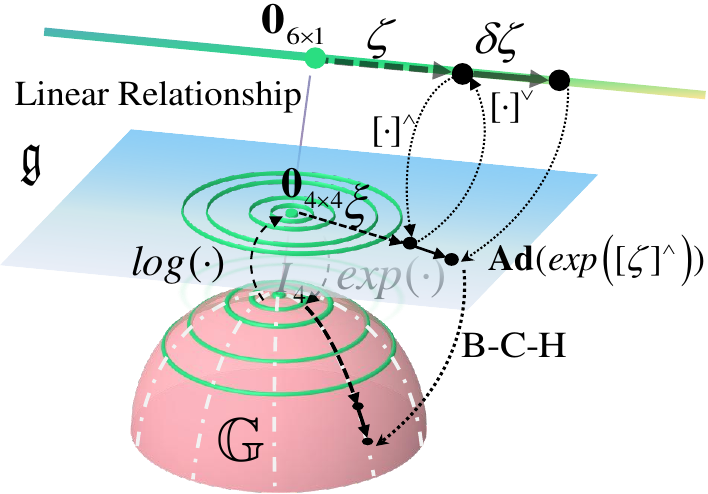}
	\caption{\ \ \leftskip=0pt \rightskip=0pt plus 0cm The relationship between the Lie groups and the Lie algebras.}
	\label{FIG_3}
\end{figure}

The Euclidean motion group, \(SE(3)\), is the semi-direct product of \(\bm{\mathit{IR}}^3\) with the special orthogonal group, \(SO(3)\), which commonly used to describe the motion of robots in a \( \textit{6}\)-dimensional space and the symbol for defining group operations is [\( \circ\)]. More accurately, the robot's pose can be denoted as \(\vectT \in \) \(SE(3)\) and can be expressed as a $4\times4$ homogeneous transformation matrix as
\begin{equation}
	\label{eq:1}
	{SE}(3) := \left\{ \left. \boldsymbol{T} = \left[ \begin{array}{rr} \boldsymbol{R} & \boldsymbol{t} \\ \mathbf{0}^{T} & \mathbf{1} \end{array} \right] \right| \, \boldsymbol{t} \in \mathbb{R}^3, \, \boldsymbol{R} \in SO(3) \right\},
\end{equation}
where the Lie algebra \( \mathfrak{se} \)(3) is shown as
\begin{equation}
	\label{eq:2}
	\mathfrak{se}(3) := \left\{ \left. [\boldsymbol{\zeta_t}]^{\wedge} = \left[ \begin{array}{cc} [\boldsymbol{\varphi}]^{\wedge} & \boldsymbol{\rho} \\ \mathbf{0}_{n \times 1}^{\top} & 0 \end{array} \right] \in \mathbb{R}^{4 \times 4} \right| \, \boldsymbol{\varphi}, \boldsymbol{\rho} \in \mathbb{R}^{3} \right\},
\end{equation}
where corresponding to the Lie group \(SE(3)\) is the \( \textit{6}\)-dimensional tangent space of the Lie group. The mapping relationships between the Lie algebra \( \mathfrak{g} \) and the Lie group \( \mathbb{G} \), as well as certain operational transformations, are shown in Figure \ref{FIG_3}. The exponential operation \( \textit{exp}(\cdot) \):  \(\mathfrak{g} \rightarrow \mathbb{G}\) and logarithm \( \textit{log}(\cdot) \): \(\mathbb{G} \rightarrow \mathfrak{g}\) operation establish a local diffeomorphism between a neighborhood of \( \mathbf{0}_{n} \) in the tangent space to a local neighborhood of the identity \( \bm{\mathit{I}}_{n} \) on the manifold. The isomorphism between the Lie algebra \( \mathfrak{g} \) and its vector space \(\mathbb{R}^{n}\) are indicated by \([\cdot]^{\wedge}\) : \(\mathbb{R}^{n} \rightarrow \mathfrak{g} \) and \([\cdot]^{\vee}\) : \(\mathfrak{g} \rightarrow \mathbb{R}^{n}\), respectively. For an element \( \boldsymbol{{\zeta}_t} \in \mathbb{R}^{6} \), which is a vector form of the Lie algebra, there exists the invertible transformation as
\begin{equation}
	\label{eq:3}
	\boldsymbol{T}=\textit{exp} \left([\boldsymbol{{\zeta}_t}]^{\wedge}\right)=\left[\begin{array}{cc}\textit{exp}\left([\boldsymbol{\varphi}]^{\wedge}\right) & \mathbf{V}_{3}(\boldsymbol{\varphi}) \boldsymbol{\rho} \\\mathbf{0}_{3 \times 1}^{\top} & 1\end{array}\right],
\end{equation}
\begin{equation}
	\label{eq:4}
	\boldsymbol{{\zeta}_t}=[\textit{log} (\vectT)]^{\vee}=\left[\begin{array}{c}{[\textit{log} (\boldsymbol{R})]^{\vee}} \\\mathbf{V}_{3}^{-1}\left([\textit{log} (\boldsymbol{R})]^{\vee}\right) \boldsymbol{t}\end{array}\right],
\end{equation}
where the operation \([\boldsymbol{\cdot}]^{-1}\) represents  the inverse transformation and \(\boldsymbol{t}\) represents a translation vector with respect to some reference frame. Correspondingly, \(\boldsymbol{R}\) describes the rotational transformation in space, where \(\boldsymbol{\varphi}\) is a vector space element associated with \(\boldsymbol{R}\). \(\mathbf{V}_{3} (\cdot)\) can be calculated using Rodrigues formula as
\begin{equation}
	\label{eq:5}
	\mathbf{V}_{3}(\boldsymbol{\varphi})=\mathit{I}_{3}+\frac{1-\cos \theta}{\theta^{2}}[\boldsymbol{\varphi}]_{\times}+\frac{\theta-\sin \theta}{\theta^{3}}[\boldsymbol{\varphi}]_{\times}^{2},
\end{equation}
where corresponding to the pure rotational motion with the axis passing through the origin and rotated by an angle \( \theta \), its definition is as: \(\theta:=\|\boldsymbol{\varphi}\|\). On top of that the operator \([\boldsymbol{\cdot}]_{\times}\) generates the skew-symmetric matrix corresponding to the input vector, whereas in reality  the operators \([\boldsymbol{\varphi}]_{\times}\) and \([\boldsymbol{\varphi}]^{\wedge}\) produce the same result for any arbitrary element \( \boldsymbol{\varphi} \in  \mathbb{R}^3\).

Given a general matrix Lie group, for elements sufficiently close to the identity \( \bm{\mathit{I}}_{n} \), the corresponding exponential \(\textit{exp}(\cdot)\) and logarithmic \(\textit{log}(\cdot)\) maps can be defined by Taylor expansions around the identity as
\begin{equation}
	\label{eq:6}
	e^{\boldsymbol{\xi}}=\sum_{k=0}^{\infty} \frac{\boldsymbol{\xi}^{k}}{k!},
\end{equation}
\begin{equation}
	\label{eq:7}
	\textit{log} (\boldsymbol{G})=\textit{log}(\boldsymbol{I}+( {G}-\boldsymbol{I}))=\sum_{k=1}^{\infty}(-1)^{k+1} \frac{(\boldsymbol{G}-\boldsymbol{I})^{k}}{k},
\end{equation}
where \(\boldsymbol{G} \in \mathbb{G}\) and \(\boldsymbol{\xi} \in \mathfrak{g}\), when \(\boldsymbol{G}\) is close to the identify matrix and \(\boldsymbol{\xi}\) is small, higher-order terms in Equations (\ref{eq:6}) and (\ref{eq:7}) may be neglected.  

The adjoint matrix for \(SE(n)\) is used to describe transformations of motion. For an element \( \vectT \in SE(3) \), the adjoint matrix is defined as 
\begin{equation}
	\label{eq:8}
	\mathbf{A d}\left(\vectT\right)=\left[\begin{array}{cc}\boldsymbol{R} & \mathbf{0}_{3} \\{[\boldsymbol{t}]_{\times} \boldsymbol{R}} & \boldsymbol{R}\end{array}\right] \in \mathbb{R}^{6 \times 6}.
\end{equation}

For any element \(\boldsymbol{{\xi}_t} \in \mathfrak{se}(3)\), the adjoint representation of the Lie algebra \(\mathfrak{se}(3)\) can be defined as
\begin{equation}
	\label{eq:9}
	\mathbf{a d}({\boldsymbol{{\xi}_t}})=\mathbf{a d}({[\boldsymbol{\zeta_t}]^{\wedge}})=\left[\begin{array}{cc}{[\boldsymbol{\varphi}]_{\times}} & \mathbf{0}_{3} \\{[\boldsymbol{\rho}]_{\times}} & {[\boldsymbol{\varphi}]_{\times}}\end{array}\right] \in \mathbb{R}^{6 \times 6}.
\end{equation}

For an arbitrary element \(\boldsymbol{\xi} \in \mathfrak{se}\textit{(3)}\) which is a matrix of the Lie algebra, there is a profound connection between the adjoint matrix and the adjoint operator which defined as
\begin{equation}
	\label{eq:10}
	\mathbf{A d}_{\vectT} \boldsymbol{\xi}=\left.\frac{d}{d t}\left(\vectT e^{t \boldsymbol{\xi}} \vectT^{-1}\right)\right|_{t=0}=\vectT \boldsymbol{\xi} \vectT^{-1}.
\end{equation}

A point worth noting is that if operation  \([\cdot]^{\vee}\) is applied to Equation (\ref{eq:10}), an important transformation can be obtained as 
\begin{equation}
	\label{eq:11}
	[\mathbf{A d}_{\vectT} \boldsymbol{\xi}]^\vee=[\vectT \boldsymbol{\xi} \vectT^{-1}]^\vee=\mathbf{A d}\left(\vectT\right) [\boldsymbol{\xi}]^\vee.
\end{equation}

The transformations of Lie groups and Lie algebras can be related through the Baker-Campbell-Hausdorff (BCH) formula. Taking the Euclidean group of transformations as an example, let \(\boldsymbol{X}\)=\textit{exp}(\([\boldsymbol{\zeta_x}]^{\wedge}\)) and \(\boldsymbol{Y}\)=\textit{exp}(\([\boldsymbol{\zeta_y}]^{\wedge}\)) be two elements in the Lie group \(SE(3)\), if \(\textit{exp}([\boldsymbol{\zeta}]^{\wedge})=\boldsymbol{X}\boldsymbol{Y}\), the BCH formula can be expressed as
\begin{equation}
	\label{eq:12}
	\begin{aligned}\boldsymbol{\zeta}=\boldsymbol{\zeta_x} & +\boldsymbol{\zeta_y}+\frac{1}{2} \mathbf{a d}({\boldsymbol{\zeta_x}})\boldsymbol{\zeta_y}+\frac{1}{12} \mathbf{a d}({\boldsymbol{\zeta_x}}) \mathbf{a d}({\boldsymbol{\zeta_x}})\boldsymbol{\zeta_y}\\& +\frac{1}{12} \mathbf{a d}({\boldsymbol{\zeta_y}}) \mathbf{a d}({\boldsymbol{\zeta_y}})\boldsymbol{\zeta_x}+\cdots.\end{aligned}
\end{equation}

When \(\boldsymbol{\zeta_y}\) is sufficiently small, higher-order terms in the BCH formula can be neglected. This leads to the following useful result as
\begin{equation}
	\label{eq:13}
	\left[\textit{log} \left(\textit{exp} \left([-\boldsymbol{\zeta}]^{\wedge}\right) \textit{exp} \left([\boldsymbol{\zeta}+\boldsymbol{\delta\zeta}]^{\wedge}\right)\right)\right]^{\vee} \approx \mathcal{J}_{r}(\boldsymbol{\zeta}) \boldsymbol{\delta\zeta},
\end{equation}
where \(\mathcal{J}_{r}(\boldsymbol{\zeta})\) represents the right Jacobian matrix of an element belonging to \( \mathfrak{se} \)(3) and \(\boldsymbol{\delta\zeta}\) is the perturbation of the Lie algebra element \(\boldsymbol{\zeta}\). The right Jacobian matrix can be calculated as
\begin{equation}
	\label{eq:14}
	\begin{aligned}\mathcal{J}_{r}(\boldsymbol{\zeta})= & \boldsymbol{I}_{6}-\frac{4- \theta \sin ( \theta)-4 \cos ( \theta)}{2 \theta^{2}} \mathbf{a d}(\zeta) \\& +\frac{4 \theta-5 \sin (\theta)+\theta \cos (\theta)}{2 \theta^{3}}\left(\mathbf{a d}(\zeta)\right)^{2} \\& -\frac{2-\theta \sin (\theta)-2 \cos (\theta)}{2 \theta^{4}}\left(\mathbf{a d}(\zeta)\right)^{3} \\& +\frac{2 \theta-3 \sin (\theta)+\theta \cos (\theta)}{2 \theta^{5}}\left(\mathbf{a d}(\zeta)\right)^{4}.\end{aligned}
\end{equation}

For a set of the Euclidean group \{\({\boldsymbol{A_{i}}}\)\} consisting of a continuum of elements, a fundamentally important problem is to obtain the mean and variance constructed from all the elements, where \(n\) represents the number of samples, the mean \(\boldsymbol{M_{A}}\) and variance \(\boldsymbol{\Sigma}_{\boldsymbol{A}}\) can be calculated as
\begin{equation}
	\label{eq:15}
	\sum_{i=1}^{n} \textit{log}\left(\boldsymbol{M_{A}}^{-1} \boldsymbol{A_{i}}\right)=\mathbb{O}_{4}
\end{equation}
\begin{equation}
	\label{eq:16}
	\boldsymbol{\Sigma}_{\boldsymbol{A}}= \frac{1}{n}\sum_{i=1}^{n} \textit{log} ^{\vee}\left(\boldsymbol{M_{A}}^{-1} \boldsymbol{A_{i}}\right)\left[\textit{log} ^{\vee}\left(\boldsymbol{M_{A}}^{-1} \boldsymbol{A_{i}}\right)\right]^{T}.
\end{equation}

Given \{\({\boldsymbol{A_{i}}}\)\} with the cloud of frames \(\boldsymbol{A_{i}}\) clustering around \(\boldsymbol{M_{A}} \in \mathbb{R}^{4 \times 4}\), decentralization can be achieved by used for \(\boldsymbol{M_{A}}\), and the variance \(\boldsymbol{\Sigma}_{\boldsymbol{A}} \in \mathbb{R}^{6 \times 6}\) can be calculated to describe the dispersion of the cloud of frames.

\subsection{3.2. AX = YB problem analysis for uncertain source dataset}\label{sec:2.2}

The problem of solving for the unknown
transformation \(\vectXX\) and \(\vectYY\) in the equation \(\vectAA\)\(\vectXX\)=\(\vectYY\)\(\vectBB\), which is essentially a two-frame sensor calibration problem: given pairs of homogeneous rigid-body transformation matrices \{(\(\boldsymbol{A_{i}},\boldsymbol{B_{i}}\))\}, where each \(\boldsymbol{A_{i}}\) and \(\boldsymbol{B_{i}}\) is a \(4 \times 4\) transformation matrix belonging to the Special Euclidean group from sensor (probe, camera, etc.) readings, can be expressed as
\begin{equation}
	\label{eq:17}
	\begin{array}{l}\boldsymbol{R}_{A} \boldsymbol{R}_{X}=\boldsymbol{R}_{Y} \boldsymbol{R}_{B},  \boldsymbol{R}_{A} \boldsymbol{t}_{X}+\boldsymbol{t}_{A}=\boldsymbol{R}_{Y} \boldsymbol{t}_{B}+\boldsymbol{t}_{Y}.\end{array}
\end{equation}

In an ideal scenario, minimizing the solution error, either analytically or iteratively, based on the obtained source data, turns out to be a desirable solution. In parallel, the associated solving procedures for obtaining the final solutions are highly dependent on the source dataset  \{(\(\boldsymbol{A_{i}},\boldsymbol{B_{i}}\))\}. More accurately, a higher solution accuracy means that for each pair \{(\(\boldsymbol{A_{i}},\boldsymbol{B_{i}}\))\}, there are two constraints: one is that their sequences correspond to each other, and the other is that the transformation between them is solely influenced by \(\vectXX\) and \(\vectYY\), with no other factors involved.

\begin{figure}[!htbp]\centering
	\includegraphics[width=0.35\textwidth]{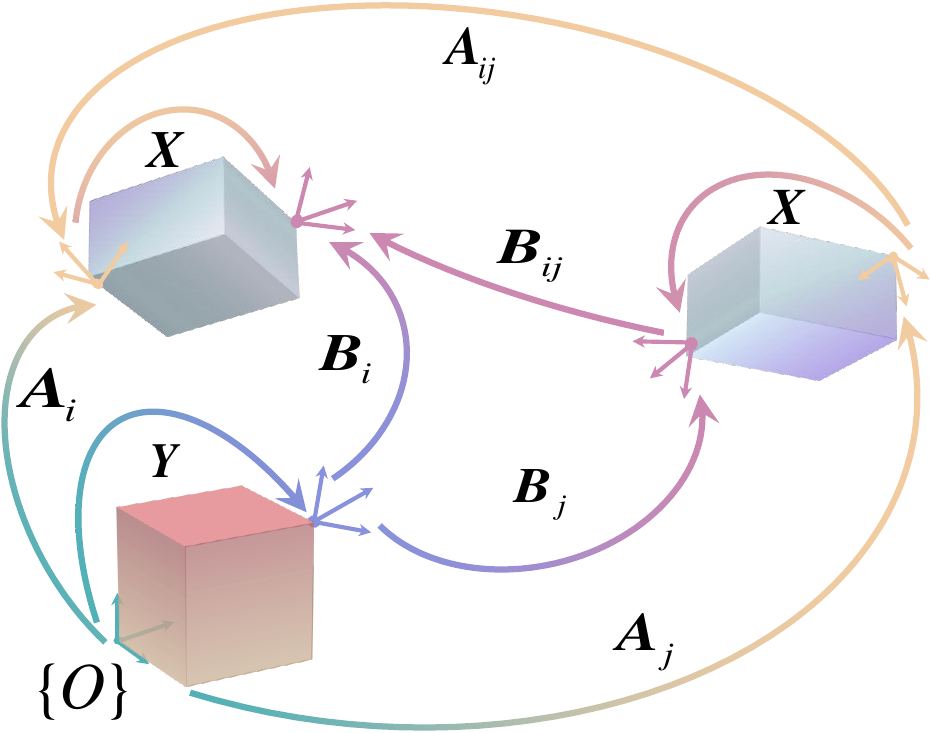}
	\caption{\ \ \leftskip=0pt \rightskip=0pt plus 0cm The transformation between the \(\vectAA\)\(\vectXX\)=\(\vectYY\)\(\vectBB\)  problem and the \(\vectAA\)\(\vectXX\)=\(\vectXX\)\(\vectBB\)  problem.}
	\label{FIG_4}
\end{figure}
The first constraint can be solved through invariants in the calibration process. As illustrated in Figure \ref{FIG_4}, when there are two sets of motions, the unknown matrice \(\vectYY\) exhibits invariance, the problem of solving for two unknown matrices \(\vectAA\)\(\vectXX\)=\(\vectYY\)\(\vectBB\) can be reformulated as a single pose matrix solving problem  \(\vectAA\)\(\vectXX\)=\(\vectXX\)\(\vectBB\), which leads to the following as
\begin{equation}
	\label{eq:18}
{\vectAA_j}^{-1}{\vectAA_i}\left(\vectXX\right)=\left(\vectXX\right){\vectBB_j}^{-1}{\vectBB_i},
\end{equation}

Let (\({\vectAA_j}^{-1}{\vectAA_i}\)) be defined as \(\vectAA_{ij}\) and (\({\vectBB_j}^{-1}{\vectBB_i}\)) be defined as \(\vectBB_{ij}\), the problem \(\vectAA\)\(\vectXX\)=\(\vectYY\)\(\vectBB\) is transformed into as
\begin{equation}
	\label{eq:19}
	\vectAA_{ij}\vectXX =\vectXX\vectBB_{ij},
\end{equation}
where is essentially a one-frame sensor calibration problem, and the transformation matrix \(\vectXX\) makes insignificant effect in the rotation angle \(\theta\) around the axis of rotation and the displacement \({h}\) along the rotation axis per unit angle. 

Hence, for the source dataset \{(\(\boldsymbol{A_{i}},\boldsymbol{B_{i}}\))\}, if the elements constructed in Equation (\ref{eq:19}) satisfy the conditions of Equation (\ref{eq:20}),  it can be considered that the sequences of \(\boldsymbol{A_{i}}\) and \(\boldsymbol{B_{i}}\) correspond to each other, and vice versa.  
\begin{equation}
	\label{eq:20}
	\left\{
		\begin{array}{l}
		\left|\theta_{\vectBB_{ij}} - \theta_{\vectAA_{ij}}\right| < \boldsymbol{\varepsilon}_{\theta}, \quad \text{where} \quad \boldsymbol{\varepsilon}_{\theta} \to 0 \\
		\left|{h}_{\vectBB_{ij}} - {h}_{\vectAA_{ij}}\right| < \varepsilon_{{h}}, \quad \text{where} \quad \varepsilon_{{h}} \to 0.
		\end{array}
		\right.
\end{equation}

However, when the source dataset  \{(\(\boldsymbol{A_{i}},\boldsymbol{B_{i}}\))\} does not satisfy the second constraint, it becomes difficult for Equation (\ref{eq:20}) to be satisfied as well. The condition that leads to the failure to satisfy the second constraint is that, for a pair of \(\boldsymbol{A_{i}} \) and \( \boldsymbol{B_{i}} \), the motion contained in \( \boldsymbol{A_{i}} \) cannot be obtained precisely from the motion contained in \( \boldsymbol{B_{i}} \) through the homogeneous transformation matrices \( \vectXX \) and \( \vectYY \). Such the source dataset is defined as uncertain source dataset, which implies that there is some level of uncertainty between the data pairs (\(\boldsymbol{A_{i}},\boldsymbol{B_{i}}\)).

Unfortunately, for robotic systems, especially those with large-scale or heavy loads, the above-mentioned problem is often encountered. The fundamental reason, as described in Section 1\ref{sec:1}, is that the data flow information obtained from different apparatus does not exact. Therefore, for the \(\vectAA\)\(\vectXX\)=\(\vectYY\)\(\vectBB\) problem with uncertain source dataset, regarding the accuracy of the propagation formula, the results remain demonstrably suboptimal, the error distance metric should be defined as
\begin{equation}
	\label{eq:21}
	[\textit{log}({\boldsymbol{d}})]^\vee=[\textit{log}(\boldsymbol{A} \vectXX \boldsymbol{B}^{-1} \vectYY^{-1})]^\vee-\boldsymbol{\delta \vecte},
\end{equation}
where \(\boldsymbol{\delta\vecte}\) is the error term that represents the uncertainty in the source dataset. 

Consequently, it is well known that there is no bi-invariant distance on \(SE(3)\), so the intrinsic significance of minimizing the solution error remains an unresolved question, e.g, \( \vecte\left(\vectAA_{i} \vectXX, \vectYY \vectBB_{i}\right)\) and \(\vecte\left(\vectBB_{i} \vectXX^{-1}, \vectYY^{-1} \vectAA_{i}\right)\) may lead to divergent solution outcomes. Thus, the accuracy of the final solution may be fundamentally contingent upon the specific formulation of the chosen minimization model. 

With regard to the methodology for solving the problem, albeit the analytical solution can yield good results, for uncertain source datasets, the analytical solution is often prone to errors, therefore, an iterative solution is a good choice for this type of problem. For the problem in Equation (\ref{eq:17}), whether separating the rotation \(\boldsymbol{R}\) and translation \(\boldsymbol{t}\) for iteration or iterating the \( \vectXX \) and \( \vectYY \) matrices separately, both approaches lead to the propagation of computational errors. Thus, the optimal iteration method is synchronized iteration. Additionally, the best iterative model should be independent of the initial values; the initial values only affect the efficiency of the iteration. As a result, the convex optimization theory plays a particularly important role in the \(\vectAA\)\(\vectXX\)=\(\vectYY\)\(\vectBB\) problem for uncertain source dataset. 

In summary, the formulation of the objective function and information processing of source data constitute a pivotal factor in the precise resolution of the \(\vectAA\)\(\vectXX\)=\(\vectYY\)\(\vectBB\) problem for uncertain source datasets. Furthermore, the construction of the distance metric constitutes an ambiguity factor.

\section{4. Methodology}\label{sec:3}
\subsection{4.1. Multi-layer local convex optimization based exact calibration model using lie group}\label{sec:3.1}

Under ideal circumstances, for the \(\vectAA\)\(\vectXX\)=\(\vectYY\)\(\vectBB\) problem, there exists a pair of ideal solutions, which can be defined as \(\vectXX_o\) and \(\vectYY_o\). Building upon this ensemble of ideal solutions, the associated distance metric can be  constructed as
\begin{equation}
	\label{eq:22}
	\boldsymbol{d}_o=\boldsymbol{A} \vectXX_o \boldsymbol{B}^{-1} \vectYY_o^{-1} = \bm{\mathit{I}}_{4}.
\end{equation}

However, when the \(\vectAA\)\(\vectXX\)=\(\vectYY\)\(\vectBB\) problem originating from an uncertain source dataset, securing a ideal solution is virtually unattainable. For a particular ensemble of solutions as \(\vectXX_i\) and \(\vectYY_i\), the error metric can be defined based on Equation (\ref{eq:22}) as
\begin{equation}
	\label{eq:23}
	\begin{aligned}
	\boldsymbol{e}_i=\boldsymbol{A} \vectXX_i &\boldsymbol{B}^{-1} \vectYY_i^{-1}\\&=\boldsymbol{A} (\vectXX_o+\Delta\vectXX) \boldsymbol{B}^{-1} (\vectYY_o^{-1}+\Delta\vectYY^{-1}).
	\end{aligned}
\end{equation}

Since \(\boldsymbol{e}_i\) is close proximity to \(\boldsymbol{d}_o\), \(\boldsymbol{e}_i\) can be obtained as a minor perturbation of the identity element within \( \vectT \in SE(3) \) Lie group framework. Through using the first-order approximation, \(\textit{log}(\boldsymbol{e})\) can be expressed based on Equation (\(\ref{eq:7}\)) as
\begin{equation}
	\label{eq:24}
	\textit{log} (\boldsymbol{e}_i) \approx \boldsymbol{e}_i-\boldsymbol{I}_{4}.
\end{equation}

By integrating Equations (\ref{eq:22}) through (\ref{eq:24}), the Lie group and Lie algebra maintain a robust linear relationship in the vicinity of the identity element, and simultaneously ignoring high-order errors, then Equation (\ref{eq:24}) is transformed as
\begin{equation}
	\label{eq:27}
	\textit{log} (\boldsymbol{e}_i) = \boldsymbol{A} \Delta\vectXX \boldsymbol{B}^{-1} \vectYY^{-1} + \boldsymbol{A} \vectXX \boldsymbol{B}^{-1} \Delta\vectYY^{-1}.
\end{equation}

Given the closed-loop relationship \(\vectAA\)\(\vectXX\)=\(\vectYY\)\(\vectBB\), Equation (\ref{eq:27}) can be expressed as
\begin{equation}
	\label{eq:28}
	\begin{aligned}
		\textit{log} (\boldsymbol{e}_i) =\boldsymbol{A} \Delta\vectXX \vectXX^{-1}\boldsymbol{A}^{-1} +  \vectYY\Delta\vectYY^{-1}.
	\end{aligned}
\end{equation}

Let \(\boldsymbol{X}\)=\textit{exp}(\([\boldsymbol{\zeta_X}]^{\wedge}\)), \(\boldsymbol{Y}\)=\textit{exp}(\([\boldsymbol{\zeta_Y}]^{\wedge}\)) and  \(\boldsymbol{A}\)=\textit{exp}(\([\boldsymbol{\zeta_A}]^{\wedge}\)), and using Equations (\ref{eq:11}) and (\ref{eq:28}) can be rewritten as
\begin{equation}
	\label{eq:29}
	\begin{aligned}
		[\textit{log} (\boldsymbol{e})]^{\vee}=&\mathbf{A d}\left(\boldsymbol{A}\right)[\Delta\boldsymbol{X} \boldsymbol{X}^{-1}]^{\vee}+[\vectYY\Delta\vectYY^{-1}]^\vee\\
		=&\mathbf{A d}\left(\boldsymbol{A}\right)[\Delta \textit{exp}([\boldsymbol{\zeta_X}]^{\wedge})\textit{exp}([\boldsymbol{-\zeta_X}]^{\wedge})]^{\vee}\\+& [\textit{exp}([\boldsymbol{\zeta_Y}]^{\wedge})\Delta \textit{exp}([\boldsymbol{-\zeta_Y}]^{\wedge})]^{\vee}.
	\end{aligned}
\end{equation}

According to the definition in Equation (\ref{eq:23}), \([\Delta \textit{exp}([\boldsymbol{\zeta}]^{\wedge})\textit{exp}([\boldsymbol{\zeta}^{-1}]^{\wedge})]^{\vee}\) can be approximated as
\begin{equation}
	\label{eq:30}
	\begin{aligned}
		&[\Delta \textit{exp}([\boldsymbol{\zeta}]^{\wedge})\textit{exp}([\boldsymbol{-\zeta}]^{\wedge})]^{\vee} \\ 
		&=[[\textit{exp}([\boldsymbol{\zeta+\delta\zeta}]^{\wedge})-\textit{exp}([\boldsymbol{\zeta}]^{\wedge})]\textit{exp}([\boldsymbol{-\zeta}]^{\wedge})]^{\vee} \\
		&=[\textit{exp}([\boldsymbol{\zeta+\delta\zeta}]^{\wedge})\textit{exp}([\boldsymbol{-\zeta}]^{\wedge})-\boldsymbol{I}_{4}]^{\vee}
		\\
		& \approx [\textit{log}(\textit{exp}([\boldsymbol{\zeta+\delta\zeta}]^{\wedge})\textit{exp}([\boldsymbol{-\zeta}]^{\wedge}))]^{\vee}.
	\end{aligned}
\end{equation}

For the right Jacobian matrix defined in Equation (\(\ref{eq:13}\)), the corresponding left Jacobian matrix relates to it as
\begin{equation}
	\label{eq:31}
	\mathcal{J}_{l}(\boldsymbol{\zeta})=\mathcal{J}_{r}(-\boldsymbol{\zeta}).
\end{equation}

By combining Equations (\ref{eq:30}) and (\ref{eq:31}), Equation (\ref{eq:30}) can be rewritten as
\begin{equation}
	\label{eq:32}
	[\textit{log}(\textit{exp}([\boldsymbol{\zeta+\delta\zeta}]^{\wedge})\textit{exp}([\boldsymbol{-\zeta}]^{\wedge}))]^{\vee}=\mathcal{J}_{l}(\boldsymbol{\zeta})\delta\boldsymbol{\zeta}.
\end{equation}

Through the manipulation of Equation (\ref{eq:29}) to Equation (\ref{eq:32}), the deviation \(\textit{log} (\boldsymbol{e})\) can be obtained as
\begin{equation}
	\label{eq:33}
	\begin{aligned}
		[\textit{log} (\boldsymbol{e})]^{\vee} =&\mathbf{A d}\left(\boldsymbol{A}\right)\mathcal{J}_{l}(\boldsymbol{\zeta_X})\delta\boldsymbol{\zeta_X}+ \mathcal{J}_{r}(-\boldsymbol{\zeta_Y})\delta\boldsymbol{\zeta_Y}\\ =& \boldsymbol{J}_{X}\delta\boldsymbol{\zeta_X}+\boldsymbol{J}_{Y}\delta\boldsymbol{\zeta_Y}.
	\end{aligned}
\end{equation}

Based on Equation (\ref{eq:33}), the solution to the \(\vectAA\)\(\vectXX\)=\(\vectYY\)\(\vectBB\) problem can be formulated as a convex optimization problem, and the goal is to identify the optimal solution \(\hat{\vectXX}^{*}\) and \(\hat{\vectYY}^{*}\) that can converge infinitely close to the theoretically established solution \(\vectXX_o\) and \(\vectYY_o\). Based on Equation (\ref{eq:24}), a new deviation metric \(\boldsymbol{\gamma} \in \mathbb{R}^{6}\) is defined as
\begin{equation}
	\label{eq:34}
	\boldsymbol{\gamma} = [\textit{log} (\boldsymbol{e})]^{\vee} -(\boldsymbol{J}_{X}\delta\boldsymbol{\zeta_X}+\boldsymbol{J}_{Y}\delta\boldsymbol{\zeta_Y}). 
\end{equation}

Considering that \(\boldsymbol{\zeta} = [\boldsymbol{\varphi}, \boldsymbol{\rho}]^{T}\), where the operation \([\cdot]^{T}\) denotes the transpose, a fundamentally important problem is that the rotational component \(\boldsymbol{\varphi}\) and translational component \(\boldsymbol{\rho}\) have varying physical units, and their optimization variables may differ greatly in orders of magnitude. In order to avoid this issue, the Mahalanobis distance is used to define the objective function based on the \(N\) pairs of data sets \{(\(\boldsymbol{A_{n}},\boldsymbol{B_{n}}\))\} as 
\begin{equation}
	\label{eq:35}
	\underset{\delta \boldsymbol{\zeta}_{X},\delta \boldsymbol{\zeta}_{Y}}{\arg \min }~\mathcal{F}(\delta\boldsymbol{\zeta_X},\delta\boldsymbol{\zeta_Y})= \sum_{n=1}^{{N}}\boldsymbol{\gamma}^{T}_{n}\boldsymbol{\Sigma}_{\boldsymbol{e}}\boldsymbol{\gamma}_{n}. 	
\end{equation}

The covariance matrix \(\boldsymbol{\Sigma}_{\boldsymbol{e}} \) primarily serves to standardize the magnitude, and the objective function \(\mathcal{F}(\delta\boldsymbol{\zeta_X},\delta\boldsymbol{\zeta_Y})\) is used to minimize the deviation \(\boldsymbol{\gamma}\) in order to \(\boldsymbol{e} \to \bm{\mathit{I}}_{4}\). Taking the construction of Equation (\ref{eq:23}) as an example, for a specific pair of estimated solutions \( \hat{\vectXX} \) and \( \hat{\vectYY} \), the covariance matrix \(\boldsymbol{\Sigma}_{\boldsymbol{e}}^{-1}\) can be calculated as
\begin{equation}
	\label{eq:36}
	\begin{aligned}
	\hat{\boldsymbol{\Sigma}}_{\boldsymbol{e}}&=diag(\\&\frac{1}{N-1} \sum_{n=1}^{N}\left([\textit{log} (\boldsymbol{e})]^{\vee}_{n}-\hat{\boldsymbol{\mu}}\right)\left([\textit{log} (\boldsymbol{e})]^{\vee}_{n}-\hat{\boldsymbol{\mu}}\right)^{T}),\\
	& \text{Let} ~\hat{\boldsymbol{\mu}}=\frac{1}{N} \sum_{n=1}^{N} [\textit{log} (\boldsymbol{e})]^{\vee}_{n},	
	\end{aligned}
\end{equation}
where the operation \(diag(\cdot)\) is to extract the diagonal elements and thereby construct a new diagonal matrix. Let \(\boldsymbol{J}_{X Y}=\left[\begin{array}{llll}\boldsymbol{J}_{X},\boldsymbol{J}_{Y}\end{array}\right], \quad \delta \boldsymbol{\zeta}_{XY}=\left[\begin{array}{lll}\delta \boldsymbol{\zeta}_{X}^{\mathrm{T}} , \delta \boldsymbol{\zeta}_{Y}^{\mathrm{T}}\end{array}\right]^{\mathrm{T}}\), Equation (\ref{eq:34}) can be simplified as
\begin{equation}
	\label{eq:37}
	\boldsymbol{\gamma} = [\textit{log} (\boldsymbol{e})]^{\vee} -\boldsymbol{J}_{X Y}\delta \boldsymbol{\zeta}_{X Y}. 	
\end{equation}

To mitigate oscillations during the iterative process, the gradients can be formulated as
\begin{equation}
	\label{eq:38}
	\begin{array}{c}\mathbf{v}_{t}=\beta \mathbf{v}_{t-1}+(1-\beta) \nabla \mathcal{F}[\left(\delta \boldsymbol{\zeta}_{X Y}\right)_{t-1}] \\\left(\delta \boldsymbol{\zeta}_{X Y}\right)_{t} =\left(\delta \boldsymbol{\zeta}_{X Y}\right)_{t-1} -\alpha \mathbf{v}_{t},\end{array}
\end{equation}
where \(\mathbf{v}_{t}\) is the momentum vector, which could accelerate convergence and reduce oscillations, and \(\alpha\) is the learning rate, and $t$ is the number of iterations, and \(\beta\) is the momentum factor. The core part of this gradient formula is the calculation of the gradient \(\nabla \mathcal{F}[\left(\delta \boldsymbol{\zeta}_{X Y}\right)_{t-1}]\), which can be expressed as
\begin{equation}
	\label{eq:39}
	\nabla \mathcal{F}[\left(\delta \boldsymbol{\zeta}_{X Y}\right)_{t-1}]=-2 \sum_{n=1}^{N} \left[\begin{array}{c}\boldsymbol{J}_{X, n} \\\boldsymbol{J}_{Y, n}\end{array}\right]\boldsymbol{\Sigma}_{e}\boldsymbol{\gamma}_{n}.
\end{equation}

By integrating Equations (\ref{eq:34})-(\ref{eq:39}) and substituting the initial parameters, the optimization objectives are updated until the stopping criteria are satisfied. Since the error metric involves matrix multiplication, which can cause the iterative process to become trapped in local optima, small stochastic perturbation terms \(\delta_{\textit{rand}}\) are introduced to Equation (\ref{eq:39}) when necessary to facilitate the computation of a globally optimal solution as
\begin{equation}
	\label{eq:26}
	\nabla \mathcal{F}[\left(\delta \boldsymbol{\zeta}_{X Y}\right)_{t-1}]_{\textit{rand}}=-2 \sum_{n=1}^{N} \left[\begin{array}{c}\boldsymbol{J}_{X, n} \\\boldsymbol{J}_{Y, n}\end{array}\right]\boldsymbol{\Sigma}_{e}\boldsymbol{\gamma}_{n}+\delta_{\textit{rand}}.
\end{equation}

Based on the adjustment of Equation (\(\ref{eq:26}\)), apply an stochastic perturbation when Equation (\(\ref{eq:35}\)) falls into a local optimum to enter the next local optimum calculation. For every calculation that has a promoting effect, output the current iteration result \(\hat{\delta\boldsymbol{\zeta}}_{\boldsymbol{X Y}}^{k}\). As such, an optimal solution \(\hat{\vectXX}^{*}\) and \(\hat{\vectYY}^{*}\) can be updated as:
\begin{equation}
	\label{eq:40}
	\begin{aligned}
		\left\{
			\begin{array}{l}
		\hat{\vectXX}^{k+1} = \textit{exp}([\hat{\boldsymbol{\zeta}}^{k}_{\boldsymbol{X}}+\hat{\delta\boldsymbol{\zeta}}^{k}_{\boldsymbol{X}}]^{\wedge}),\\
		\hat{\vectYY}^{k+1} = \textit{exp}([\hat{\boldsymbol{\zeta}}^{k}_{\boldsymbol{Y}}+\hat{\delta\boldsymbol{\zeta}}^{k}_{\boldsymbol{Y}}]^{\wedge}),
		\end{array}	
		\right.
	\end{aligned}
\end{equation}
where, \(\textit{exp}([\hat{\boldsymbol{\zeta}}^{k}_{\boldsymbol{X}}])=\hat{\vectXX}^{k}\) and \(\textit{exp}([\hat{\boldsymbol{\zeta}}^{k}_{\boldsymbol{Y}}])=\hat{\vectYY}^{k}\) are the the current optimal solutions of the \(\vectAA\)\(\vectXX\)=\(\vectYY\)\(\vectBB\) problem, and \(\hat{\vectXX}^{*}\) and \(\hat{\vectYY}^{*}\) are updated iteratively until either the maximum number of iterations is reached or the norm of \(\hat{\delta\boldsymbol{\zeta}}_{\boldsymbol{X Y}}\) falls below a predefined threshold.
A point worth noting is that the optimal solution \(\hat{\vectXX}^{*}\) and \(\hat{\vectYY}^{*}\) are not necessarily the ideal solutions \(\vectXX_o\) and \(\vectYY_o\), they can become infinitely close to the ideal solutions.  

\begin{figure}[!htbp]\centering
	\includegraphics[width=0.45\textwidth]{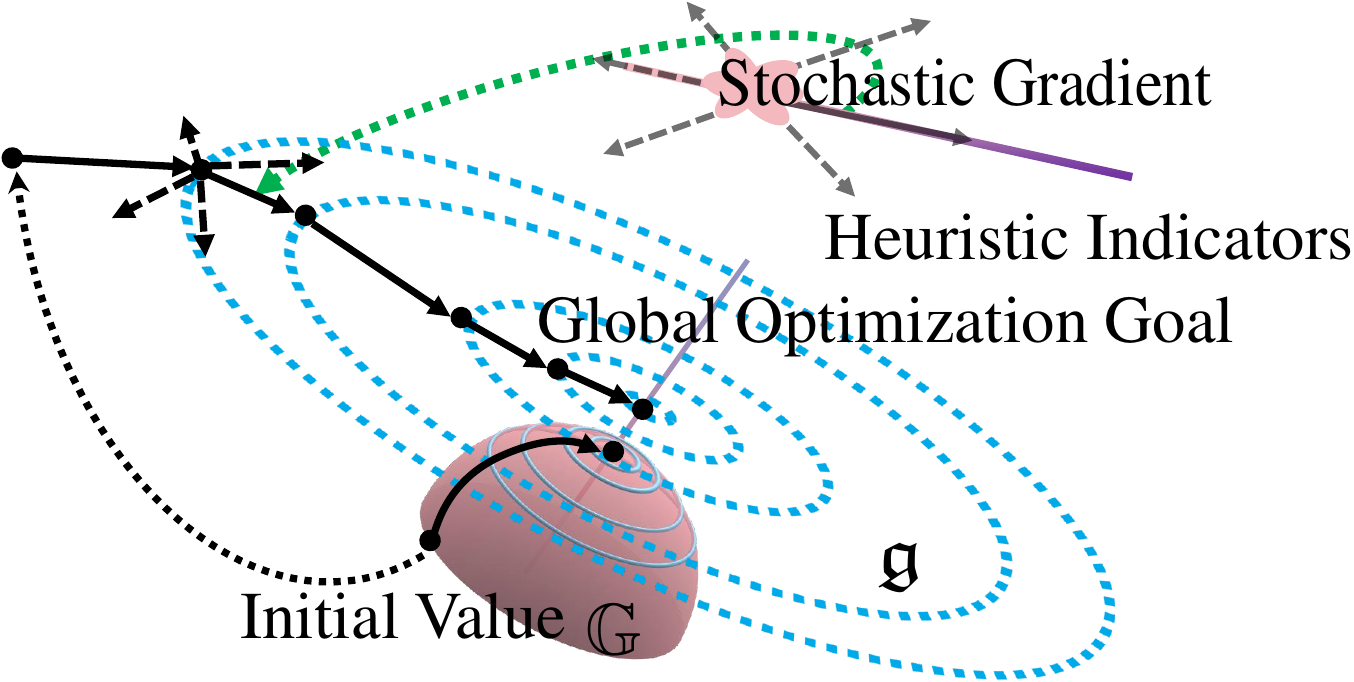}
	\caption{\ \ \leftskip=0pt \rightskip=0pt plus 0cm Schematic diagram of the L-HED method principle.}
	\label{s10}
\end{figure}

Simply adding stochastic perturbations in Equation (\(\ref{eq:26}\)) would disrupt the convergence properties; therefore, the perturbations need to be carefully selected. The selected tool is a Lie algebra-based residual metric, which is expressed as
\begin{equation}
	\label{eq:201}
		\begin{array}{l}
\begin{aligned}  \tau_{i} & ={\frac{1}{N} \sum_{i=1}^{N}\left\|\log \left(\boldsymbol{T}_{i}- {\boldsymbol{T}_{opt}}\right)\right\|} 
\end{aligned},
\end{array}
\end{equation}
where $\tau_{i}$ represents the heuristic metric at the $i^{th}$ iteration. The overall sequence of the iterative process is shown in Figure \ref{s10}.

To ensure that Equation (\ref{eq:35}) represents a convex optimization function, two conditions must be satisfied during the solution process. The first condition is that Equation (\ref{eq:34}) must constitute an affine function of \( \delta \boldsymbol{\zeta}_{X Y} \). Consequently, for each iteration given an values  \(\hat{\vectXX}^{k}\) and \(\hat{\vectYY}^{k}\), Equation (\ref{eq:33}) needs to be modified as
\begin{equation}
	\label{eq:41}
	\begin{aligned}
		[\textit{log} (\boldsymbol{e})]^{\vee}=&\mathbf{A d}\left(\boldsymbol{A}\right)\mathcal{J}_{l}(\boldsymbol{{\zeta_{\hat{\vectXX}^{k}}}})\delta\boldsymbol{\zeta_X}+ \mathcal{J}_{r}(-\boldsymbol{{\zeta_{\hat{\vectYY}^{k}}}})\delta\boldsymbol{\zeta_Y}\\ =& \boldsymbol{J}_{X}\delta\boldsymbol{\zeta_X}+\boldsymbol{J}_{Y}\delta\boldsymbol{\zeta_Y}.
	\end{aligned}
\end{equation}

The other is \(\boldsymbol{\Sigma}_{\boldsymbol{e}}^{-1}\) must be a positive definite matrix, if not, appropriate regularization, introduced through a small perturbation \( \boldsymbol{\epsilon} \), can be applied to the covariance matrix as
\begin{equation}
	\label{eq:42}
	\boldsymbol{\Sigma}_{\boldsymbol{e}} = \boldsymbol{\Sigma}_{\boldsymbol{e}}+\boldsymbol{\epsilon} \boldsymbol{I}_{6}.
\end{equation}

To sum up, the complete process of the convex optimization based exact calibration model using Lie Group theory is presented in Algorithm \ref{alg:1}.

\begin{algorithm}[!htbp]
    \caption{\small{\textbf{The synchronous iterative method for an exact solution of the AX=YB problem}}}
    \label{alg:1}
    \begin{algorithmic}
    \State \textbf{Input:} 
	\State The set $\{\boldsymbol{A_{i}}\}$ ; // \small{The source dataset from the controller indicates robot's pose in its base frame.}
    \State The set $\{\boldsymbol{B_{i}}\}$ ; // \small{The source dataset from the measurement equipment indicates robot's pose in measurement frame}
	\State $\hat{\vectXX}^{1}, \hat{\vectYY}^{1}$ ; // \small{Initial calibration results and the computational description is in Section \ref{sec:3.3}}
	\State $\alpha, \beta, \epsilon$; // \small{Iterative parameter}
	\State  $\delta \vecte$;  // \small{Uncertainty correction of the source dataset and the computational description is in Section \ref{sec:3.2}}
    \State \textbf{Output:} 
	\State $\hat{\vectXX}^{*}, \hat{\vectYY}^{*}$ ; // \small{Accurate calibration results}
    \end{algorithmic}
    \begin{algorithmic}[1] % Restart algorithmic with numbering
    \State Initialization: MaxIters, iter$\leftarrow$1, 
    \State Transform the input data into Lie algebra form base on Equations (\ref{eq:3})-(\ref{eq:5})
	\State \textbf{while} iter $<$ MaxIters \textbf{do}
    \State Compute the error metric \(\boldsymbol{e}_i\)  based on Equations (\ref{eq:21}) and (\ref{eq:23}) and convert the error metric into Lie algebra form 
    \State Compute the covariance matrice \(\boldsymbol{\Sigma}_{\boldsymbol{e}}\) based on Equation (\ref{eq:36})
    \State Compute Jacobian matrices \(\boldsymbol{J}_{X}\) and \(\boldsymbol{J}_{Y}\) based on Equations (\ref{eq:13}) and (\ref{eq:33})
    \State Construct $\mathcal{F}(\delta\boldsymbol{\zeta_X},\delta\boldsymbol{\zeta_Y})$ based on Equations (\ref{eq:34}) and (\ref{eq:35})
	\State Ensure the properties of convex functions based on Equations (\ref{eq:41}) and (\ref{eq:42})
    \State Compute an optimal perturbation \(\hat{\delta\boldsymbol{\zeta}}_{\boldsymbol{X Y}}^{k}\) based on Equations (\ref{eq:38}) and (\ref{eq:39})
    \State Update the optimal solution \(\hat{\vectXX}^{*}\) and \(\hat{\vectYY}^{*}\) based on Equation (\ref{eq:40})
	\State iter $\leftarrow$ iter + 1
	\State \textbf{until} $\hat{\delta\boldsymbol{\zeta}}_{\boldsymbol{X Y}} < \epsilon$
	\State \textbf{end while}
    \State Compute the final calibration results $\hat{\vectXX}^{*}$ and $\hat{\vectYY}^{*}$
    \end{algorithmic}
\end{algorithm}

\subsection{4.2. Uncertainty modeling of the source dataset}\label{sec:3.2}

The primary objective of this Section is to construct a convex optimization model for the \(\vectAA\)\(\vectXX\)=\(\vectYY\)\(\vectBB\) problem, and the optimal solution \(\hat{\vectXX}\) and \(\hat{\vectYY}\) can be obtained through synchronized iteration. Under these circumstances, it is assumed that the value of the error \(\boldsymbol{e}\) is influenced solely by the solution accuracy of \( \vectXX \) and \( \vectYY \). In other words, the distance metric \( \boldsymbol{d} \) defined as Equation (\ref{eq:22}) differs from \( \boldsymbol{I}_{4} \) precisely because the computed \( \hat{\vectXX} \) and \( \hat{\vectYY} \) are not the ideal solutions. Thus, synchronous iteration is required to pursue the best attainable solution by minimizing Equation \((\ref{eq:35})\).

Unfortunately, in scenarios where the source data contains inherent uncertainty, due to inherent errors present in each corresponding data pair (\(\boldsymbol{A_{i}}, \boldsymbol{B_{i}}\)), e,g., \(\boldsymbol{A_{i}}\) denotes the pose of the robot's end-effector relative to the robot base, disregarding the robot's precision inaccuracies, \(\boldsymbol{B_{i}}\) represents the pose of the robot's end-effector relative to the measurement apparatus, encompassing both the robot's precision errors and measurement uncertainties, the accuracy of the solutions \( \vectXX \) and \( \vectYY \) is consequently compromised. For this reason, when there is uncertainty in the source data used, Equation (\ref{eq:21}) should be used to correct Equation (\(\ref{eq:22}\)). So that the error term \(\boldsymbol{\delta \vecte}\) can be incorporated as 
\begin{equation}
	\label{eq:72}
	[\textit{log} (\boldsymbol{e})]^{\vee} = [\textit{log} (\boldsymbol{e})]^{\vee} - \boldsymbol{\delta \vecte}.
\end{equation}

Another perspective on the aforementioned issue is that the relationship between \(\boldsymbol{A_{i}}\) and \(\boldsymbol{B_{i}}\) is not merely  \(\vectAA\)\(\vectXX\)=\(\vectYY\)\(\vectBB\). Instead, it should incorporate a correction defined by Equation (\ref{eq:73}) using \(\boldsymbol{\delta \xi}\) which images the error from various sources  between \(\boldsymbol{A_{i}}\) and \(\boldsymbol{B_{i}}\) as
\begin{equation}
	\label{eq:73}
	\boldsymbol{d}=\boldsymbol{A} \vectXX \textit{exp}(\mathbf{A d}_{B} \boldsymbol{\delta \xi})\boldsymbol{B}^{-1} \vectYY^{-1}.
\end{equation}

Considering that modeling the uncertainties in the source data set inherently serves as a correction to the solution accuracy, utilizing Equation (\ref{eq:73}) for this purpose can degrade the solution accuracy if the modeling is imprecise. Therefore, by integrating the approach outlined in Section \ref{sec:3.1}, Equation (\ref{eq:42}) is employed to eliminate the errors introduced by the uncertainty components of the source data sets.

For the source data sets \{\(\boldsymbol{A_{i}}\)\}, the corresponding \(SE(3)\) means, as shown in Equation (\ref{eq:15}), can be computed iteratively for each respective set as
\begin{equation}
	\label{eq:43}
	{ }^{k+1} \boldsymbol{M_{A}}={ }^{k} \boldsymbol{M_{A}} \circ \textit{exp} \left[\frac{1}{N} \sum_{i=1}^{N} \log \left({ }^{k} \boldsymbol{M_{A_{i}}}^{-1}  \boldsymbol{A}\right)\right]
\end{equation}
where \( k \) represents the iteration count, and an initial estimate for the iterative procedure can be chosen as
\begin{equation}
	\label{eq:44}
	{ }^{1} \boldsymbol{M_{A}}=\textit{exp} \left(\frac{1}{N} \sum_{i=1}^{N} \log \left(\boldsymbol{A_{i}}\right)\right).
\end{equation}

Using Equations (\ref{eq:43}) and (\(\ref{eq:44}\)), the problem is formulated as
\begin{equation}
	\label{eq:68}
	\mathcal{M}_{\text {opt }} =\frac{1}{N} \sum_{i=1}^{n} \textit{log} \left(\textit{exp} \left([-\delta \boldsymbol{\zeta}_{M}]^{\wedge}\right) { }^{1}\boldsymbol{M}_{\boldsymbol{A}}^{-1} \boldsymbol{A}_{i}\right)
\end{equation}

Identifying the optimal correction parameter \( \delta \boldsymbol{\zeta}_{M} \) that minimizes \( \mathcal{M}_{\text {opt }} \), the optimal value of \( \boldsymbol{M_{A}} \) is updated as
\begin{equation}
	\label{eq:69}
	{ }^{k+1} \boldsymbol{M_{A}}={ }^{k} \boldsymbol{M_{A}}\circ\textit{exp} \left([\delta \boldsymbol{\zeta}_{M}]^{\wedge}\right)
\end{equation}

It is noteworthy that Equations (\ref{eq:40}) and (\ref{eq:69}) have different forms due to their distinct iterative constructions. Using \(\boldsymbol{\zeta}_{i}^{\prime}\) to represent \( { }^{1}\boldsymbol{M}_{\boldsymbol{A}}^{-1} \boldsymbol{A}_{i}\), and combining Equation (\(\ref{eq:13}\)), the optimization equation can be transformed through the manipulation of \([\cdot]^{\wedge}\) as
\begin{equation}
	\label{eq:70}
	\frac{1}{N} \sum_{i=1}^{N}\left(-\mathcal{J}_{l}^{-1}\left(\boldsymbol{\zeta}_{i}^{\prime}\right)  \delta \boldsymbol{\zeta}_{M}+\boldsymbol{\zeta}_{i}^{\prime}\right)=0,
\end{equation}
where \(\delta \boldsymbol{\zeta}_{M}\) can be solved using Equation (\ref{eq:69}), and \(\boldsymbol{M_{A}}\) is updated iteratively until either the maximum number of iterations is reached or the norm of \(\delta \boldsymbol{\zeta}_{M}\) falls below a predefined threshold. Finally, the optimal value of \(\boldsymbol{M_{A}}^{*}\), which can be regarded as the mean \(\hat{\boldsymbol{M_{A}}}\) of the source data sets \( \{\boldsymbol{A}_{i}\} \), is obtained.

The next step, the variance \( \hat{\boldsymbol{\Sigma}}_{\boldsymbol{A}} \) can be obtained from the calculated \( \hat{\boldsymbol{M}}_{\boldsymbol{A}} \) using Equation (\ref{eq:16}), then the decentralized formula is defined as
\begin{equation}
	\label{eq:45}
	{{\hat{\boldsymbol{\psi}}}}_{\boldsymbol{A}}=(\hat{\boldsymbol{\Sigma}}_{\boldsymbol{A}})^{-\frac{1}{2}} \circ \textit{log} ({\hat{\boldsymbol{M}}_{\boldsymbol{A}}}^{-1}\boldsymbol{A})^{\vee}.
\end{equation}

The same approach applies to \{\(\boldsymbol{B_{i}}\)\}. By applying Equation (\ref{eq:45}), the source data sets \( \{\boldsymbol{A}_{i}\} \) and \( \{\boldsymbol{B}_{i}\} \) are each transformed into \( \{\boldsymbol{\psi}_{\boldsymbol{A}_{i}}\} \) and \( \{\boldsymbol{\psi}_{\boldsymbol{B}_{i}}\} \), respectively, with their respective means regarded as unit vectors. The means and variances of the source data \( \{\boldsymbol{A}_{i}\} \) and \( \{\boldsymbol{B}_{i}\} \) satisfy the following conditions as
\begin{equation}
	\label{eq:46}	
  \left\{
		\begin{array}{l}
			{\boldsymbol{M}}_{\boldsymbol{A}}^{-1}\boldsymbol{A} =\mathbf{A d}\left(\vectXX\right){\boldsymbol{M}}_{\boldsymbol{B}}^{-1}\boldsymbol{B}   \\
			\hat{\boldsymbol{\Sigma}}_{\boldsymbol{A}}= \mathbf{A d}\left(\vectXX\right)\hat{\boldsymbol{\Sigma}}_{\boldsymbol{B}}\mathbf{A d}\left(\vectXX\right)^{T}.
		\end{array}
		\right.
\end{equation}

By leveraging the relationship in Equation (\ref{eq:46}), it can be observed that for the elements constructed in Equation (\ref{eq:45}), for each corresponding pair in the source data sets \((\boldsymbol{A_{i}},\boldsymbol{B_{i}})\), if uncertainties do not exist, the following conditions must be satisfied as
\begin{equation}
	\label{eq:47}
	\lVert {{{\boldsymbol{\psi}}}}_{\boldsymbol{A}} \rVert_2 = \lVert {{{\boldsymbol{\psi}}}}_{\boldsymbol{B}} \rVert_2.
\end{equation}

Conversely, Equation (\ref{eq:47}) does not hold, which indicates the presence of uncertainty factors. A possibility being investigated is utilizing the information provided by the source data, in order to elucidate the uncertainty information to the fullest possible extent. At the factual level, the inequality in Equation (\ref{eq:47}) provides a metric for quantifying uncertainty between the source data sets \( \{\boldsymbol{A}_{i}\} \) and \( \{\boldsymbol{B}_{i}\} \), which can be defined as an invariant in the \(\vectAA\)\(\vectXX\)=\(\vectYY\)\(\vectBB\) problem.

Under ideal conditions,  \( \textit{det}({\boldsymbol{\Sigma}}_{\boldsymbol{A}}) = \textit{det} ({\boldsymbol{\Sigma}}_{\boldsymbol{B}}) \) signifies that there is no uncertainty among the data, and measurements are entirely free of noise. Unfortunately, this does not occur in real-world scenarios. The uncertainty factors in the data are defined by two components. 

The first component, for each data pair \((\boldsymbol{A}_{i}, \boldsymbol{B}_{i})\), unique discrepancies arise due to the robot's absolute positioning precision errors and the camera's intrinsic measurement deviations, such as, the robot's limited localization accuracy, which leads to a deviation in the relative center position of the robot's end pose in the real world is different from the deviation in the ideal world,leading to variations specific to each individual set. The other component is caused by measurement noise, which is a random variable and cannot be modeled through a single element; instead, it must be modeled across all elements.

Considering the error metric established in Equation (\ref{eq:23}), the impact of the uncertainty error \(\delta\boldsymbol{e}\) shown in Equation (\ref{eq:72}) between the dataset \( \{\boldsymbol{A}_{i}\} \) and the dataset \( \{\boldsymbol{B}_{i}\} \) is defined as
\begin{equation}
	\label{eq:52}
	\boldsymbol{\delta \vecte_{i}} = \mathcal{J}_{l}(\boldsymbol{{A}_{i}})\delta\boldsymbol{\zeta}_{i},
\end{equation}
where, the derivation of Equation (\ref{eq:52}) can be obtained through the transformations from  Equations (\ref{eq:27}) to (\ref{eq:29}).
The uncertainty correction term \(\boldsymbol\delta\boldsymbol{\zeta}_{i}\) can be calculated as
\begin{equation}
	\label{eq:51}
	\left\{
		\begin{array}{l}
			\boldsymbol{\delta \zeta}_{1 \times 3} =\omega {\boldsymbol{M}_{\boldsymbol{\psi}_{\boldsymbol{B}}}}_{1 \times 3} \odot  \boldsymbol{ \chi}_{1 \times 3}  \\
			\boldsymbol{\delta \zeta}_{4 \times 6} = \omega {\boldsymbol{M}_{\boldsymbol{\psi}_{\boldsymbol{B}}}}_{4 \times 6} \odot\boldsymbol{ \chi}_{4 \times 6},
		\end{array}
		\right.
\end{equation}
where \( \omega \) is the standard deviation obtained through variance calculation \({\boldsymbol{\Sigma}}_{\boldsymbol{A}}^{\frac{1}{2}}\), and \(\boldsymbol{ \chi}\) is a ratio, which is defined as
\begin{equation}
	\label{eq:48}
	\begin{aligned}
	\boldsymbol{\chi_i} =&(1-\lambda) (\frac{\left\|\boldsymbol{\psi}_{\boldsymbol{B_i}}\right\|}{\left\|\boldsymbol{\psi}_{\boldsymbol{A_i}}\right\|}-1) \cdot \frac{\boldsymbol{\psi}_{\boldsymbol{A_i}}}{\left\|\boldsymbol{\psi}_{\boldsymbol{A_i}}\right\|} \\&+\lambda \cdot \textit{diag}(\frac{\left\|{\boldsymbol{\Sigma}}_{\mathbf{B}}\right\|}{\left\|{\boldsymbol{\Sigma}}_{\mathbf{A}}\right\|}-1) \cdot \left(\frac{{\boldsymbol{\Sigma}}_{\mathbf{A}}}{\left\|{\boldsymbol{\Sigma}}_{\mathbf{A}}\right\|}\right),
	\end{aligned}
\end{equation}
where \( \lambda \) is the influence factor, specifically the proportion of the variance's contribution to the overall metric. The mean and variance can be defined as
\begin{equation}
	\label{eq:49}
	\left\{
		\begin{array}{l}
			\boldsymbol{M_{\psi}}=\frac{1}{N}\sum_{i=1}^{N}\boldsymbol{\psi}_{i} \\
			\boldsymbol{\Sigma}_{\boldsymbol{\psi}}=\frac{1}{N}\sum_{i=1}^{N}\mathbb{E}\left[\left(\boldsymbol{\psi}_{\boldsymbol{i}}-\boldsymbol{I}_{1 \times 6}\right)\left(\boldsymbol{\psi}_{\boldsymbol{i}}-\boldsymbol{I}_{1 \times 6}\right)^{T}\right].
		\end{array}
	\right.
\end{equation}

It is noteworthy that \( N \) is employed here instead of \( N-1 \) because the mean is assumed to be the known value \( \boldsymbol{I}_{1 \times 6} \). According to Equation (\ref{eq:49}), \( \lambda \) is defined as
\begin{equation}
	\label{eq:50}
	\lambda=log\left(1+\frac{diag({\boldsymbol{\Sigma}}_{\mathbf{A}})}{diag({\boldsymbol{\Sigma}}_{\mathbf{B}})}\right)log\left(1+\frac{diag({\boldsymbol{\Sigma}_{\boldsymbol{\psi}_{\boldsymbol{A}}}})}{diag({\boldsymbol{\Sigma}_{\boldsymbol{\psi}_{\boldsymbol{B}}}})}\right)^{-1}.
\end{equation}

By utilizing Equations (\ref{eq:48})-(\(\ref{eq:50}\)), \(\boldsymbol{\delta \xi}\) in Equation (\ref{eq:52}) can be computed as Equation (\ref{eq:51}). By correcting Equation (\ref{eq:37}) through the aforementioned uncertainty modeling as
\begin{equation}
	\label{eq:71}
	\boldsymbol{\gamma} = [\textit{log} (\boldsymbol{e})]^{\vee}-\mathcal{J}_{l}(\boldsymbol{{A}_{i}})\delta\boldsymbol{\zeta}_{i}-(\boldsymbol{J}_{X}\delta\boldsymbol{\zeta_X}+\boldsymbol{J}_{Y}\delta\boldsymbol{\zeta_Y}). 
\end{equation}

On the whole, the complete process of the method for uncertain modeling is presented in Algorithm \ref{alg:2}.
\begin{algorithm}[!htbp]
    \caption{\small{\textbf{The proposed method for uncertainty modeling of the source dataset}}}
    \label{alg:2}
    \begin{algorithmic}
    \State \textbf{Input:} 
	\State The set $\{\boldsymbol{A_{i}}\}$ ; // \small{The source dataset from the controller indicates robot's pose in its base frame.}
    \State The set $\{\boldsymbol{B_{i}}\}$ ; // \small{The source dataset from the measurement equipment indicates robot's pose in measurement frame}
	\State $\epsilon$; // \small{Iterative parameter}
    \State \textbf{Output:} 
	\State $\{\boldsymbol{\delta \vecte_{i}}\}$ ; // \small{Characterizing the uncertainties in the source data, which are used to correct the deviation metric \(\boldsymbol{\gamma}\) used Equation (\ref{eq:71}) instead of Equation (\(\ref{eq:34}\))}
    \end{algorithmic}
    \begin{algorithmic}[1] % Restart algorithmic with numbering
    \State Calculate the mean values ${ }^{0} \boldsymbol{M_{A}}, { }^{0} \boldsymbol{M_{B}}$ based on Equation (\ref{eq:44}) 
    \State Initialization: MaxIters, iter$\leftarrow$1,
	\State \textbf{while} iter $<$ MaxIters \textbf{do}
    \State Construct $\mathcal{M}_{\text {opt}}^{\boldsymbol{A}}$ and $\mathcal{M}_{\text {opt}}^{\boldsymbol{B}}$ based on Equation (\ref{eq:68})
	\State Solve \(\delta \boldsymbol{\zeta}_{M}^{\boldsymbol{A}}\) and \(\delta \boldsymbol{\zeta}_{M}^{\boldsymbol{B}}\) based on Equation (\ref{eq:69})
    \State Update the optimal solution \(\boldsymbol{M_{A}},  \boldsymbol{M_{B}}\) based on Equation (\ref{eq:69})
	\State iter $\leftarrow$ iter + 1
	\State \textbf{until} $\delta \boldsymbol{\zeta}_{M}^{\boldsymbol{A}} < \epsilon$ and $\delta \boldsymbol{\zeta}_{M}^{\boldsymbol{B}} < \epsilon$
	\State \textbf{end while}
    \State Compute the final mean values $\hat{\boldsymbol{M_{A}}}$ and $\hat{\boldsymbol{M_{B}}}$
	\State Calculate the variance $\hat{\boldsymbol{\Sigma}}_{\boldsymbol{A}}$ and $\hat{\boldsymbol{\Sigma}}_{\boldsymbol{B}}$ based on Equation (\ref{eq:16})
	\State Calculate $ \{\boldsymbol{\psi}_{\boldsymbol{A}_{i}}\} $ and $ \{\boldsymbol{\psi}_{\boldsymbol{B}_{i}}\} $ based on Equation (\ref{eq:45})
	\State Construct $\{\boldsymbol{\chi_i}\}$ based on Equations (\ref{eq:49})-(\ref{eq:50}) and compute based on Equation (\ref{eq:48})
	\State Calculate $\{\boldsymbol{\delta \vecte_{i}}\}$ based on Equations (\ref{eq:51})-(\ref{eq:52})
    \end{algorithmic}
\end{algorithm}

\subsection{4.3. Initial calibration solver}\label{sec:3.3}

Although the final iterative formulation in Section 4.1 constitutes a convex function, rendering the optimal solution theoretically independent of the initial values, a well-chosen initial estimate can significantly enhance computational efficiency. In addition, the calibration model construction employs a linear approximation that capitalizes on the properties of the Euclidean group in the vicinity of the identity element. Thus, a positive initial calibration solution is paramount for the precise resolution of the calibration model.

For the obtained the source dataset \{(\(\boldsymbol{A_{i}},\boldsymbol{B_{i}}\))\}, before proceeding with the initial solution, generally, one can first apply a filtering process based on Equation (\ref{eq:20}) to check the impact of errors in the data. By doing so, it becomes possible to screen out data pairs characterized by erroneous data.

Based on screw theory, an element \(\boldsymbol{T} \in SE(3)\) can be represented by four screw parameters (\(\theta,h,\boldsymbol{k},\boldsymbol{c}\)) as
\begin{equation}
	\label{eq:53}
	\boldsymbol{T}=\left(\begin{array}{cc}e^{\theta [\boldsymbol{k}]^{\wedge}} & \left(I_{3}-e^{\theta [\boldsymbol{k}]^{\wedge}}\right) \boldsymbol{c}+h \theta [\boldsymbol{k}]^{\wedge} \\0^{T} & 1\end{array}\right),
\end{equation}
where \(\theta\) and \(h\) have already been introduced in Equation (\ref{eq:20}), and \(\boldsymbol{k}\) denotes the rotation axis and \(\boldsymbol{c}\) is the position of a point on the line relative to the origin of a space-fixed reference frame with \(\boldsymbol{c} \cdot \boldsymbol{k} =\boldsymbol{0} \). By combining Equation (\ref{eq:9}), the matrix \(\boldsymbol{T}\) can be expressed in a form analogous to the adjoint representation of the Lie algebra as
\begin{equation}
	\label{eq:54}
	\mathbf{a d}({[\boldsymbol{\zeta_t}]^{\wedge}})=\theta\left[\begin{array}{cc}{[\boldsymbol{k}]^{\wedge}} & \mathbf{O} \\{[\boldsymbol{c} \times \boldsymbol{k}+h \boldsymbol{k}]^{\wedge}} & {[\boldsymbol{k}]^{\wedge}}\end{array}\right]\in \operatorname{ad}(\operatorname{se}(3)).
\end{equation}

For the \(\vectAA\)\(\vectXX\)=\(\vectYY\)\(\vectBB\)  problem, by applying Equation (\ref{eq:18}), it can be reformulated as
\begin{equation}
	\label{eq:55}
	\vectAA_{ij}\vectXX =\vectXX\vectBB_{ij}.
\end{equation}

Further, leveraging Equation (\ref{eq:54}) and the properties of the adjoint matrix, Equation (\ref{eq:55}) can be rewritten as
\begin{equation}
	\label{eq:56}
	\mathbf{a d}({[\boldsymbol{\zeta_{A_{ij}}}]^{\wedge}})\mathbf{Ad}(\vectXX)=\mathbf{Ad}(\vectXX)\mathbf{a d}({[\boldsymbol{\zeta_{B_{ij}}}]^{\wedge}}).
\end{equation}

Based on Equations (\ref{eq:10}) and (\(\ref{eq:11}\)), let \(\mathbf{A}\) substitute for \(\boldsymbol{A_{ij}}\), \(\mathbf{B}\) substitute for \(\boldsymbol{B_{ij}}\), Equation (\ref{eq:56}) transforms as
\begin{equation}
	\label{eq:57}
	\begin{aligned}\left\{\begin{array}{c}\theta_{\mathrm{A}} \boldsymbol{k}_{\mathrm{A}} \\\theta_{\mathrm{A}} \boldsymbol{c}_{\mathrm{A}} \times \boldsymbol{k}_{\mathrm{A}}\end{array}\right\} & =\left[\begin{array}{cc}\boldsymbol{R}_{\mathbf{X}} & 0 \\{\left[\boldsymbol{t}_{\mathbf{X}}\right]^{\wedge} \boldsymbol{R}_{\mathrm{X}}} & \boldsymbol{R}_{\mathrm{X}}\end{array}\right]\left\{\begin{array}{c}\theta_{\mathrm{B}} \boldsymbol{k}_{\mathrm{B}} \\\theta_{\mathrm{B}} \boldsymbol{c}_{\mathrm{B}} \times \boldsymbol{k}_{\mathrm{B}}\end{array}\right\} \\\left\{\begin{array}{c}\mathbf{0} \\\theta_{\mathrm{A}} \boldsymbol{h}_{\mathrm{A}} \boldsymbol{k}_{\mathrm{A}}\end{array}\right\} & =\left[\begin{array}{cc}\boldsymbol{R}_{\mathrm{X}} & \mathbf{0} \\{\left[\mathbf{t}_{\mathbf{X}}\right]^{\wedge} \boldsymbol{R}_{\mathbf{X}}} & \boldsymbol{R}_{\mathbf{X}}\end{array}\right]\left\{\begin{array}{c}\boldsymbol{0} \\\theta_{\mathbf{B}} \boldsymbol{h}_{\mathrm{B}} \boldsymbol{k}_{\mathrm{B}}\end{array}\right\}\end{aligned}.
\end{equation}

Assuming there are \(N\) groups of data, the rotational part of initial calibration solution can be obtained by solving the following optimization problem as
\begin{equation}
	\label{eq:58}
	F\left(\boldsymbol{R}_{\mathbf{X}}\right)=\sum_{n=1}^{N}\left(\boldsymbol{R}_{\mathbf{X}} \cdot \theta_{\mathbf{B}_{n}} \boldsymbol{k}_{\mathbf{B}_{n}}-\theta_{\mathbf{A}_{n}} \boldsymbol{k}_{\mathbf{A}_{n}}\right)^{2}.
\end{equation}

Based on Equation (\ref{eq:58}), the optimal solution \(\boldsymbol{R}_{\vectXX}\) needs to satisfy the following equation as
\begin{equation}
	\label{eq:59}
	\boldsymbol{R}_{\vectXX}=\arg \max \left(\sum_{n=1}^{N}\left(\theta_{\mathbf{A}_{n}} \boldsymbol{k}_{\mathbf{A}_{n}}\right)^{\mathrm{T}} \boldsymbol{R}_{\mathbf{X}}\left(\theta_{\mathbf{B}_{n}} \boldsymbol{k}_{\mathbf{B}_{n}}\right)\right),
\end{equation}
where using the properties of the trace, the rotation matrix \(\boldsymbol{R}_{\vectXX}\) can be obtained by decomposing \(\sum_{n=1}^{N}\left(\left(\theta_{\mathbf{A}_{n}} \boldsymbol{k}_{\mathbf{A}_{n}}\right)^{T}\left(\theta_{\mathbf{B}_{n}} \boldsymbol{k}_{\mathbf{B}_{n}}\right)\right)\) using singular value decomposition (SVD) as
\begin{equation}
	\label{eq:60}
	\boldsymbol{R}_{\vectXX}=\boldsymbol{V}\left[\begin{array}{ccc}\mathbf{1} & \mathbf{0} & \mathbf{0} \\\mathbf{0} & \mathbf{1} & \mathbf{0} \\\mathbf{0} & \mathbf{0} & \operatorname{det}\left(\boldsymbol{V} \boldsymbol{U}^{\mathrm{T}}\right)\end{array}\right] \boldsymbol{U}^{\mathrm{T}},
\end{equation}
where \(\boldsymbol{U}\) and \(\boldsymbol{V}\) are orthogonal matrices by SVD. Using the iterative procedure outlined in Equations (\(\ref{eq:44}\))-(\(\ref{eq:70}\)), the means of the source data sets \(\{\mathbf{A_{n}}\}\) and \(\{\mathbf{B_{n}}\}\) can be obtained, the translational part of \(\vectXX\) can be calculated as
\begin{equation}
	\label{eq:61}
	\boldsymbol{t}_{\vectXX}=\boldsymbol{R}_{\boldsymbol{M}_\mathrm{A}} \boldsymbol{t}_{\vectXX}+\boldsymbol{t}_{\boldsymbol{M}_\mathrm{A}}-\boldsymbol{R}_{\vectXX} \boldsymbol{t}_{\boldsymbol{M}_\mathrm{B}}.
\end{equation}

Since solving the rotational and translational components separately can lead to error propagation, it is necessary to eliminate the errors. Considering the impact of dimensionality, the iterative errors for rotation and translation are constructed separately as
\begin{equation}
	\label{eq:62}
	E_{n}(\boldsymbol{R}_\vectXX, \boldsymbol{t}_\vectXX)=\left[\begin{array}{c}\boldsymbol{R}_\vectXX \left(\theta_{\mathbf{B}_{n}} \boldsymbol{k}_{\mathbf{B}_{n}}\right)-\left(\theta_{\mathbf{A}_{n}} \boldsymbol{k}_{\mathbf{A}_{n}}\right) \\ \boldsymbol{R}_{\mathbf{A}_{n}}  \boldsymbol{t}_\vectXX+ \boldsymbol{t}_{\mathbf{A}_{n}}-\boldsymbol{R}_\vectXX \boldsymbol{t}_{\mathbf{B}_{n}}-\boldsymbol{t}_\vectXX\end{array}\right].
\end{equation}

The objective function \(\mathcal{F}(\boldsymbol{R}_\vectXX, \boldsymbol{t}_\vectXX)\) can be defined as
\begin{equation}
	\label{eq:63}
	\begin{aligned}
	\underset{\boldsymbol{R}_{X},\boldsymbol{t}_{X}}{\arg \min }\mathcal{F}&(\boldsymbol{R}_\vectXX, \boldsymbol{t}_\vectXX)=\\&\frac{1}{2}\sum_{n=1}^{N}E_{n}(\boldsymbol{R}_\vectXX, \boldsymbol{t}_\vectXX)^{\top} E_{n}(\boldsymbol{R}_\vectXX, \boldsymbol{t}_\vectXX).
	\end{aligned}
\end{equation}

Record the incremental parameter as \(\delta \boldsymbol{x}=[\delta \boldsymbol{\omega} ; \delta \boldsymbol{t}] \in \mathbb{R}^{6}\), for every pair \(\{\mathbf{A}_n, \mathbf{B}_n\}\) , the Jacobian matrix \(\boldsymbol{J}_n\) of \(E_n(\mathbf{R}_X, \mathbf{t}_X)\) can be computed. Constructing the total Jacobian matrix \(\boldsymbol{J}\in \mathbb{R}^{6 N \times 6}\) from all Jacobian matrices yields and \(E_{N}(\boldsymbol{R}_\vectXX, \boldsymbol{t}_\vectXX) \in \mathbb{R}^{6 N}\) the iterative equation as
\begin{equation}
	\label{eq:64}
	\left(\boldsymbol{J}^{\top} \boldsymbol{J}+\lambda \boldsymbol{I}\right) \delta \boldsymbol{x}=-\boldsymbol{J}^{\top}E_{N}(\boldsymbol{R}_\vectXX, \boldsymbol{t}_\vectXX),
\end{equation}
where \(\lambda\) is a damping parameter for the L-M (Levenberg-Marquardt) method, and the solution \(\delta \boldsymbol{x}\) is obtained iteratively as
\begin{equation}
	\label{eq:65}
	\delta \boldsymbol{x}=-\left(\boldsymbol{J}^{\top} \boldsymbol{J}+\lambda \boldsymbol{I}\right)^{-1} \boldsymbol{J}^{\top} E_{N}(\boldsymbol{R}_\vectXX, \boldsymbol{t}_\vectXX).
\end{equation}

Further \(\boldsymbol{R}_\vectXX, \boldsymbol{t}_\vectXX\) can be updated as
\begin{equation}
	\label{eq:66}
	\left\{
		\begin{array}{l}
			\boldsymbol{R}_\vectXX=\boldsymbol{R}_\vectXX\textit{exp}([\delta \boldsymbol{\omega}]^{\wedge}) \\
			\boldsymbol{t}_\vectXX=\boldsymbol{t}_\vectXX+\delta \boldsymbol{t}
		\end{array}.
		\right.
\end{equation}

As the iterations continue, adjusting \(\lambda\) by factor \(\mu\) as
\begin{equation}
	\label{eq:67}
		\lambda^{(k+1)}=\left\{\begin{array}{ll}\max \left(\frac{\lambda^{(k)}}{\mu}, \lambda_{\min }\right),\text { if } \mathcal{F}^{(k+1)}<\mathcal{F}^{(k)} \\\min \left(\mu \times \lambda^{(k)}, \lambda_{\max }\right),\text { else }\end{array}\right.,
\end{equation}
where \(\left|\mathcal{F}^{(k)} - \mathcal{F}^{(k+1)}\right| < \epsilon\), the error update of \(\mathcal{F}\) is deemed to be below the prescribed minimum threshold \(\operatorname{tol}\), and the optimal initial matrix \(\hat{\vectXX}^{1}\) is finally obtained. Thereafter, the initial matrix \(\hat{\vectYY}^{1}\) can be calculated through Equation (\ref{eq:17}).

In summary, the complete process of the method for initial calibration \(\hat{\vectXX}^{1}\) and \(\hat{\vectYY}^{1}\) is presented in Algorithm~\ref{alg:3}.
\begin{algorithm}[!htbp]
    \caption{\small{\textbf{Initial calibration solving method based on the L-M iterative algorithm}}}
    \label{alg:3}
    \begin{algorithmic}
    \State \textbf{Input:} 
	\State The set $\{\boldsymbol{A_{i}}\}$ ; // \small{The source dataset from the controller indicates robot's pose in its base frame.}
    \State The set $\{\boldsymbol{B_{i}}\}$ ; // \small{The source dataset from the measurement equipment indicates robot's pose in measurement frame}
	\State $\lambda, \lambda_{\max }, \lambda_{\min },  \epsilon, \mu$ ; // \small{Iterative parameter}
    \State \textbf{Output:} 
	\State  \(\hat{\vectXX}^{1}\) and \(\hat{\vectYY}^{1}\) ; // \small{The initial solution for the  \(\vectAA\)\(\vectXX\)=\(\vectYY\)\(\vectBB\) problem}
    \end{algorithmic}
    \begin{algorithmic}[1] % Restart algorithmic with numbering
	\State Compute $\{\vectAA_{ij}\}$ and $\{\vectBB_{ij}\}$ based on Equation (\ref{eq:19}) and defined \(\mathbf{A}\) and \(\mathbf{B}\)
	\State Compute the means of the source data sets \(\{\mathbf{A_{n}}\}\) and \(\{\mathbf{B_{n}}\}\) based on Equations (\ref{eq:44})-(\ref{eq:70})
	\State Compute four screw parameters (\(\theta,h,\boldsymbol{k},\boldsymbol{c}\)) of $\{\mathbf{A}_i\}$ and $\{\mathbf{B}_i\}$ based on Equation (\ref{eq:53})
	\State Compute the rotational part of the initial calibration solution $\boldsymbol{R}_{\vectXX}$ based on Equations (\ref{eq:57})-(\ref{eq:60})
	\State Compute the translational part of the initial calibration solution $\boldsymbol{t}_{\vectXX}$ based on Equation~(\ref{eq:61})
    \State Initialization: MaxIters, iter$\leftarrow$1, 
	\State \textbf{while} iter $<$ MaxIters \textbf{do}
    \State Compute Jacobian matrices \(\boldsymbol{J}\) defined as Equations (\ref{eq:62}) and (\ref{eq:63})
	\State Compute \(\delta \boldsymbol{x}\) based on Equations (\ref{eq:64}) and (\ref{eq:65})
    \State Update the optimal solution \(\boldsymbol{R}_{\vectXX}\) and \(\boldsymbol{t}_{\vectXX}\) based on Equation (\ref{eq:66})
	\State Update the damping parameter \(\lambda\) based on Equation (\ref{eq:67})
	\State iter $\leftarrow$ iter + 1
	\State \textbf{until} \(\left|\mathcal{F}^{(k)} - \mathcal{F}^{(k+1)}\right| < \epsilon\)
	\State \textbf{end while}
	\State Compute the \(\hat{\vectYY}^{1}\) based on Equation (\ref{eq:17})
    \State Compute the final calibration results \(\hat{\vectXX}^{1}\) and \(\hat{\vectYY}^{1}\)
    \end{algorithmic}
\end{algorithm}

\section{5. Evaluations}
\label{sec:4}
\subsection{5.1. Numerical simulations}\label{sec:4.1}
\subsubsection{5.1.1.  Simulation data statement}

In the simulations, synthetic data were generated to emulate realistic HEC scenarios. The base dataset comprises 6-DoF robot poses represented as position and orientation (Euler-angles) and 6-DoF camera observation data. To better simulate real-world conditions, the synthesized data assumes error-free six-dimensional data from the robot, while the camera observation data contains errors, which is intended to reflect practical applications, where dataset \{\({\boldsymbol{A_{i}}}\)\} is typically obtained from the teach pendant and represents ideal computed values. Camera errors are relatively complex. In real-world scenarios, the camera observations are obtained by starting from the ideal robot pose, adding the robot's AU and EU as 
\begin{equation}
	\label{eq:101}
   \left\{
		\begin{array}{l}
\bm{\mathit{p}}_{\text{noise}}^{(i)} = \bm{\mathit{p}}_{opt}^{(i)} + \bm{\mathit{e}}^{(i)}_{\text{EU}} + \bm{\mathit{e}}_{\text{AU}}^{(i)},\\
\bm{\mathit{\omega}}_{\text{noise}}^{(i)} = \bm{\mathit{\omega}}_{opt}^{(i)} + \bm{\mathit{\theta}}^{(i)}_{\text{EU}} + \bm{\mathit{\theta}}_{\text{AU}}^{(i)},
		\end{array}
		\right.
\end{equation}
where $\bm{\mathit{p}}_{opt} \in \mathbb{R}^{3}$ denotes the synthesized robot position, and $\bm{\mathit{\omega}}_{opt} \in \mathbb{R}^{3}$ denotes the synthesized robot orientation. $\bm{\mathit{e}}_{\text{AU}}^{(i)}$ and $\bm{\mathit{\theta}}_{\text{AU}}^{(i)}$ represents random error, characterized by the following as
\begin{equation}
	\label{eq:102}
   \left\{
		\begin{array}{l}
\bm{\mathit{e}}_{\text{AU}}^{(i)} \sim \mathcal{N}\left(0_{3 \times 1}, \Sigma_{p}\right),\\
\bm{\mathit{\theta}}_{\text{AU}}^{(i)} \sim \mathcal{N}\left(0_{3 \times 1}, \Sigma_{w}\right),
		\end{array}
		\right.
\end{equation}
where $\mathcal{N}$ denotes a Gaussian distribution. Additionally, $\bm{\mathit{e}}^{(i)}_{\text{EU}}$ and $ \bm{\mathit{\theta}}^{(i)}_{\text{EU}}$ represent repeatable errors as
\begin{equation}
	\label{eq:103}
   \left\{
		\begin{array}{l}
\bm{\mathit{e}}_{\text{EU}}^{(i)} \sim \mathcal{D}\left(\bm{\mathit{p}}_{opt}^{(i)}, \bm{\mathit{p}}_{opt}^{(o)}\right),\\
\bm{\mathit{\theta}}_{\text{EU}}^{(i)} \sim \mathcal{D}\left(\bm{\mathit{\omega}}_{opt}^{(i)}, \bm{\mathit{\omega}}_{opt}^{(o)}\right),
		\end{array}
		\right.
\end{equation}
where $\mathcal{D}$ denotes a distribution distance function. The parameters \(\Sigma_{p}\) and \(\Sigma_{w}\) are the covariance matrices of the random errors, while \(\bm{\mathit{p}}_{opt}^{(o)}\) and \(\bm{\mathit{\omega}}_{opt}^{(o)}\) are the position and orientation of the origin. $[\bm{\mathit{p}}_{\text{noise}}^{(i)}, \bm{\mathit{\omega}}_{\text{noise}}^{(i)}]$ obtained from Equation (\ref{eq:101}) are converted into HTMs, and transformed using the defined ideal HECPs $\boldsymbol{X}_{opt}$ and $\boldsymbol{Y}_{opt}$ into the camera coordinate frame. By converting the homogeneous form into Position-Euler angle form and applying the corresponding measurement errors according to the logic of Equation (\ref{eq:101}), the synthesized data \{\({\boldsymbol{B_{i}}}\)\} can be obtained.

\begin{table}[!htbp]
\centering
\begin{center}
\caption{\ \ \leftskip=0pt \rightskip=0pt plus 0cm Parameter range settings for injection of uncertainty at different levels into the synthesized data.}
\label{tab:noise_injection}
\footnotesize
\begin{tabular}{m{1.5cm}<{\centering} m{2.8cm}<{\centering} m{2.8cm}<{\centering}}
\toprule
\color{black}{Uncertainty} & \color{black}{Robot} & \color{black}{Camera} \\ \midrule
AUP    & $Par(\mathit{e}_{\text{AU}}^x,\mathit{e}_{\text{AU}}^y,\mathit{e}_{\text{AU}}^z)$         & $Par(\mathit{e}_{\text{AU}}^x,\mathit{e}_{\text{AU}}^y,\mathit{e}_{\text{AU}}^z)$         \\
       &  $\in (0, 1)$                                     &  $\in(0, 0.5)$                             \\
AUR & $\mathit{Par(\theta}_{\text{AU}}^x,\mathit{\theta}_{\text{AU}}^y,\mathit{\theta}_{\text{AU}}^z)$         &Par($\mathit{\theta}_{\text{AU}}^x,\mathit{\theta}_{\text{AU}}^y,\mathit{\theta}_{\text{AU}}^z)$         \\
       &  $\in(0, 0.4)$                                     & $\in(0, 0.2)$                             \\
EU   & $Par(\bm{\mathit{e}}_{\text{EU}}^{(i)},\bm{\mathit{\theta}}_{\text{EU}}^{(i)})$         & $Par(\bm{\mathit{e}}_{\text{EU}}^{(i)},\bm{\mathit{\theta}}_{\text{EU}}^{(i)} p_z)$         \\
       &  $\in(0, 0.004)$                                     &  $\in(0, 0.004)$                                 \\ \bottomrule
\end{tabular}
\end{center}
\end{table}

\subsubsection{5.1.2. Effectiveness of uncertainty metrics}

 Due to the complexity of constructing the synthesized data \{\({\boldsymbol{B_{i}}}\)\}, the uncertainties affecting its generation are divided into six components, each with its own corresponding numerical level which shown in Table \ref{tab:noise_injection}. Each of the six types of uncertainty is assigned 18 different levels, and their combinations are used to validation of the effectiveness of uncertainty metrics.
After forming 18 ordered combinations of the 6 types of noise, each combination is used to generate 100 data samples, and the uncertainty metric is then averaged for each combination.

\begin{figure}[!htbp]\centering
	\includegraphics[width=0.45\textwidth]{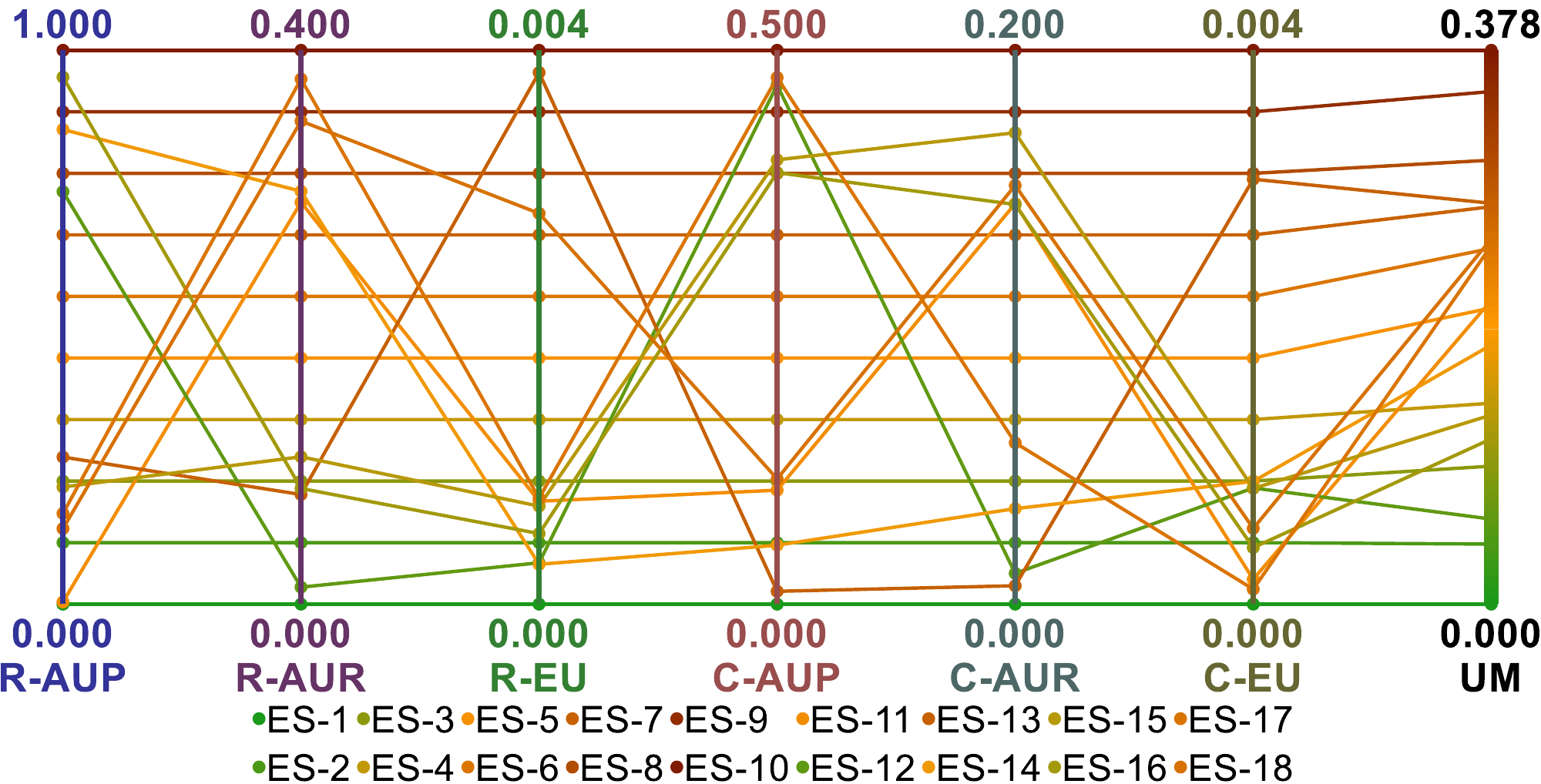}
	\caption{\ \ \leftskip=0pt \rightskip=0pt plus 0cm Relationship between different uncertainty combinations and uncertainty metrics.}
	\label{s1}
\end{figure}

As illustrated Figure \ref{s1}, for the 6 types of uncertainty across Error Scenarios-$1$ to Error Scenarios-$10$, their numerical magnitudes increase uniformly. Correspondingly, the uncertainty metrics exhibit an approximately proportional tendency, i.e as the magnitude of uncertainty increases, the values of the uncertainty metrics also increase, which indicates that the proposed uncertainty metric effectively captures the relative uncertainty relationship between  \{\({\boldsymbol{A_{i}}}\)\} and  \{\({\boldsymbol{B_{i}}}\)\}, that is, the greater the uncertainty in \{\({\boldsymbol{B_{i}}}\)\}, the larger the metric value. Another point worth noting is that across Error Scenarios 11 to 18, high AU in the rotational component has a greater impact on the uncertainty metric than high AU in the translational component. Additionally, to better simulate real-world conditions, data with higher EU values also leads to increased values in the uncertainty metric.
\begin{figure}[!htbp]\centering
	\includegraphics[width=0.45\textwidth]{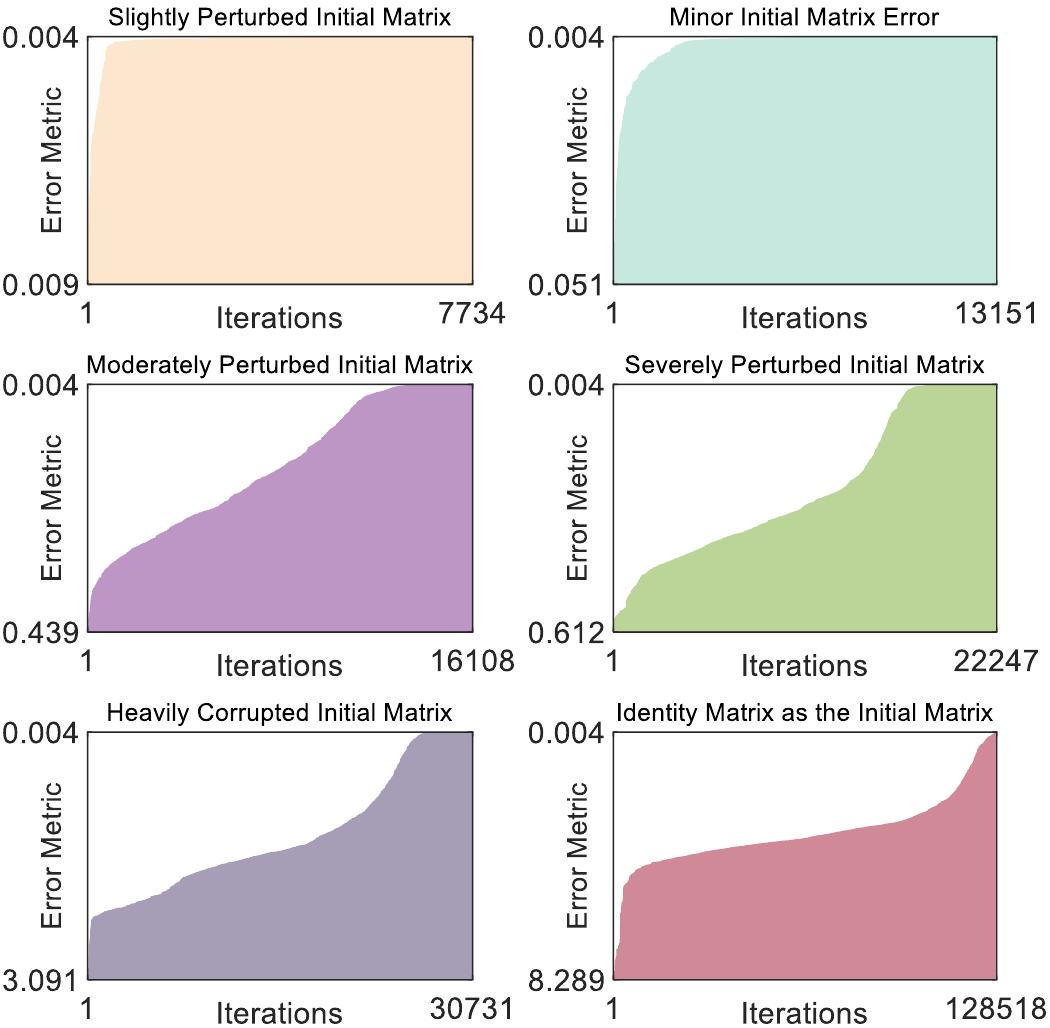}
	\caption{\ \ \leftskip=0pt \rightskip=0pt plus 0cm Relationship between heuristic error metrics and number of iterations under different initial values.}
	\label{s2}
\end{figure}

\subsubsection{5.1.3. Validation of global convergence}

 To verify the global convergence capability of the proposed L-HED method, iterations are performed using different initial values to compute the HECPs  $\boldsymbol{X}$ and $\boldsymbol{Y}$.
As illustrated Figure \ref{s2}, under different initial values, the heuristic error metrics all converge to the same level of accuracy. When the initial value is close to the final $\boldsymbol{X}_{opt}$ and $\boldsymbol{Y}_{opt}$, the algorithm reaches the designed stopping threshold in just 7,734 iterations, when the initial values are far from $\boldsymbol{X}_{opt}$ and $\boldsymbol{Y}_{opt}$, convergence is achieved after 30,731 iterations. In a more extreme case, when the \(\mathbf{I}_4\) is used as the initial value, convergence is still achieved; however, the number of iterations increases significantly, reaching 128518. It can also be observed that the closer the initial matrices are to $\boldsymbol{X}_{opt}$ and $\boldsymbol{Y}_{opt}$, the fewer iterations are required for convergence. However, it is worth noting that due to the involvement of heuristic escape mechanisms, the number of iterations under the same initial values is not fixed and may exhibit fluctuations.

\begin{figure*}[!htbp] % 
	\centering
	\includegraphics[width=\textwidth]{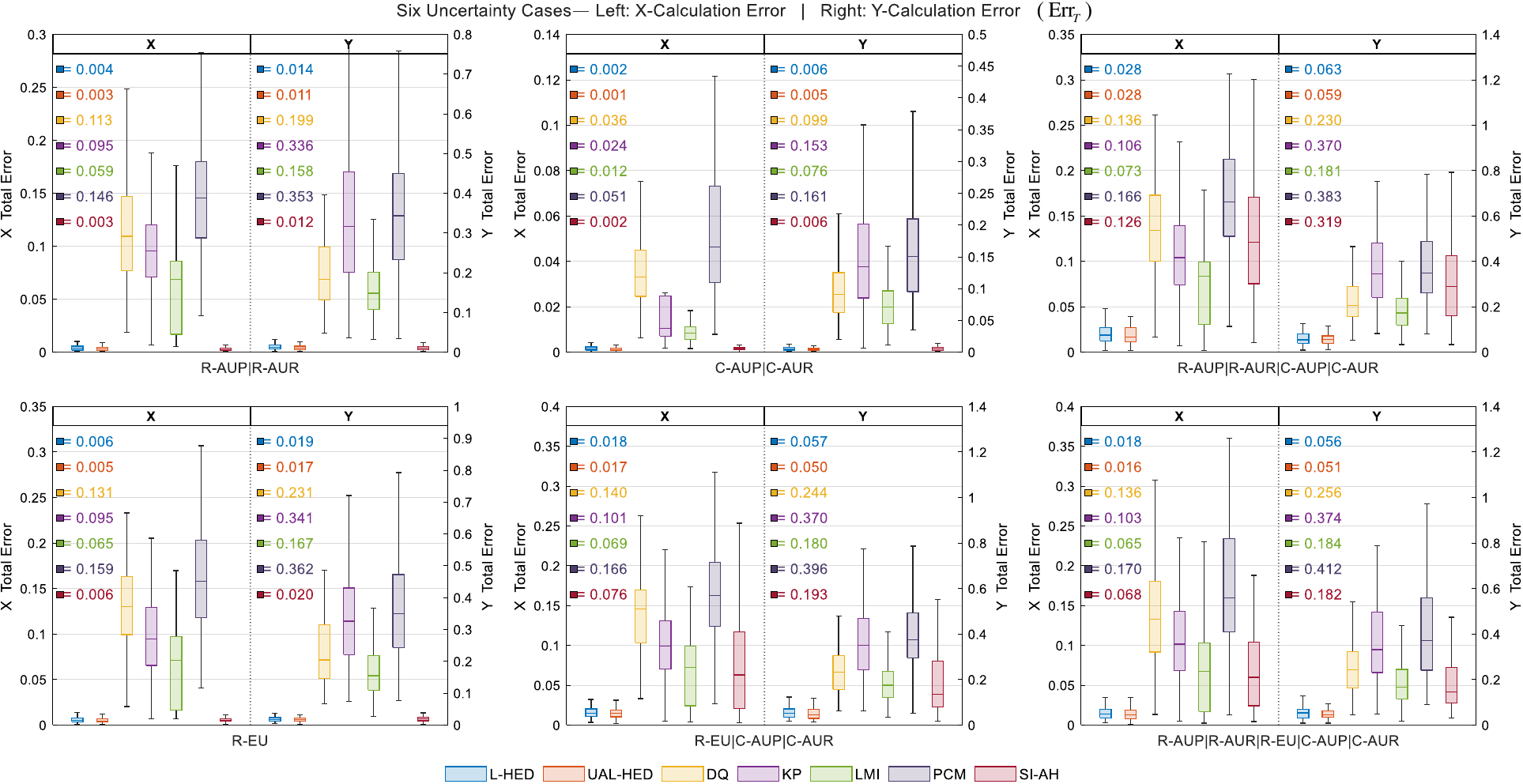} 
	\caption{\ \ \leftskip=0pt \rightskip=0pt plus 0cm Computation accuracy of seven methods under different uncertainty combinations.}
	\label{s3}
\end{figure*}
\begin{table}[!htbp]
\centering
\begin{center}
\caption{\ \ \leftskip=0pt \rightskip=0pt plus 0cm Six uncertainty combination settings}
\label{tab:s3}
\footnotesize
\begin{tabular}{m{2cm}<{\centering} m{2.5cm}<{\centering} m{2.5cm}<{\centering}}
\toprule
\color{black}{Uncertainty} & \color{black}{Robot} & \color{black}{Camera} \\ \midrule
R-AU    & $Par(\mathit{e}_{\text{AU}}~/~\mathit{\theta}_{\text{AU}})$         & $--$         \\
       &  $1~/~0.4$                                     &  $-- $                           \\
C-AU & $-- $         &Par($\mathit{e}_{\text{AU}}~/~\mathit{\theta}_{\text{AU}})$         \\
       &  $-- $                                     & $0.5~/~0.2$                             \\
R-AU/C-AU   & Par($\mathit{e}_{\text{AU}}~/~\mathit{\theta}_{\text{AU}})$         & $Par(\mathit{e}_{\text{AU}}~/~\mathit{\theta}_{\text{AU}})$         \\
       &  $1~/~0.4$                                     &   $0.5~/~0.2$                                 \\ 
R-EU   & $Par(\mathit{e}_{\text{EU}}~/~\mathit{\theta}_{\text{EU}})$         & $--$         \\
       &  $0.004$                                     &   $--$                                 \\ 
R-EU/C-AU   & $Par(\mathit{e}_{\text{EU}}~/~\mathit{\theta}_{\text{EU}})$         & $Par(\mathit{e}_{\text{AU}}~/~\mathit{\theta}_{\text{AU}})$         \\
       &  $0.004$                                     &   $0.5~/~0.2$                                 \\ 
R-AU-EU/C-AU   & $Par({\text{AU}}/{\text{EU}})$         & $Par(\mathit{e}_{\text{AU}}~/~\mathit{\theta}_{\text{AU}})$         \\
       &  $(1~,~0.4)~/~0.004$                                     &   $0.5~/~0.2$                                 \\ 
	   \bottomrule
\end{tabular}
\end{center}
\end{table}

\subsubsection{5.1.4. Validation of the effectiveness of the proposed method}

To validate the effectiveness of the proposed method, the six types of uncertainty are combined according to scenarios that may occur in real-world applications, resulting in six representative combinations as shown in Table \ref{tab:s3}. The last combination represents a scenario frequently encountered in real industrial settings, where the robot's uncertainty includes both AU and EU, while the measurement camera exhibits AU-type uncertainty. Since different types of error scenarios are constructed, the comparison methods selected are all general-purpose solutions that do not rely on explicit modeling of uncertainty. L-HED, UAL-HED, and SI-AH are all methods proposed in this paper, as described in Section 4. Dual Quaternions and the Kronecker Product are two classical solutions commonly used in HEC problem and they are summarized in  \cite{w31}. The LMI-SDP optimization (\cite{w47}) is an iterative method with favorable performance, abbreviated as LMI in the experimental section of this paper. The Point Cloud Matching in \cite{w52} algorithm differs from the six methods mentioned above in that it does not involve any synchronized or jointly solved steps; instead, it is a decoupled solution approach. Figure \ref{s3} presents the validation results base on 120 sets of synthesized data \{\({\boldsymbol{A_{i}}}\)\} and  \{\({\boldsymbol{B_{i}}}\)\} for each experimental error scenarios. The total error is computed based on as
\begin{equation}
	\label{eq:104}
   \left\{
		\begin{array}{l}
\begin{aligned}\operatorname{Err}_{\boldsymbol{R}} & ={\frac{1}{N} \sum_{i=1}^{N}\left\|\log \left(\boldsymbol{R}_{i}^{\mathrm{T}} {\boldsymbol{R}_{opt}}\right)\right\|} \\
\operatorname{Err}_{\boldsymbol{t}} & ={\frac{1}{N} \sum_{i=1}^{N}\left\|\boldsymbol{t}_{i}-{\boldsymbol{t}_{opt}}\right\|}\\
\operatorname{Err}_{T}&=(\operatorname{Err}_{\boldsymbol{R}}+\operatorname{Err}_{\boldsymbol{t}})\times 100\%\end{aligned},
		\end{array}
		\right.
\end{equation}
where $\operatorname{Err}_{T}$ represents the total error, which corresponds to the metric shown in the Figure \ref{s3}. It is visually evident that, under all six uncertainty combinations, the proposed methods L-HED and UAL-HED achieve relatively low computational errors. Compared to L-HED, UAL-HED reduces the average estimation error of the HECPs for $\boldsymbol{X}$ by 7.1\%, and similarly achieves a 10.6\% reduction in the average estimation error for $\boldsymbol{Y}$ . Compared to LMI, UAL-HED reduces the average estimation error of the HECPs for $\boldsymbol{X}$ by 81.1\%, and similarly achieves a 79.5\% reduction in the average estimation error for $\boldsymbol{Y}$ . The decoupled Point Cloud Matching method performs the worst, in terms of estimation error for  $\boldsymbol{X}$, UAL-HED achieves a 91.7\% lower error, and for $\boldsymbol{Y}$, UAL-HED also reduces the error by 90.6\% compared to the point cloud method. The estimation total error of $\boldsymbol{X}$ obtained using the Dual Quaternion method is lower than that obtained with the Kronecker Product method; however, the situation is reversed for the estimation of$\boldsymbol{Y}$, where the Kronecker product yields better accuracy.

\begin{table*}[t]
\centering
\caption{\ \ \leftskip=0pt \rightskip=0pt plus 0cm Estimation errors of HECPs with multiple error measurements}
\label{tab:hecps_full_errors}
\footnotesize
\setlength{\tabcolsep}{3.5pt} % åå°åé´è·ä»¥éåºæ´å¤å1¤7
\begin{tabular}{@{}ll*{12}{S[table-format=1.4]}@{}}
\toprule
\multirow{2}{*}{Method} & 
\multirow{2}{*}{Uncertainty Setting} & 
\multicolumn{12}{c}{Error Types} \\
\cmidrule(lr){3-14}
& & \multicolumn{3}{c}{Rotational of $\boldsymbol{X}$ $(10^-3)$} & \multicolumn{3}{c}{Rotational of $\boldsymbol{Y}$  $(10^-3)$} & \multicolumn{3}{c}{Traditional of $\boldsymbol{X}$  $(mm)$} & \multicolumn{3}{c}{Traditional of $\boldsymbol{Y}$  $(mm)$} \\
\cmidrule(lr){3-5}\cmidrule(lr){6-8}\cmidrule(lr){9-11}\cmidrule(lr){12-14}
& & {Mean} & {Max} & {Min} & {Mean} & {Max} & {Min} & {Mean} & {Max} & {Min} & {Mean} & {Max} & {Min} \\
\midrule
L-HED    & R-AU    & 0.000 & 0.000 & 0.000 & 0.033 & 0.089 & 0.002 & 0.039 & 0.161 & 0.005 & 0.0020 & 0.0017 & 0.0022 \\
         & C-AU    & 0.000 & 0.000 & 0.000 & 0.013 & 0.038 & 0.002 & 0.017 & 0.069 & 0.001 & 0.0021 & 0.0018 & 0.0023 \\
         & R-AU/C-AU & 0.122 & 0.927 & 0.000 & 0.150 & 0.511 & 0.015 & 0.165 & 0.411 & 0.096 & 0.0022 & 0.0019 & 0.0024 \\
         & R-EU    & 0.000 & 0.000 & 0.000 & 0.039 & 0.109 &  0.005 & 0.061 & 0.167 & 0.011 & 0.0023 & 0.0020 & 0.0025 \\
         & R-EU/C-AU & 0.031 & 0.627 & 0.000 & 0.130 & 0.321 & 0.013 & 0.154 & 0.472 & 0.029 & 0.0024 & 0.0021 & 0.0026 \\
         & R-AU-EU/C-AU & 0.022 & 0.609 & 0.000 & 0.130 & 0.382 & 0.018 & 0.152 & 0.493 & 0.034 & 0.0025 & 0.0022 & 0.0027 \\
\addlinespace
UAL-HED & R-AU    & 0.000 & 0.000 & 0.000 & 0.027 & 0.085 & 0.002 &0.033 & 0.129 & 0.004 & 0.0020 & 0.0017 & 0.0022 \\
         & C-AU    & 0.000 & 0.000 & 0.000 & 0.010 & 0.030 & 0.001 & 0.014 & 0.069 & 0.001 & 0.0021 & 0.0018 & 0.0023 \\
         & R-AU/C-AU & 0.113 &  1.056 & 0.000 & 0.139 & 0.430 & 0.026 &  0.164 & 0.390 & 0.015 & 0.0022 & 0.0019 & 0.0024 \\
         & R-EU    & 0.000 & 0.000 & 0.000 & 0.035 & 0.091 & 0.003 & 0.051 & 0.150 & 0.009 & 0.0023 & 0.0020 & 0.0025 \\
         & R-EU/C-AU & 0.018 & 0.632 & 0.000 & 0.121 & 0.314 & 0.029 & 0.151 & 0.396 &  0.021 & 0.0024 & 0.0021 & 0.0026 \\
         & R-AU-EU/C-AU & 0.021 & 0.684 & 0.000 & 0.119 & 0.331 & 0.015 & 0.142 & 0.389 & 0.012 & 0.0025 & 0.0022 & 0.0027 \\
\addlinespace

DQ     & R-AU    & 0.411 & 1.308 & 0.000 & 0.437 & 0.965 & 0.148 & 0.720 & 1.666 & 0.121 & 0.0020 & 0.0017 & 0.0022 \\
         & C-AU    & 0.031 & 0.624 & 0.000 & 0.224 & 0.562 & 0.049 &0.326 & 0.682 & 0.062 & 0.0021 & 0.0018 & 0.0023 \\
         & R-AU/C-AU & 0.573 & 1.416 & 0.000 & 0.518 & 1.175 & 0.035 & 0.791 & 1.739 & 0.165 & 0.0022 & 0.0019 & 0.0024 \\
         & R-EU    & 0.532 & 1.629 & 0.000 & 0.498 & 1.254 & 0.115 & 0.776 & 1.696 & 0.043 & 0.0023 & 0.0020 & 0.0025 \\
         & R-EU/C-AU & 0.559 & 1.463 & 0.000 & 0.544 & 1.149 & 0.120 & 0.825 & 1.966 & 0.103 & 0.0024 & 0.0021 & 0.0026 \\
         & R-AU-EU/C-AU & 0.533 & 1.760 & 0.000 & 0.539 & 1.356 & 0.079 & 0.813 & 1.876 & 0.133 & 0.0025 & 0.0022 & 0.0027 \\
\addlinespace

KP & R-AU    & 0.706 & 1.839 & 0.000 & 0.753 & 1.843 & 0.078 & 0.248 & 0.609 & 0.070 & 0.0020 & 0.0017 & 0.0022 \\
         & C-AU    & 0.138 & 0.933 & 0.000 & 0.359& 0.988 & 0.021 & 0.103 & 0.284 & 0.017 & 0.0021 & 0.0018 & 0.0023 \\
         & R-AU/C-AU & 0.766 & 1.888 & 0.000 & 0.815 & 1.818 & 0.178 & 0.280 & 0.553 & 0.053 & 0.0022 & 0.0019 & 0.0024 \\
         & R-EU    & 0.681 & 1.974 & 0.000 & 0.719 & 1.494 & 0.161 & 0.245 & 0.753 & 0.036 & 0.0023 & 0.0020 & 0.0025 \\
         & R-EU/C-AU & 0.764 & 1.886 & 0.000 & 0.847 & 1.680 & 0.189 & 0.254 & 0.795 & 0.047 & 0.0024 & 0.0021 & 0.0026 \\
         & R-AU-EU/C-AU & 0.733 & 1.745 & 0.000 & 0.804 & 1.885 & 0.173 & 0.259 & 1.028 & 0.046 & 0.0025 & 0.0022 & 0.0027 \\
\addlinespace

LMI & R-AU    & 0.434 & 1.562 & 0.000 & 0.369 & 0.886 & 0.066 & 0.177 & 0.429 &  0.052 & 0.0020 & 0.0017 & 0.0022 \\
         & C-AU    & 0.034 & 0.595 & 0.000 & 0.178 & 0.526 & 0.028 & 0.082 & 0.183 &  0.016 & 0.0021 & 0.0018 & 0.0023 \\
         & R-AU/C-AU &  0.507 & 1.486 & 0.000 & 0.402 & 0.883 & 0.058 & 0.232 & 0.434 & 0.015 & 0.0022 & 0.0019 & 0.0024 \\
         & R-EU    & 0.449 & 1.303 & 0.000 & 0.359 & 0.925 & 0.037 & 0.188 &0.483 & 0.053 & 0.0023 & 0.0020 & 0.0025 \\
         & R-EU/C-AU & 0.487 & 1.320 & 0.000 & 0.421 & 0.968 & 0.044 & 0.195 & 0.484 & 0.036 & 0.0024 & 0.0021 & 0.0026 \\
         & R-AU-EU/C-AU &  0.444 & 1.668 & 0.000 & 0.407 & 1.017 & 0.072 & 0.198 & 0.631 & 0.025 & 0.0025 & 0.0022 & 0.0027 \\
\addlinespace

PCM & R-AU    &  0.714 & 1.999 & 0.000 & 0.754 & 1.817 & 0.062 & 0.763 & 1.881 & 0.209 & 0.0020 & 0.0017 & 0.0022 \\
         & C-AU    &  0.134 & 0.942 & 0.000 & 0.359 & 0.992 & 0.017 & 0.380 & 0.888 & 0.079 & 0.0021 & 0.0018 & 0.0023 \\
         & R-AU/C-AU &  0.767 & 1.894 & 0.000 & 0.816 & 1.784 & 0.172 & 0.875 & 1.906 & 0.139 & 0.0022 & 0.0019 & 0.0024 \\
         & R-EU    & 0.678 & 1.949 & 0.000 & 0.715 & 1.482 & 0.154 & 0.880& 1.875 & 0.203 & 0.203 & 0.0020 & 0.0025 \\
         & R-EU/C-AU & 0.753 & 1.905 & 0.000 & 0.842 & 1.660 & 0.192 & 0.865 & 1.984 & 0.104  & 0.104 & 0.0021 & 0.0026 \\
         & R-AU-EU/C-AU & 0.725 & 1.742 & 0.000 & 0.799 & 1.858 & 0.174 & 0.888 & 1.960 & 0.126 & 0.126 & 0.0022 & 0.0027 \\
\addlinespace

SI-AH & R-AU    & 0.000 & 0.000 & 0.000 & 0.025 & 0.077 & 0.005 & 0.028 & 0.0022 & 0.0027 & 0.0026 & 0.0023 & 0.0028 \\
         & C-AU    & 0.000 & 0.000 & 0.000 & 0.012 & 0.040 & 0.002 & 0.015 & 0.0023 & 0.0028 & 0.0027 & 0.0024 & 0.0029 \\
         & R-AU/C-AU &  0.617 & 1.797 & 0.000 & 0.702 & 1.853 &  0.062 & 0.653 & 0.0024 & 0.0029 & 0.0028 & 0.0025 & 0.0030 \\
         & R-EU    & 0.000 & 0.000 & 0.000 & 0.038 & 0.086 &  0.003 & 0.057 & 0.0025 & 0.0030 & 0.0029 & 0.0026 & 0.0031 \\
         & R-EU/C-AU & 0.395 & 1.593 & 0.000& 0.412 & 1.251 & 0.045 & 0.342 & 0.0026 & 0.0031 & 0.0030 & 0.0027 & 0.0032 \\
         & R-AU-EU/C-AU & 0.346 & 1.382 & 0.000 & 0.399 &  1.189 & 0.042 & 0.335 & 0.0027 & 0.0032 & 0.0031 & 0.0028 & 0.0033 \\
\bottomrule
\end{tabular}

\begin{minipage}{\linewidth}
%\footnotesize
%\textit{Note:} Error Types: Type 1 = X error, Type 2 = Y error, Type 3 = Z error, Type 4 = Combined error. 
%Measurements: M1 = Condition 1, M2 = Condition 2, M3 = Condition 3. 
%All values are in millimeters.
\end{minipage}
\end{table*}

It is worth noting that in scenarios with relatively simple uncertainty combinations, the proposed SI-AH method originally intended for generating initial estimates also demonstrates strong performance, even achieving the best result under the first scenario. The reason for this phenomenon is that, although L-HED possesses the capability for globally optimal iteration, the heuristic metric is not constructed based on the true ideal matrices $\boldsymbol{X}_{opt}$ and $\boldsymbol{Y}_{opt}$. Consequently, when the initial matrices $\boldsymbol{X}_{0}$ and $\boldsymbol{Y}_{0}$ are already close in accuracy to $\boldsymbol{X}_{opt}$ and $\boldsymbol{Y}_{opt}$, there is a risk of overfitting during the solution process.
Additionally, when both robot and camera uncertainties are present, the solution accuracy of SI-AH tends to degrade. Across the six scenarios, UAL-HED achieves, on average, a 74.7\% lower estimation error than SI-AH for $\boldsymbol{X}$, and similarly, a 73.5\% lower estimation error for  $\boldsymbol{Y}$.

An interesting observation is that the data presented in Figure \ref{s3} reflects average performance; in simulations, for any fixed uncertainty combthere is a probability that  synthesized pair of \{\({\boldsymbol{A_{i}}}\)\} and  \{\({\boldsymbol{B_{i}}}\)\} for which another method achieves the highest estimation accuracy instead of the proposed methods. ination, However, the proposed UAL-HED method most frequently attains the highest accuracy across these instances, which ultimately results in its superior average estimation accuracy for the HECPs $\boldsymbol{X}$ and $\boldsymbol{Y}$.

\begin{table}[!htbp]
\centering
\caption{\ \ \leftskip=0pt \rightskip=0pt plus 0cm Uncertainty parameter settings for source data}
\label{tab:s4}
\footnotesize
\begin{tabular}{m{2.3cm}<{\centering} m{2.3cm}<{\centering} m{2.3cm}<{\centering}}
\toprule
\color{black}{Uncertainty level} & \{\({\boldsymbol{A_{i}}}\)\} & \{\({\boldsymbol{B_{i}}}\)\} \\ \midrule
High    & $Par(\mathit{e}_{\text{AU}}~/~\mathit{\theta}_{\text{AU}})$         &  $Par(\mathit{e}_{\text{AU}}~/~\mathit{\theta}_{\text{AU}})$         \\
       &  $1~/~0.4$                                     &  $0.5~/~0.2$                            \\
Low &  $Par(\mathit{e}_{\text{AU}}~/~\mathit{\theta}_{\text{AU}})$         &$Par(\mathit{e}_{\text{AU}}~/~\mathit{\theta}_{\text{AU}})$         \\
       & $0.2~/~0.1$                                      & $0.1~/~0.05$                             \\
	   \bottomrule
\end{tabular}
\end{table}

\subsubsection{5.1.5. Comparison of the calibration performance of seven methods in estimating HECPs under various source data conditions}

As described in Section 1, regarding the estimation of HECPs, regardless of the complexity of the uncertainty combinations from the robot and the camera, their influence on the solution accuracy ultimately manifests only in the source data \{\({\boldsymbol{A_{i}}}\)\} and  \{\({\boldsymbol{B_{i}}}\)\}. Therefore, to better investigate the effectiveness of different methods under various error conditions, uncertainty combinations are designed directly on the source data  \{\({\boldsymbol{A_{i}}}\)\} and  \{\({\boldsymbol{B_{i}}}\)\} to evaluate which method performs best under each combination. The translational and rotational uncertainties of source data \{\({\boldsymbol{A_{i}}}\)\} and \{\({\boldsymbol{B_{i}}}\)\} are set at two levels to generate 16 combinations. Each combination is used to synthesize 60 sets of samples for validation. The parameter settings for each type of uncertainty are summarized in the Table \ref{tab:s4}.
\begin{figure}[!htbp]\centering
	\includegraphics[width=0.45\textwidth]{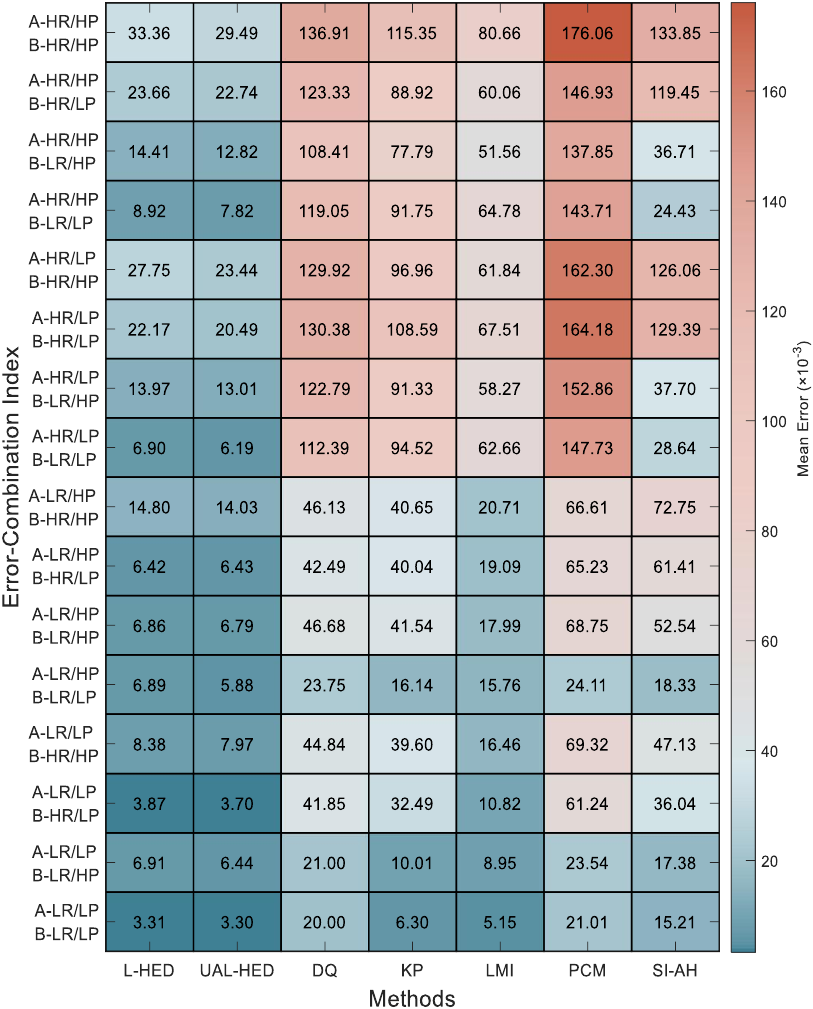}
	\caption{\ \ \leftskip=0pt \rightskip=0pt plus 0cm Mean error of $\boldsymbol{X}$ under various error conditions.}
	\label{s4-1}
\end{figure}
\begin{figure}[!htbp]\centering
	\includegraphics[width=0.4\textwidth]{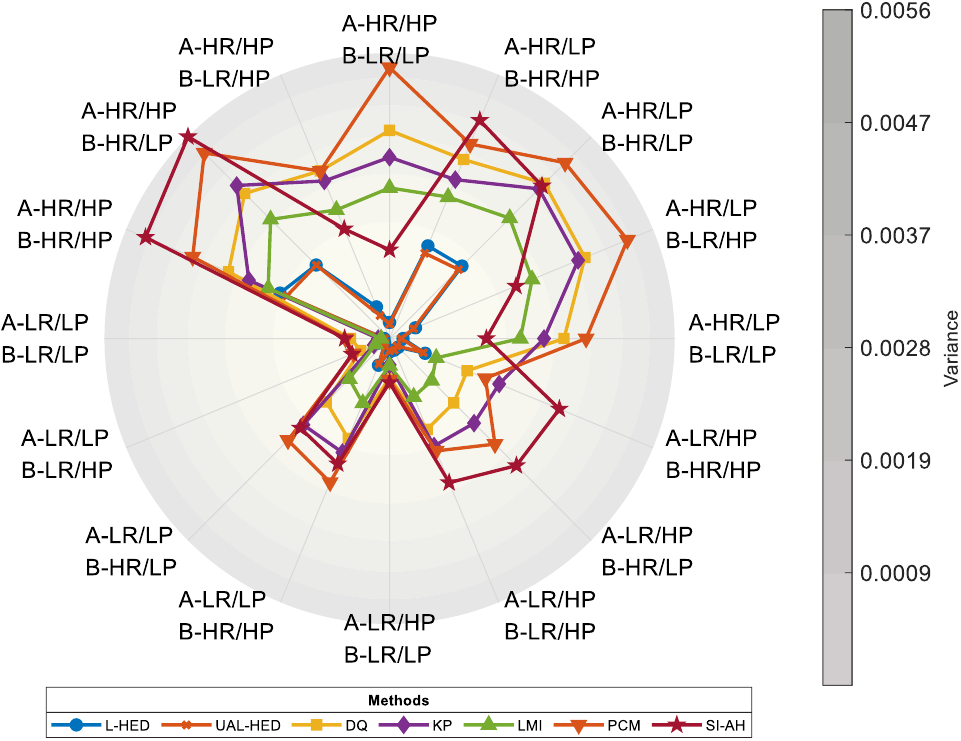}
	\caption{\ \ \leftskip=0pt \rightskip=0pt plus 0cm variance of $\boldsymbol{X}$ under various error conditions.}
	\label{s4-3}
\end{figure}

Figures \ref{s4-1} and \ref{s4-3} respectively present the average estimation accuracy and variance of  $\boldsymbol{X}$ obtained from source data \{\({\boldsymbol{A_{i}}}\)\} and \{\({\boldsymbol{B_{i}}}\)\} under 16 different uncertainty settings using the seven methods, the formula for the mean is given by Equation (\ref{eq:104}) and the variance is given as

\begin{equation}
	\label{eq:105}
   \left\{
		\begin{array}{l}
\begin{aligned}\operatorname{Var}_{\boldsymbol{R}} & ={\frac{1}{N-1} \sum_{i=1}^{N}\left\|\operatorname{Err}_{\boldsymbol{R}}-\overline{\operatorname{Err}_{\boldsymbol{R}}}\right\|^2} \\
\operatorname{Var}_{\boldsymbol{t}} & ={\frac{1}{N-1} \sum_{i=1}^{N}\left\|\operatorname{Err}_{\boldsymbol{t}}-\overline{\operatorname{Err}_{\boldsymbol{t}}}\right\|^2} \\
\operatorname{Var}_{\boldsymbol{T}} & ={\frac{1}{N-1} \sum_{i=1}^{N}\left\|\operatorname{Err}_{\boldsymbol{T}}-\overline{\operatorname{Err}_{\boldsymbol{T}}}\right\|^2}\end{aligned},
		\end{array}
		\right.
\end{equation}
where the values in Figure \ref{s4-1} are presented based on the metric $\operatorname{Err}_{\boldsymbol{T}}$, while the values in Figure \ref{s4-3} are based on the metric $\operatorname{Var}_{\boldsymbol{T}}$. 

In Figure \ref{s4-1}, by comparing the uncertainty levels of the source data, it can be observed that when both the rotational and translational uncertainties of \{\({\boldsymbol{A_{i}}}\)\} and \{\({\boldsymbol{B_{i}}}\)\} are low level, all seven methods demonstrate relatively good estimation performance. Conversely, when the rotational and translational uncertainties of source data \{\({\boldsymbol{A_{i}}}\)\} and \{\({\boldsymbol{B_{i}}}\)\} are both high level, the estimation accuracy of all seven methods is noticeably lower compared to the accuracy achieved under low uncertainty level, the estimation errors for the HECPs of $\boldsymbol{X}$ for the seven methods increased respectively by: 30.65, 26.19, 116.91, 109.05, 75.51, 155.05, and 118.64. It can be observed that the two decoupled methods for solving the HECPs of  $\boldsymbol{X}$ and  $\boldsymbol{Y}$ exhibit the greatest decrease in estimation accuracy which indicates that under high-uncertainty data sources, the propagation and accumulation of errors significantly affect the estimation accuracy.

A noteworthy observation is that uncertainty in the rotational direction has a greater tendency to disrupt the estimation of HECPs compared to positional uncertainty. The proposed methods, L-HED, UAL-HED and SI-AH exhibit the same estimation accuracy for  $\boldsymbol{X}$ when either only source data \{\({\boldsymbol{A_{i}}}\)\} or only source data \{\({\boldsymbol{B_{i}}}\)\} has high uncertainty. In contrast, the other four methods show significantly lower estimation accuracy for $\boldsymbol{X}$ when only source data \{\({\boldsymbol{A_{i}}}\)\} has high uncertainty, compared to when only source data $B$ has high uncertainty. As the uncertainty in the source data increases, the relative effectiveness of UAL-HED compared to L-HED gradually improves.

In Figure \ref{s4-3} presents the variance in estimating the HECPs for $X$ using the seven methods under source data constructed with different uncertainty levels. The variance of the estimation errors for all seven methods increases with the level of uncertainty. Specifically, high-level uncertainties, especially in the rotational component lead to a significant increase in variance. Overall, under the proposed conditions, the UAL-HED method exhibits the smallest variance in the estimation error for $\boldsymbol{X}$, followed by L-HED. The Point-Cloud Matching and SI-AH methods show the largest variances. Among the remaining three methods, the LMI method performs the best. This phenomenon implies two key points, one is no method can consistently estimate exactly the same $\boldsymbol{X}$ across different source data sets with the same level of uncertainty combination, the other is when addressing real-world HEC problems, the proposed UAL-HED method is the least sensitive to variations in the source data and achieves the highest estimation accuracy for HECPs of $\boldsymbol{X}$.
\begin{figure}[!htbp]\centering
	\includegraphics[width=0.45\textwidth]{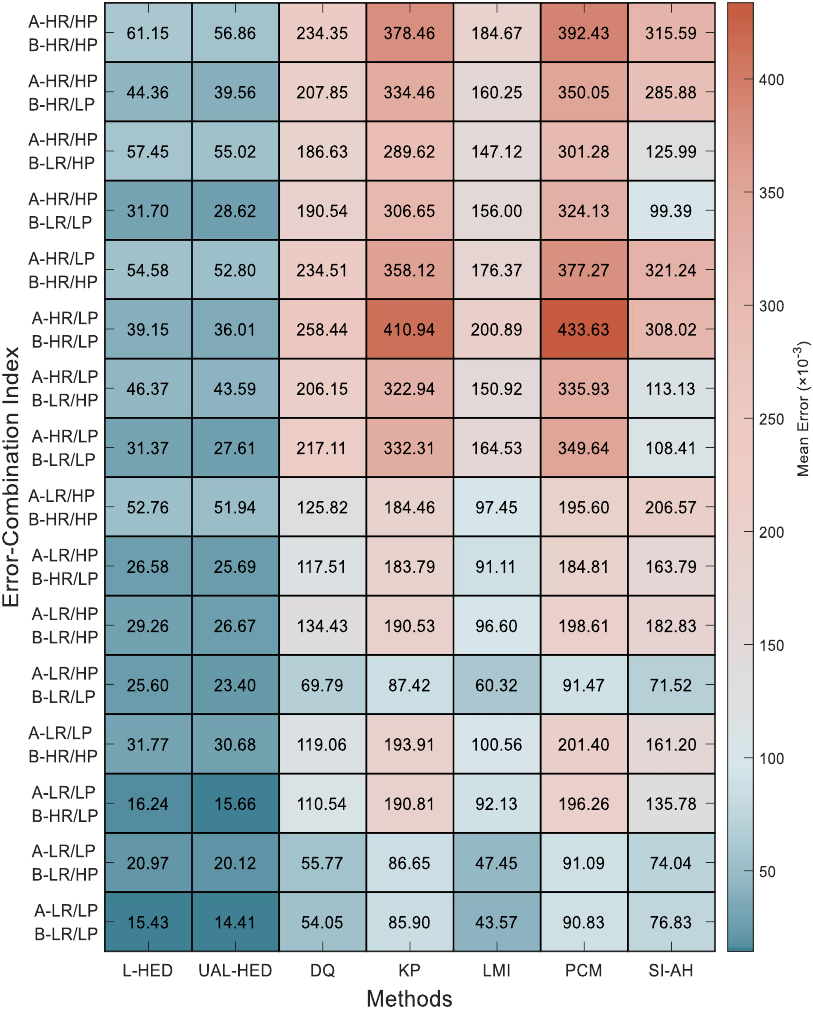}
	\caption{Mean error of $\boldsymbol{Y}$ under various error conditions.}
	\label{s4-2}
\end{figure}

\begin{figure}[!htbp]\centering
	\includegraphics[width=0.4\textwidth]{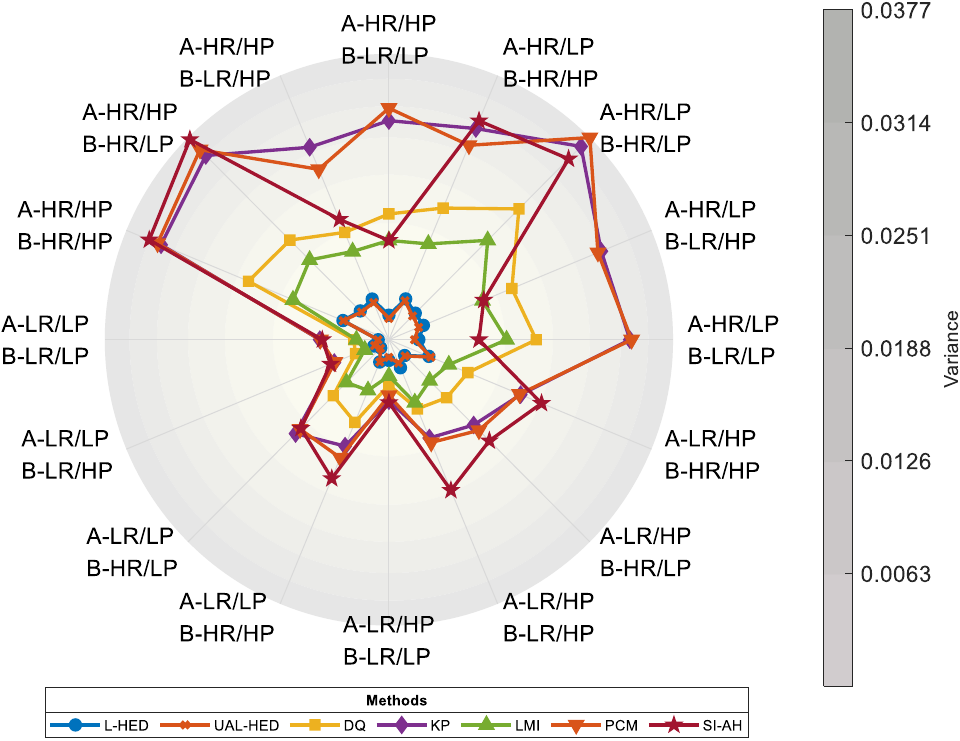}
	\caption{\ \ \leftskip=0pt \rightskip=0pt plus 0cm Variance of $\boldsymbol{Y}$ under various error conditions.}
	\label{s4-4}
\end{figure}

Similarly, Figures \ref{s4-2} and \ref{s4-4} respectively present the average estimation accuracy and variance of  $\boldsymbol{Y}$ obtained from source data \{\({\boldsymbol{A_{i}}}\)\} and \{\({\boldsymbol{B_{i}}}\)\} under 16 different uncertainty settings using the seven methods, and the observed phenomena are essentially consistent with the results shown in Figures \ref{s4-1} and \ref{s4-3}. In contrast, the estimation errors for HECPs of $\boldsymbol{Y}$ are higher than those for $\boldsymbol{X}$ across all seven methods. Additionally, the Dual-Quaternion method yields higher accuracy for $\boldsymbol{Y}$ compared to the Kronecker-Product method, which is the opposite of the trend observed in the estimation of $\boldsymbol{X}$.

\subsubsection{5.1.6. Optimal construction form for iterative estimation}

Due to the absence of a bi-invariant measure on the Euclidean group, the choice of different closed-form construction formulations for iterative estimation such as in hand-eye and robot-world calibration, formulated as $\boldsymbol{AX=YB}$ may lead to varying effects on the optimization outcome. The four Closed-Form formulations of $\boldsymbol{AX=YB}$ are defined in Table.~\ref{tab:s5}.
\begin{table}[!htbp]
\centering
\caption{\ \ \leftskip=0pt \rightskip=0pt plus 0cm Four Construction  Closed-Forms Used for Iterative Solving of the $\boldsymbol{AX=YB}$ Problem}
\label{tab:s5}
\footnotesize
\begin{tabular}{m{4.2cm}<{\centering} m{3.0cm}<{\centering}}
\toprule
\color{black}{Construction Forms} &  $\boldsymbol{AX=YB}$ \\ \midrule
Closed-Form 1    & $ \boldsymbol{Y^{-1}AXB^{-1}}$       \\ 
Closed-Form 2 &  $\boldsymbol{AXB^{-1}Y^{-1}}$       \\
Closed-Form 3  &  $\boldsymbol{XB^{-1}Y^{-1}A}$        \\
Closed-Form 4 &  $\boldsymbol{B^{-1}Y^{-1}AX}$        \\
	   \bottomrule
\end{tabular}
\end{table}

\begin{figure}[!htbp]\centering
	\includegraphics[width=0.45\textwidth]{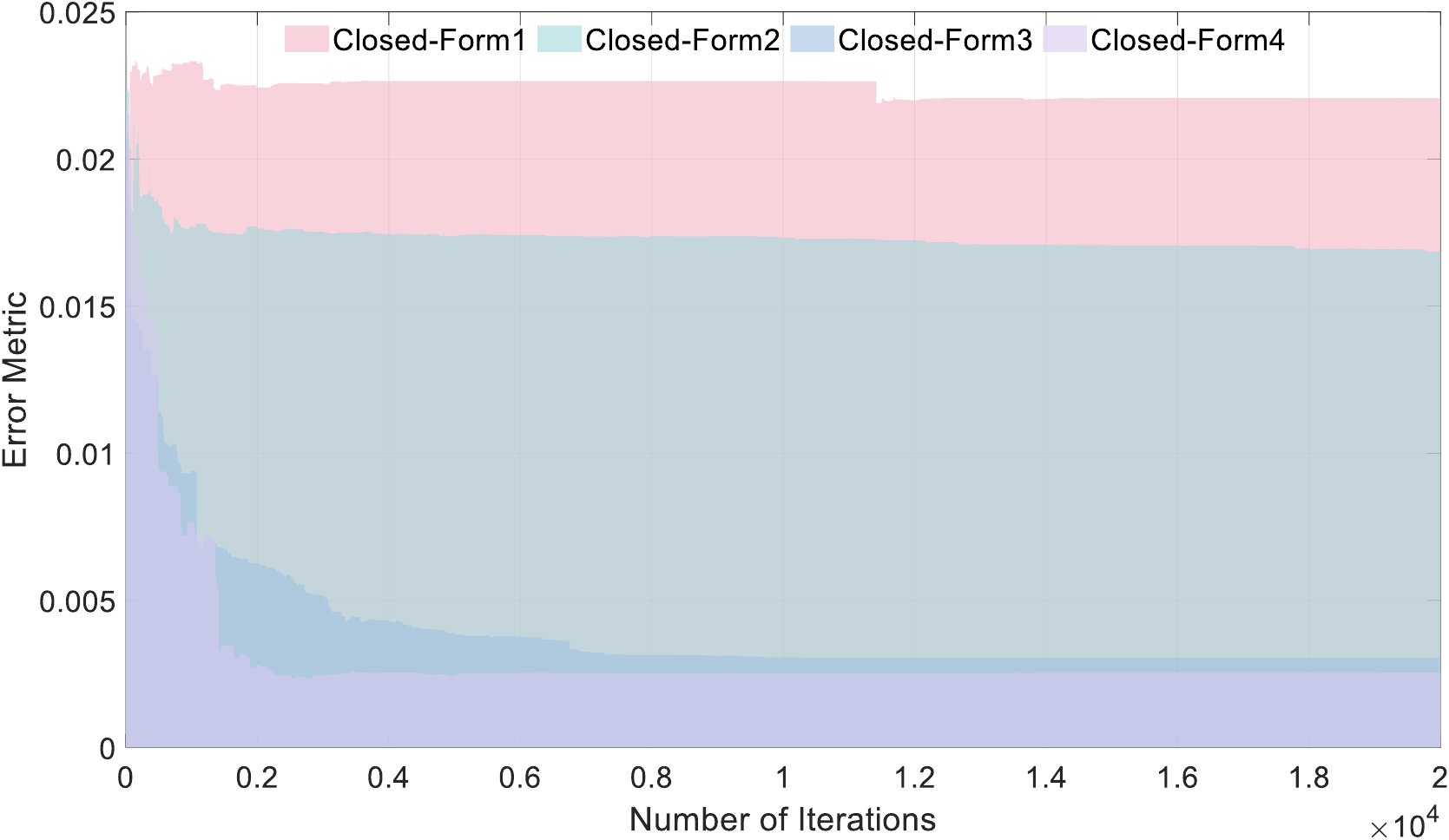}
	\caption{\ \ \leftskip=0pt \rightskip=0pt plus 0cm Impact of bi-nvariant distance absence on \(SE(3)\) iterative processes.}
	\label{s5-1}
\end{figure}

The four formulations are applied within the L-HED method using the same initial value, the relationship between the number of iterations and the estimation error is as Figure \ref{s5-1}.

As illustrated in Figure \ref{s5-1}, when the initial value is relatively close to the ground truth, Closed-Form 1 exhibits almost no further iteration, Closed-Form 2 shows poor iterative performance, while Closed-Form 3 and Closed-Form 4 demonstrate better convergence behavior, with both achieving comparable estimation accuracy after iteration. The primary reason is believed to be that the uncertainties in source data \{\({\boldsymbol{A_{i}}}\)\} and \{\({\boldsymbol{B_{i}}}\)\} interfere with the heuristic metric escape decisions. According to Equation \ref{eq:23}, the following processing is applied to the four closed-form formulations as
\begin{equation}
	\label{eq:106}
   \left\{
		\begin{array}{l}
\begin{aligned}\vectYY_i^{-1}\boldsymbol{A} \vectXX_i \boldsymbol{B}^{-1} &=\vectYY^{-1}(\boldsymbol{A_{o}}+\Delta_{\boldsymbol{A}}) \vectXX (\boldsymbol{B_{o}}^{-1}+\Delta_{\boldsymbol{B}^{-1}})  \\
\boldsymbol{A} \vectXX_i \boldsymbol{B}^{-1} \vectYY_i^{-1}&=(\boldsymbol{A_{o}}+\Delta_{\boldsymbol{A}}) \vectXX (\boldsymbol{B_{o}}^{-1}+\Delta_{\boldsymbol{B}^{-1}}) \vectYY^{-1}\\
 \vectXX_i \boldsymbol{B}^{-1} \vectYY_i^{-1}\boldsymbol{A}&= \vectXX (\boldsymbol{B_{o}}^{-1}+\Delta_{\boldsymbol{B}^{-1}}) \vectYY^{-1}(\boldsymbol{A_{o}}+\Delta_{\boldsymbol{A}})\\
 \boldsymbol{B}^{-1} \vectYY_i^{-1}\boldsymbol{A} \vectXX_i&= (\boldsymbol{B_{o}}^{-1}+\Delta_{\boldsymbol{B}^{-1}}) \vectYY^{-1}(\boldsymbol{A_{o}}+\Delta_{\boldsymbol{A}}) \vectXX\\
\end{aligned},
		\end{array}
		\right.
\end{equation}
where $\boldsymbol{A_{o}}$ and $\boldsymbol{B_{o}}$ represent uncertainty-free for source data \{\({\boldsymbol{A_{i}}}\)\} and \{\({\boldsymbol{B_{i}}}\)\}, $\Delta_{\boldsymbol{A}}$ and $\Delta_{\boldsymbol{B}^{-1}}$ are the perturbations. Applying a similar treatment as in Equations \ref{eq:24} and \ref{eq:27}, the error introduced by source data uncertainty to the closed-form formulations is given as
\begin{equation}
	\label{eq:107}
   \left\{
		\begin{array}{l}
\begin{aligned}\delta\boldsymbol{e_1} &=\vectYY^{-1}\Delta_{\boldsymbol{A}}\boldsymbol{A_{o}^{-1}}\vectYY\delta+\boldsymbol{B_{o}}\Delta_{\boldsymbol{B}^{-1}}\\
\delta\boldsymbol{e_2} &=\vectYY\boldsymbol{B_{o}}\Delta_{\boldsymbol{B}^{-1}}\vectYY^{-1}+\Delta_{\boldsymbol{A}}\boldsymbol{A_{o}^{-1}}\\
\delta\boldsymbol{e_3} &=\vectXX\Delta_{\boldsymbol{B}^{-1}}\boldsymbol{B_{o}}\vectXX^{-1}+\boldsymbol{A_{o}^{-1}}\Delta_{\boldsymbol{A}}\\
\delta\boldsymbol{e_4} &=\vectXX^{-1}\boldsymbol{A_{o}^{-1}}\Delta_{\boldsymbol{A}}\vectXX+\Delta_{\boldsymbol{B}^{-1}}\boldsymbol{B_{o}}\\
\end{aligned},
		\end{array}
		\right.
\end{equation}
where Equation \ref{eq:107} illustrates the error interference patterns introduced by the uncertainties in source data \{\({\boldsymbol{A_{i}}}\)\} and \{\({\boldsymbol{B_{i}}}\)\} under the four closed-form constructions, and it is clearly observed that the errors in Closed-Form 1 and Closed-Form 2 are related to the HECPs for $\boldsymbol{Y}$, while the errors in Closed-Form 3 and Closed-Form 4 are associated with the HECPs for $\boldsymbol{X}$. The HECPs for $\boldsymbol{X}$ and $\boldsymbol{Y}$ which are used in simulations are shown in Table \ref{tab:s5-1}. It is hypothesized that, since the translational component of $\boldsymbol{Y}$ has larger values than that of $\boldsymbol{X}$, the errors introduced by source data \{\({\boldsymbol{A_{i}}}\)\} and \{\({\boldsymbol{B_{i}}}\)\} in Closed-Form 1 and Closed-Form 2 are greater than those in Closed-Form 3 and Closed-Form 4. This, in turn, affects the decision-making during the iterative process.

\begin{table}[!hbtp]
\centering
\caption{\ \ \leftskip=0pt \rightskip=0pt plus 0cm The HECPs for $\boldsymbol{X}$ and $\boldsymbol{Y}$}
\label{tab:s5-1}
\footnotesize
\begin{tabular}{m{1.2cm}<{\centering} m{6.0cm}<{\centering}}
\toprule
\color{black}{HECPs} &  HTMs \\ \midrule
 $\boldsymbol{X}$    & $ \begin{bmatrix}
        -0.795 & -0.599 & 0.087 & 0.09 \\
         0.604 & -0.795 &  0.052 & -0.2 \\
        0.038 & 0.094 & 0.995 & 0.07 \\
        0 & 0 & 0 & 1
    \end{bmatrix}$       \\ 
$\boldsymbol{Y}$ &  $ \begin{bmatrix}
    -0.965 & -0.258 & -0.035 & 3.6 \\
        0.259 & -0.965 & -0.035 & -3.9 \\
        -0.024 & -0.043 & 0.999 & 0.3 \\
        0 & 0 & 0 & 1
    \end{bmatrix}$       \\
	   \bottomrule
\end{tabular}
\end{table}  

\begin{figure}[!htbp]\centering
	\includegraphics[width=0.45\textwidth]{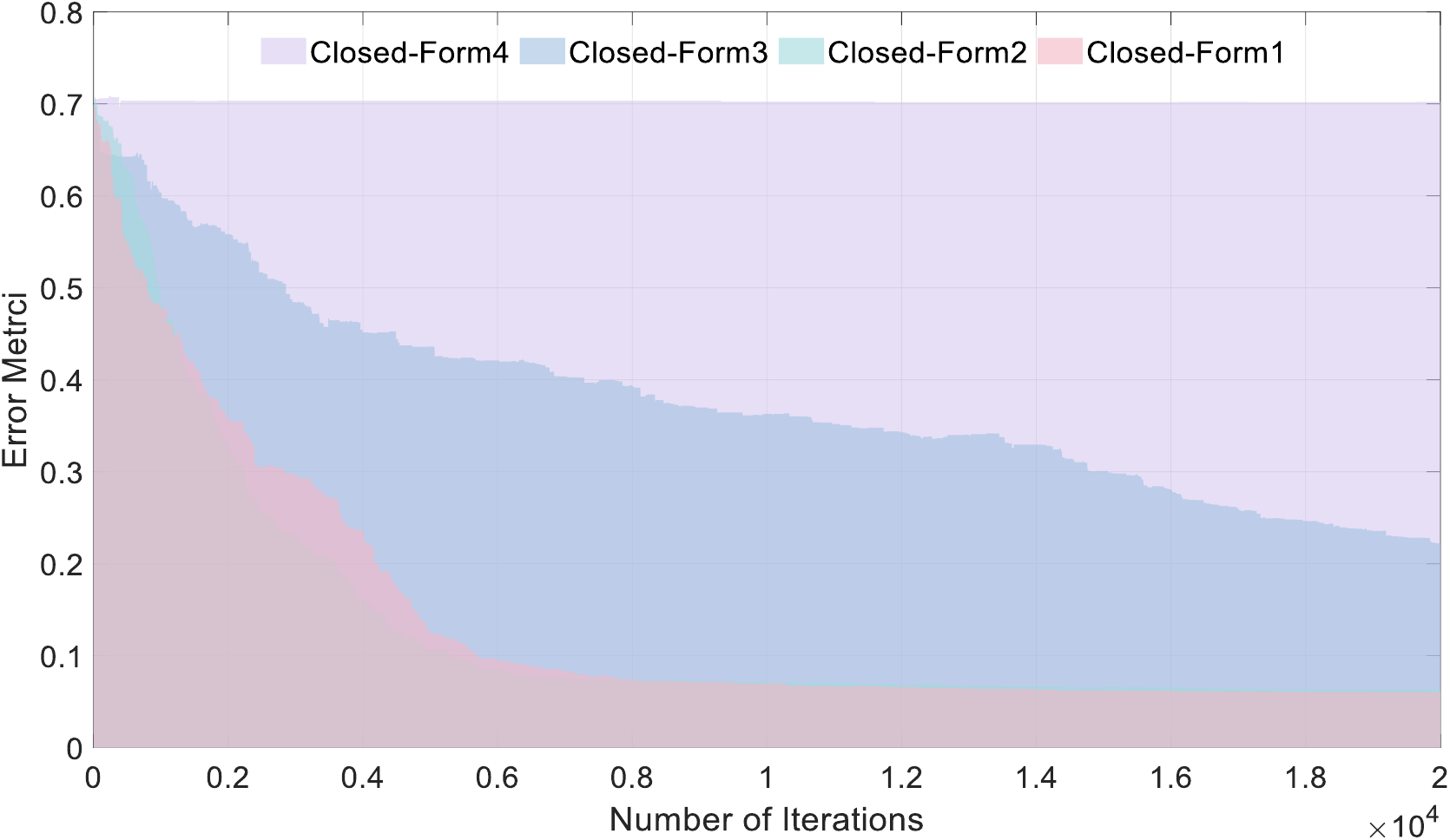}
	\caption{\ \ \leftskip=0pt \rightskip=0pt plus 0cm Swapping the translational components of $\boldsymbol{X}$ and $\boldsymbol{Y}$ impacts the Iterative Processes.}
	\label{s5-2}
\end{figure}
To validate the above hypothesis, the translational components of $\boldsymbol{X}$ and $\boldsymbol{Y}$ are swapped while keeping the rest of the synthesized data unchanged. As shown in Figure \ref{s5-2}, after the above adjustment, Closed-Form 4 exhibits almost no further iteration, Closed-Form 3 shows poor iterative performance, while Closed-Form 1 and Closed-Form 2 demonstrate better convergence behavior, with both achieving comparable estimation accuracy after iteration, this confirms the validity of the hypothesis that the choice of closed-form construction is related to the relative magnitude of the HECPs being estimated.
\begin{table}[!htbp]
\centering
\caption{\ \ \leftskip=0pt \rightskip=0pt plus 0cm Selection strategy for uncertainty metric-based data selection}
\label{tab:s7}
\footnotesize
\begin{tabular}{m{2.7cm}<{\centering} m{4.2cm}<{\centering}}
\toprule
\textrm{\color{black}{Selection strategy}} & \color{black}{Data Selection}   \\
\midrule
1 & {$\displaystyle\max_{1 \leq i \leq 10} M_{\mathrm{unc}}(i)$}  \\
2 &{$\displaystyle\max_{1 \leq i \leq 20} M_{\mathrm{unc}}(i)$}  \\
3 & {$\displaystyle\max_{1 \leq i \leq 50} M_{\mathrm{unc}}(i)$}\\
4 & {$\displaystyle\max_{1 \leq i \leq 100} M_{\mathrm{unc}}(i)$}  \\
5 & {$\displaystyle\max_{10 \leq i \leq 20} M_{\mathrm{unc}}(i)$}\\
6 & {$\displaystyle\max_{10 \leq i \leq 50} M_{\mathrm{unc}}(i)$}\\
7 & {$\displaystyle\max_{10 \leq i \leq 100} M_{\mathrm{unc}}(i)$}\\
8 & {$\displaystyle\max_{50 \leq i \leq 100} M_{\mathrm{unc}}(i)$}\\
\bottomrule
\end{tabular}
\end{table}

\subsubsection{5.1.7. Impact of the number of source data sets used for HECPs estimation on solution accuracy}

\begin{figure}[!htbp]\centering
	\includegraphics[width=0.35\textwidth]{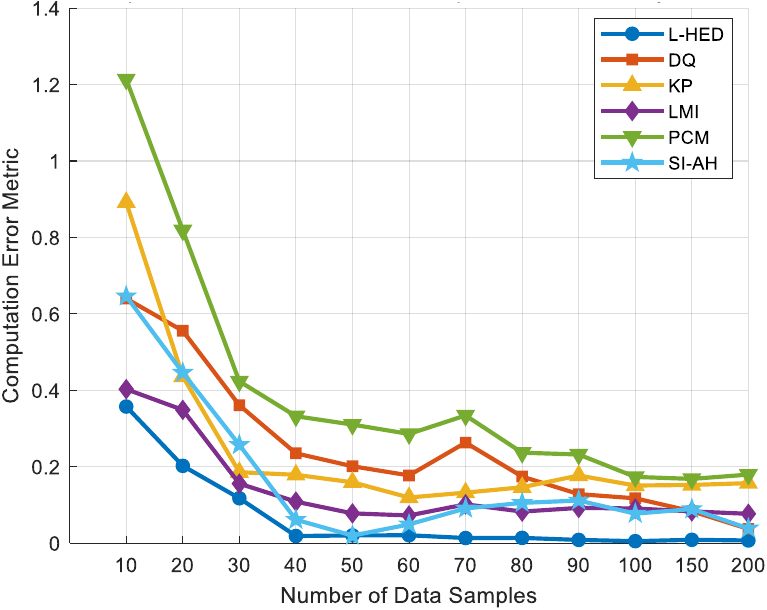}
	\caption{\ \ \leftskip=0pt \rightskip=0pt plus 0cm Impact of the number of data sets on the accuracy of $\boldsymbol{X}$.}
	\label{s6-1}
\end{figure}

\begin{figure}[!htbp]\centering
	\includegraphics[width=0.35\textwidth]{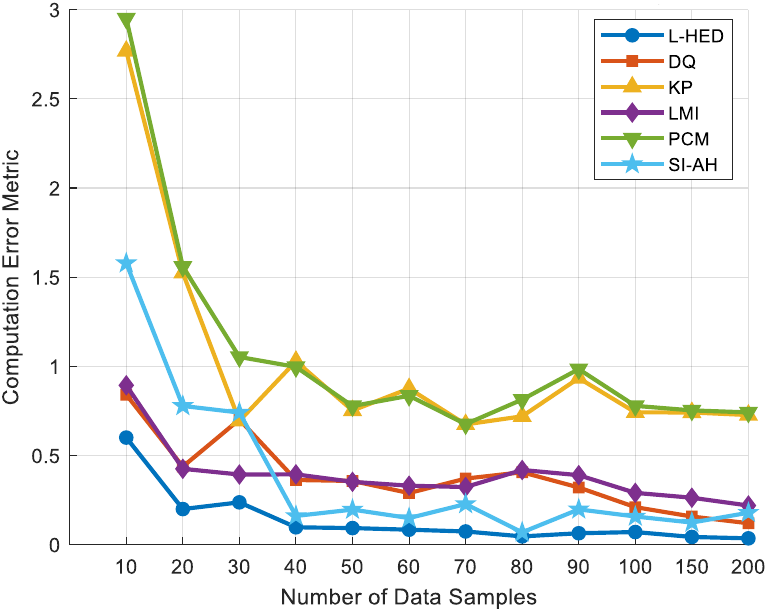}
	\caption{\ \ \leftskip=0pt \rightskip=0pt plus 0cm Impact of the number of data sets on the accuracy of $\boldsymbol{Y}$.}
	\label{s6-2}
\end{figure}

The number of source data sets used for estimating the HECPs is a crucial factor that directly affects the estimation accuracy. In this section, the impact of the number of source data sets on the estimation accuracy of HECPs is investigated. The number of source data sets is varied from 10 to 200, and the average estimation error is calculated for each case. The results are shown in Figures \ref{s6-1} and \ref{s6-2}.

The results from Figures \ref{s6-1} and \ref{s6-2} show that when the number of data sets used to estimate $\boldsymbol{X}$ and $\boldsymbol{X}$ is relatively small, increasing the number leads to improved HECP estimation accuracy across all seven methods. However, once the number of synthesized data sets reaches 100, the accuracy tends to stabilize. Therefore, to balance computational efficiency and precision, selecting 100 sets of \{\({\boldsymbol{A_{i}}}\)\} and \{\({\boldsymbol{B_{i}}}\)\} for each computation is optimal. In the synthesized data used in this paper, each set contains 100 source data pairs \{\({\boldsymbol{A_{i}}}\)\} and \{\({\boldsymbol{B_{i}}}\)\}.

\subsubsection{5.1.8. Effectiveness of uncertainty metric based data selection}

For data selection in the HEC problem, traditional methods in \cite{w33} are primarily based on the $\boldsymbol{AX = XB}$ formulation and typically use Axis-Angle representations for filtering. However, there is a lack of a standardized approach for data selection under the $\boldsymbol{AX = YB}$ formulation. In this section, the effectiveness of the proposed uncertainty metric-based data selection method under the $\boldsymbol{AX = YB}$ formulation is validated. Since the proposed method exhibits low dependency on data selection, the effectiveness of the uncertainty metric-based data selection is validated using the Dual -Quaternion(DQ), Kronecker-Product(KP), LMI, and Point-Cloud Matching (PCM) algorithms for solving HECPs. Each method is provided with the same 100 data sets for HECPs estimation. The data selection criteria are defined in Table \ref{tab:s7}.

\begin{figure}[!htbp]\centering
	\includegraphics[width=0.48\textwidth]{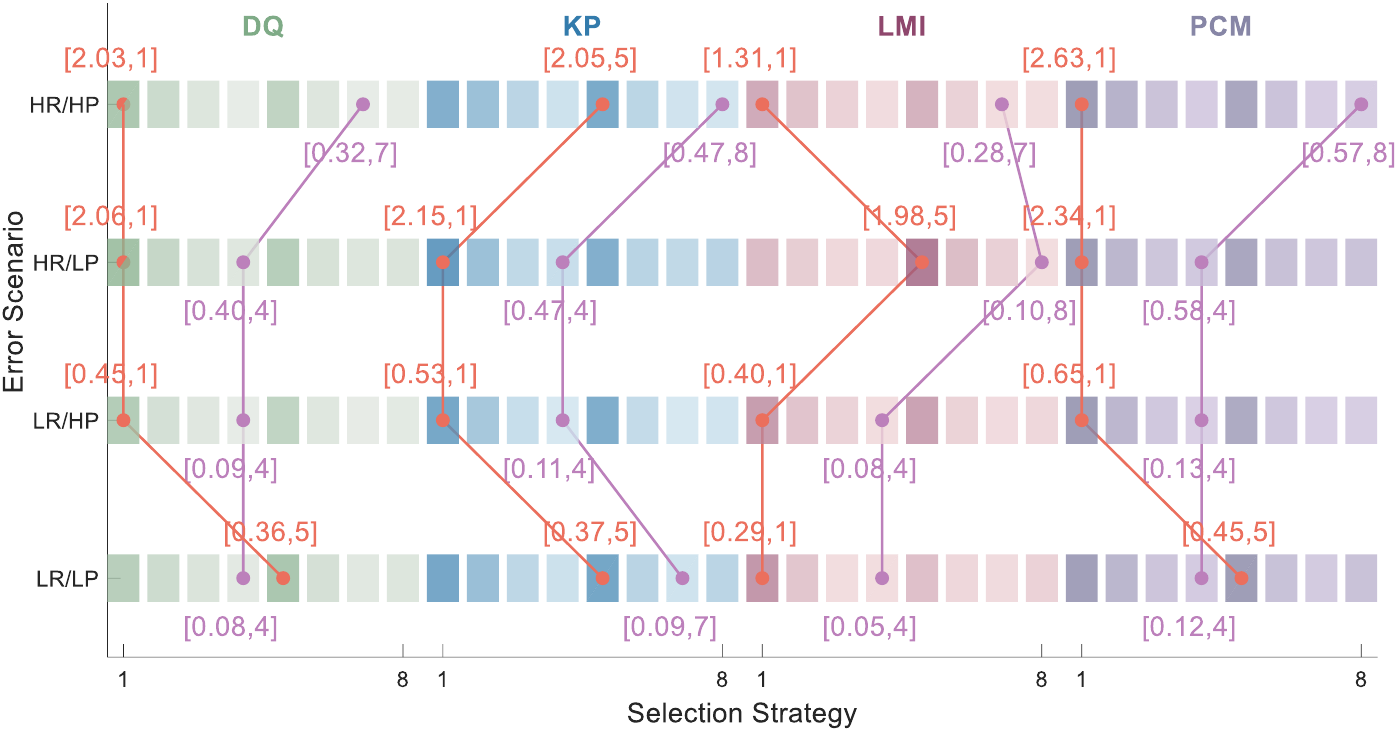}
	\caption{\ \ \leftskip=0pt \rightskip=0pt plus 0cm Impact of data selection on four methods.}
	\label{s7}
\end{figure}

Where Max\_Uncertain\_Metric(1:10) indicates that the first 10 data sets with the highest uncertainty metric values are selected, and so on. The results of the data selection are shown in Figure \ref{s7}.

Figure \ref{s7} presents the HECP estimation errors of the four methods under different error conditions after applying eight data selection strategies. The purple line represents the data selection strategy that yields the highest estimation accuracy for each method, while the red line represents the strategy with the lowest estimation accuracy. It can be observed that different methods exhibit varying performance under different data selection strategies. However, under high uncertainty conditions, the estimation accuracy obtained using source data with lower uncertainty metrics is higher than that achieved without any data selection. Therefore, when using these four methods to solve the $\boldsymbol{AX = YB}$ formulation of the HEC problem, the uncertainty metric proposed in this paper can be considered as an effective criterion for selecting source data \{\({\boldsymbol{A_{i}}}\)\} and \{\({\boldsymbol{B_{i}}}\)\}.

\subsubsection{5.1.9. Discussion on the optimal residual form for reflecting HECPs estimation error}

Unlike in simulation, in the real world, obtaining the ideal $\boldsymbol{X}_{opt}$ and $\boldsymbol{Y}_{opt}$ is impossible due to the uncertainty in the source data. Therefore, to evaluate the estimation error of the HECPs in practical applications, one must rely solely on the constructed residuals. The fundamental logic of residual construction is to use the estimated $X$ and $Y$ to form expressions such as $\left\|\boldsymbol{AX}-\boldsymbol{YB}\right\|_{F}$. 

\begin{figure}[!htbp]\centering
	\includegraphics[width=0.4\textwidth]{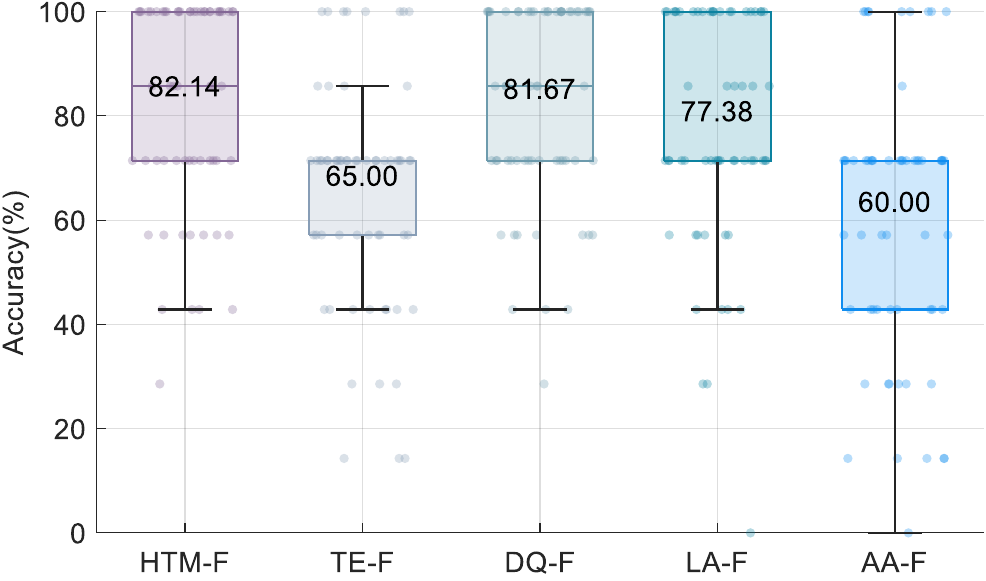}
	\caption{\ \ \leftskip=0pt \rightskip=0pt plus 0cm Accuracy comparison of five residual forms.}
	\label{s8}
\end{figure}

However, due to the lack of a bi-invariant measure on the Euclidean group and the presence of uncertainty in \{\({\boldsymbol{A_{i}}}\)\} and \{\({\boldsymbol{B_{i}}}\)\}, the relative magnitude of the residuals does not accurately reflect the relative precision of the estimated HECPs. To identify the residual construction form that best reflects the relative accuracy of the estimated HECPs, 60 pairs of synthesized data were generated. The seven methods were used to compute the HECPs, and the resulting $\boldsymbol{X}$ and $\boldsymbol{Y}$ were compared to their respective ground truth values $\boldsymbol{X}_{opt}$ and $\boldsymbol{Y}_{opt}$  to obtain the true estimation errors, which were then ranked. Subsequently, residuals were constructed using five different representations: HTM, Position and Euler Angles, Dual Quaternions, Lie Algebra, and Axis-Angle. The corresponding residual-based rankings were compared to the ground-truth-based rankings, and the accuracy of each residual form in reflecting relative error magnitude is shown in Figure \ref{s8}.

\begin{figure}[!htbp]\centering
	\includegraphics[width=0.45\textwidth]{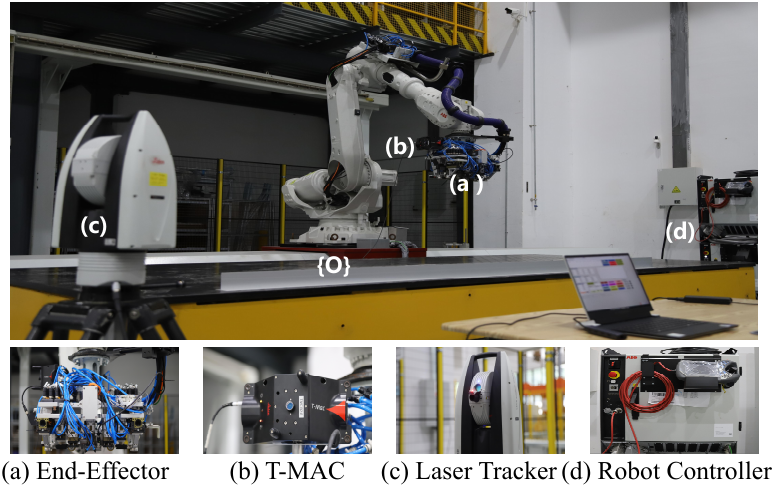}
	\caption{\ \ \leftskip=0pt \rightskip=0pt plus 0cm Real-Word experiments are carried out with an ABB IRB 6700 155/2.85 robot equipped with Lecia AT 960 Laser Tracker.}
	\label{s9-1-2}
\end{figure}

\begin{figure*}[!htbp]\centering
	\includegraphics[width=0.9\textwidth]{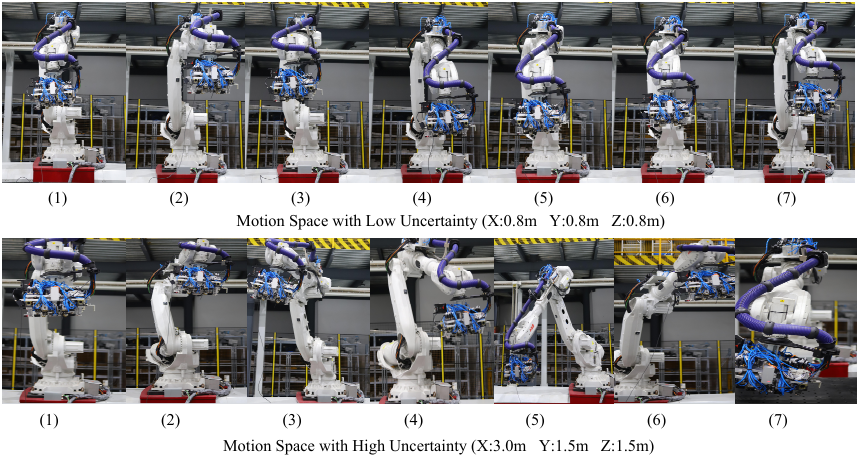}
	\caption{\ \ \leftskip=0pt \rightskip=0pt plus 0cm Two uncertainty workspace configurations.}
	\label{s9-2-1}
\end{figure*}

According to the results in  Figure \ref{s8}., the residual constructed using the HTM-based form achieves the highest accuracy. Therefore, in the subsequent real-world experiments, the residual used to evaluate the estimation accuracy of HECPs is constructed based on HTM, which is shown as 
\begin{equation}
	\label{eq:108}
	 r = \frac{1}{N} \sum_{i=1}^{N}\left\|\boldsymbol{AX}-\boldsymbol{YB}\right\|_{F} .
\end{equation}

\subsection{5.2. Real-Word Experiments}\label{sec:4.2}

\subsubsection{5.2.1. Introduction to real world experimental setup}

In addition to simulated experiments, real-word experimental calibrations were performed on a IR experiment platform, as depicted in  Figure \ref{s9-1-2}. The experimental setup consists of a ABB IRB 6700 155/2.85 robot and a Lecia AT 960 Laser Tracker, which is used to measure the position of the robot's pose. The robot carries an end-effector with a gravity-compensated calculated weight of 47 kg. It is also equipped with a T-Mac, which, in conjunction with a laser tracker, provides 6-DoF measurement data. To validate different uncertainties, source data were collected under two robot workspace configurations, which are defined in Figure \ref{s9-2-1}.
\begin{figure}[!htbp]\centering
	\includegraphics[width=0.35\textwidth]{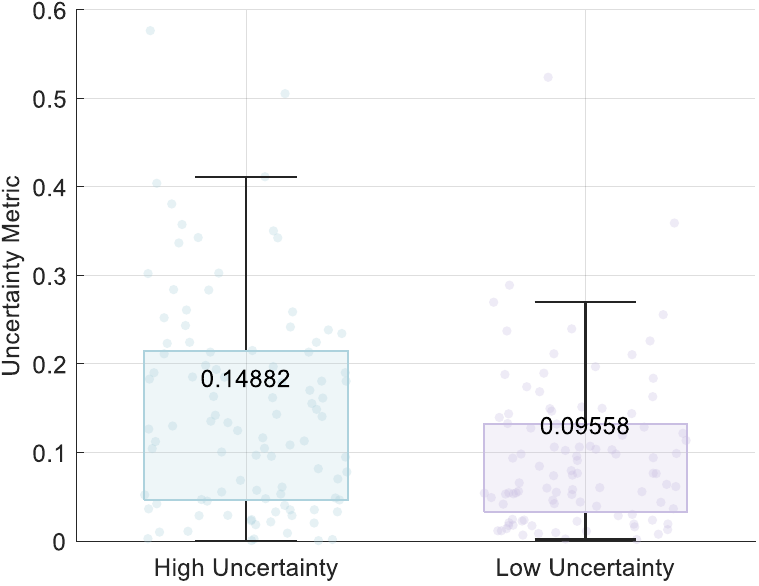}
	\caption{\ \ \leftskip=0pt \rightskip=0pt plus 0cm Uncertainty metric for two workspace configurations.}
	\label{s9-3}
\end{figure}

 \begin{figure}[!htbp]\centering
	\includegraphics[width=0.45\textwidth]{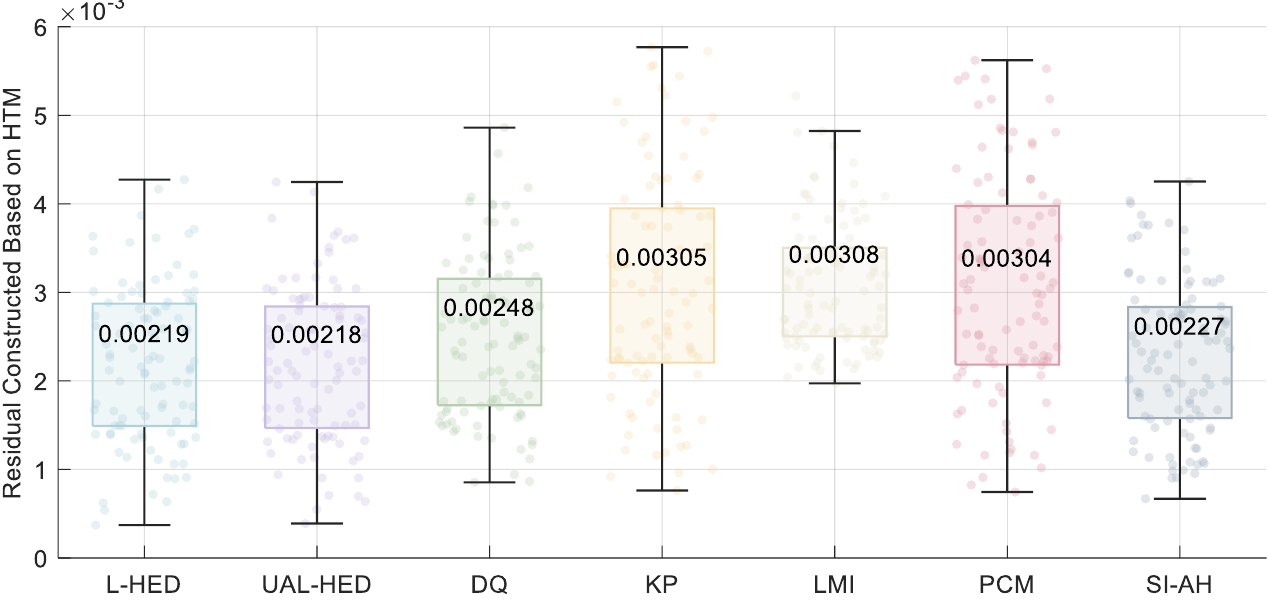}
	\caption{\ \ \leftskip=0pt \rightskip=0pt plus 0cm Performance of the seven methods in the large uncertainty workspace.}
	\label{s9-4}
\end{figure}

 \begin{figure}[!htbp]\centering
	\includegraphics[width=0.45\textwidth]{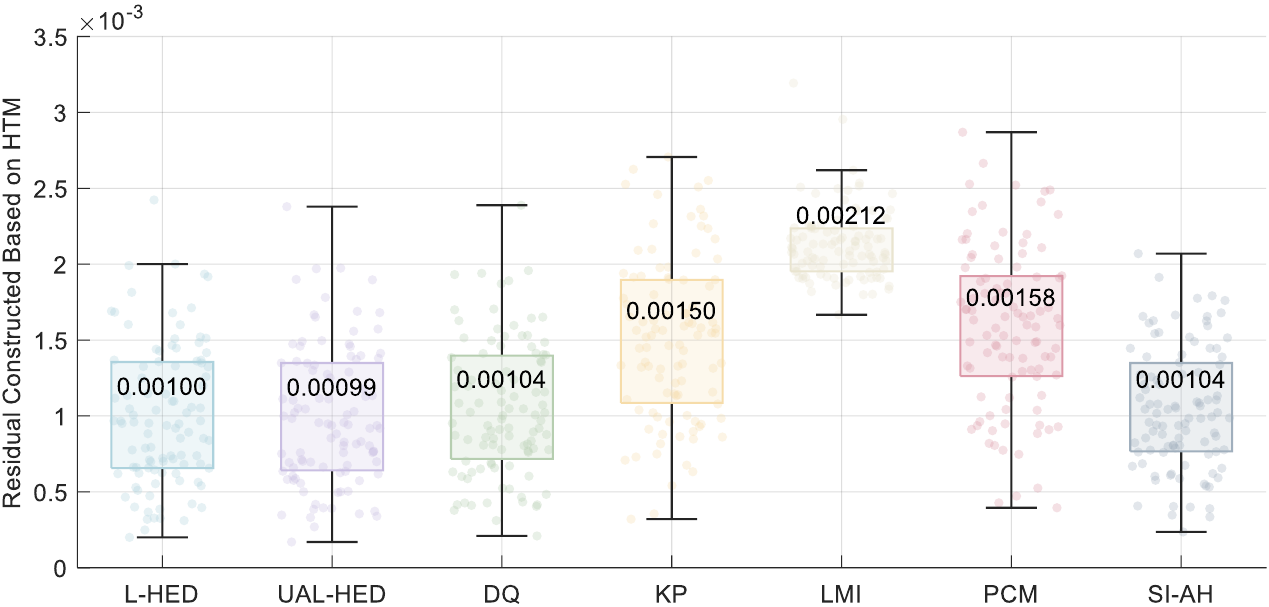}
	\caption{\ \ \leftskip=0pt \rightskip=0pt plus 0cm Performance of the seven methods in the small uncertainty workspace.}
	\label{s9-5}
\end{figure}

\subsubsection{5.2.2 Real world experimental validation of the uncertainty metric}

For the two workspace configurations, 100 sets of source data were collected for each. Based on the uncertainty metric, the corresponding uncertainty levels for each workspace are shown in Figure \ref{s9-3}. Figure \ref{s9-3} shows that, compared to the uncertainty level in the small-motion workspace, the large workspace exhibits greater dispersion in uncertainty, with the maximum uncertainty reaching 0.6. The average uncertainty in the large workspace is 1.56 times that of the small workspace. It can also be concluded that as the robot's workspace increases, the uncertainty of the source data increases as well, which is consistent with the analysis presented in Section 1.

\begin{figure*}[h]\centering
	\includegraphics[width=1\textwidth]{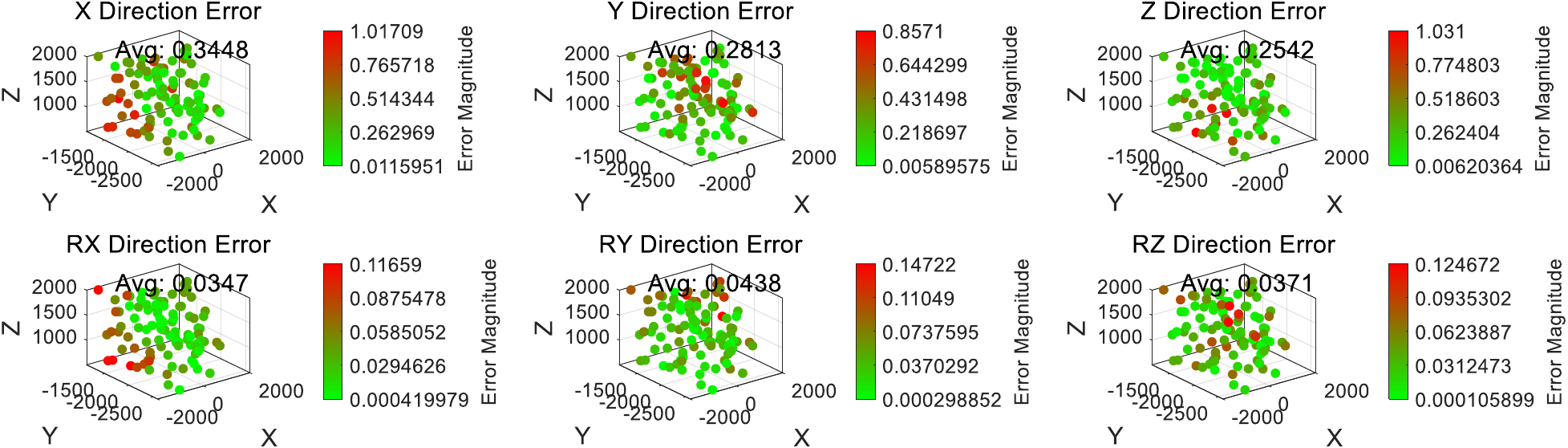}
	\caption{\ \ \leftskip=0pt \rightskip=0pt plus 0cm Six dimensional difference obtained under large uncertainty space.}
	\label{s9-6}
\end{figure*}

\begin{figure*}[!htbp]\centering
	\includegraphics[width=1\textwidth]{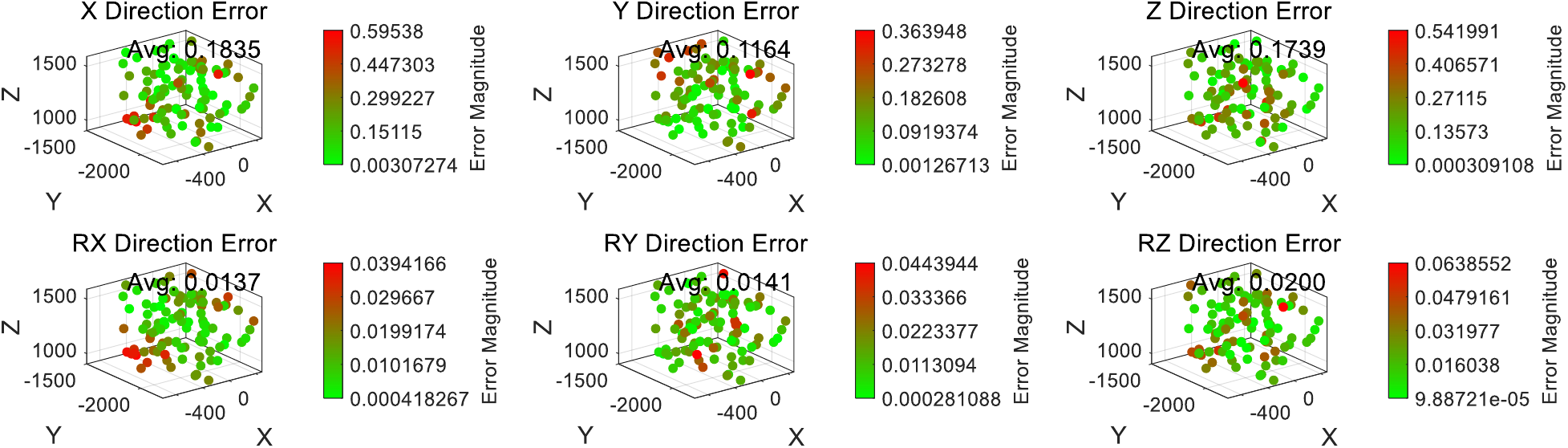}
	\caption{\ \ \leftskip=0pt \rightskip=0pt plus 0cm Six dimensional difference obtained under small uncertainty space.}
	\label{s9-7}
\end{figure*}

\subsubsection{5.2.3. Accuracy comparison of seven methods in real world experiments}

In both uncertainty workspace configurations,  100 source data pairs \{\({\boldsymbol{A_{i}}}\)\} and \{\({\boldsymbol{B_{i}}}\)\} were collected for estimating the HECPs, followed by 99 additional data pairs for validation. Figure \ref{s9-4} presents the performance of the seven methods under high-uncertainty workspace conditions using the HTMs-based residual construction base on Equation (\ref{eq:108}), while Figure \ref{s9-5} shows their performance under low-uncertainty workspace conditions using the same residual construction approach based on Equation (\ref{eq:108}).

It can be observed that in real-world experiments, the proposed UAL-HED method consistently achieves the best performance based on the constructed evaluation metric, regardless of whether the workspace is characterized by high or low uncertainty. The other two methods proposed in this paper also perform well in real-world scenarios, ranking just behind UAL-HED in terms of estimation accuracy.

By comparing the residual values across all seven methods, it is evident that the HECPs estimated in the high-uncertainty workspace consistently yield higher residual metrics than those obtained in the low-uncertainty workspace. This indicates that the uncertainty of the source data significantly affects the accuracy of HECPs estimation.

In practical industrial scenarios, the most commonly encountered use case is applying the obtained HECPs to transform measurement data from the camera's coordinate frame to the robot's coordinate frame. Using the 100 collected source data pairs $\{\boldsymbol{A_{i}}\}$ and $\{\boldsymbol{B_{i}}\}$, the HECPs $\boldsymbol{X}$ and $\boldsymbol{Y}$ are computed via UAL-HED. The 6-DoF measurement data from the camera are then transformed into the robot coordinate frame and compared with the corresponding data obtained from the teach pendant. The resulting differences for the two uncertainty workspace configurations are shown in Figures \ref{s9-6} and \ref{s9-7}.
\begin{figure*}[!htbp]\centering
	\includegraphics[width=1\textwidth]{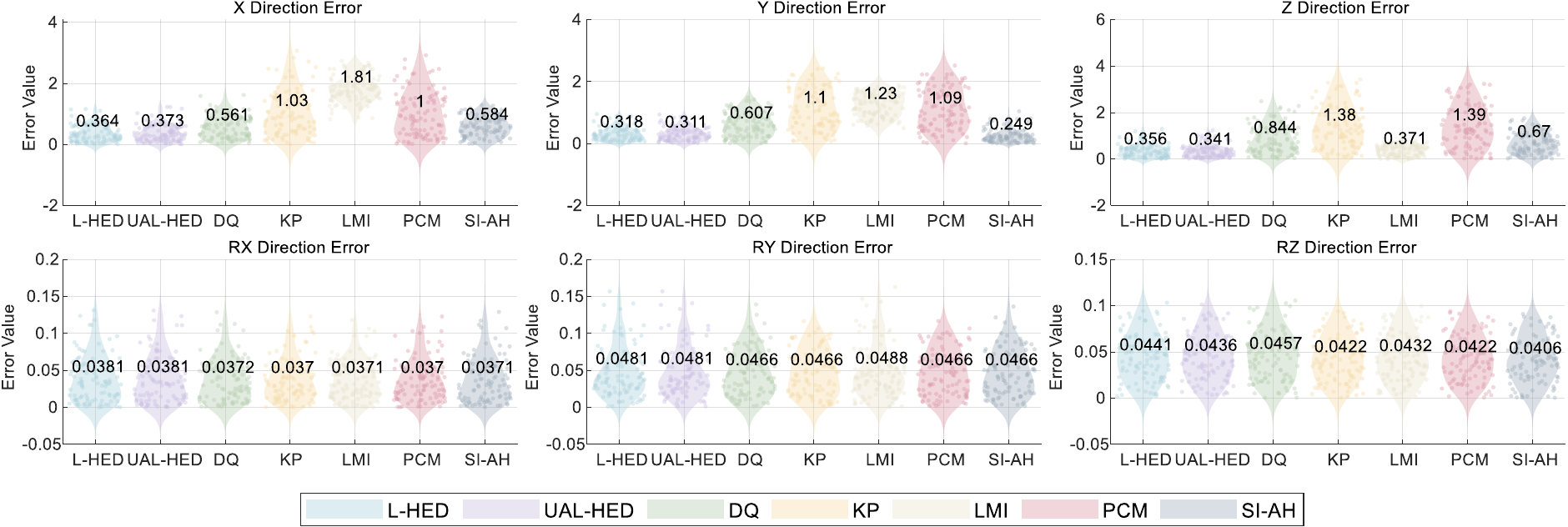}
	\caption{\ \ \leftskip=0pt \rightskip=0pt plus 0cm 6-DoF differences obtained by the seven methods in the large uncertainty workspace.}
	\label{s9-8}
\end{figure*}

\begin{figure*}[!htbp]\centering
	\includegraphics[width=1\textwidth]{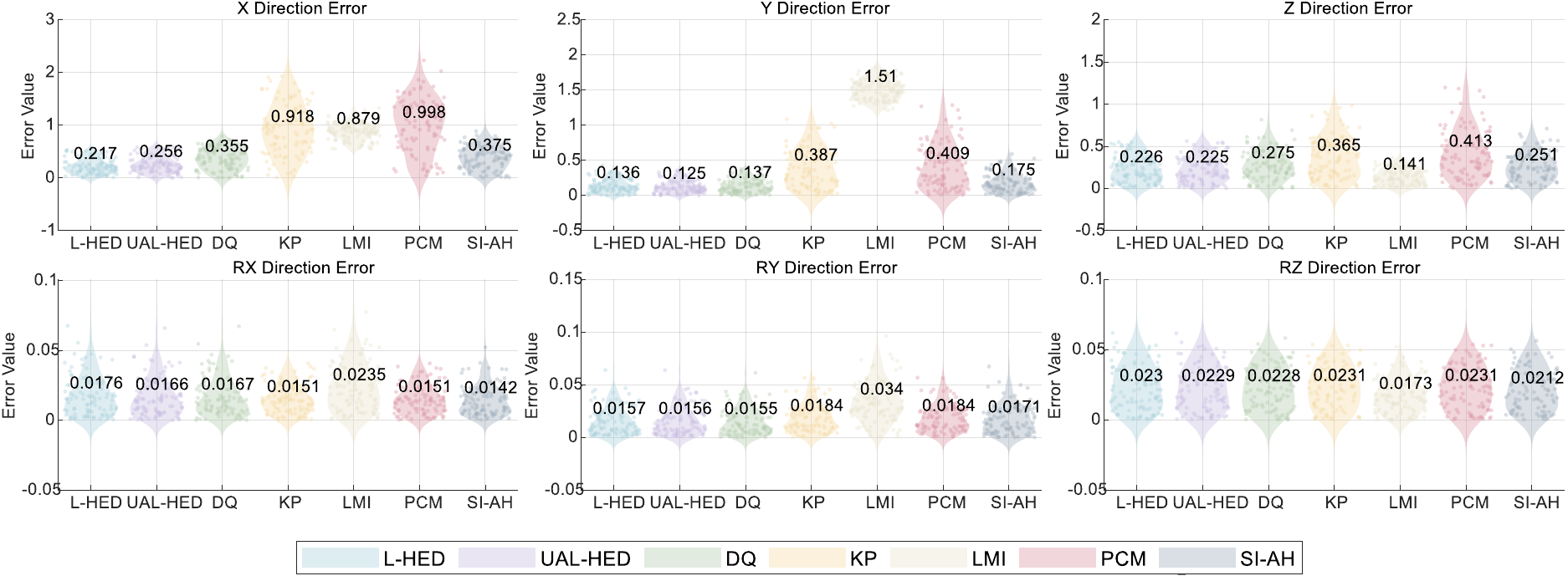}
	\caption{\ \ \leftskip=0pt \rightskip=0pt plus 0cm 6-DoF differences obtained by the seven methods in the small uncertainty workspace.}
	\label{s9-9}
\end{figure*}

The results in Figures \ref{s9-6} and \ref{s9-7} show that in all six directions, both the position differences and Euler angle differences obtained in the high-uncertainty workspace are greater than those obtained in the low-uncertainty workspace. In the high-uncertainty workspace, the average position difference is 0.293 mm and the average Euler angle difference is \(0.039^\circ\), while in the low-uncertainty workspace, the average position difference is 0.158 mm and the average Euler angle difference is \(0.016^\circ\). This further confirms that the uncertainty of source data has a significant impact on the accuracy of HECPs estimation.

Finally, the 99 additional data pairs unused during HECPs estimation were used to validate the six-dimensional differences across the seven methods. The results are shown in Figures \ref{s9-8} and \ref{s9-9}. 
 
\begin{table}[!htbp]
\centering
\caption{\ \ \leftskip=0pt \rightskip=0pt plus 0cm Conversion errors in seven methods for solving high uncertainty spaces}
\label{tab:s8}
\footnotesize  
\begin{tabular}{m{3.7cm}<{\centering} m{3.7cm}<{\centering}}
\toprule
\textrm{\color{black}{Method}} & \color{black}{Conversion error} \\
\midrule
L-HED & 0.346mm \\
UAL-HED & 0.342mm \\
Dual-Quaternion & 0.671mm\\
Kronecker-Product & 1.17mm\\
LMI & 1.137mm\\
Point-Cloud Matching & 1.16mm \\
SI-AH & 0.501mm\\
\bottomrule
\end{tabular}
\end{table}

For high-uncertainty workspace, the 7 methods show significant differences in displacement direction, with average displacement differences of conversion errors are shown in Table \ref{tab:s8}. For small-uncertainty workspace, the 7 methods show smaller differences in displacement direction, with average displacement differences of conversion errors are shown in Table \ref{tab:s9}. 
\begin{table}[!htbp]
\centering
\caption{\ \ \leftskip=0pt \rightskip=0pt plus 0cm Conversion errors in seven methods for solving small uncertainty spaces}
\label{tab:s9}
\footnotesize
\begin{tabular}{m{3.7cm}<{\centering} m{3.7cm}<{\centering}}
\toprule
\textrm{\color{black}{Method}} & \color{black}{Conversion error} \\
\midrule
L-HED & 0.193mm \\
UAL-HED & 0.202mm\\
Dual-Quaternion & 0.256mm\\
Kronecker-Product & 0.557mm\\
LMI & 0.843mm\\
Point-Cloud Matching & 0.607mm \\
SI-AH & 0.267mm\\
\bottomrule
\end{tabular}
\end{table}

\section{6. Conclusions}
\label{sec:5}
In HEC problems, the uncertainty of the source data has a significant impact on calibration accuracy particularly when there is a large disparity in uncertainty between $\{\boldsymbol{A_{i}}\}$ and $\{\boldsymbol{B_{i}}\}$. Given the inherent difficulty of modeling uncertainty and the fact that, in practice, it is typically encapsulated within the source data $\{\boldsymbol{A_{i}}\}$ and $\{\boldsymbol{B_{i}}\}$, a novel metric \textit{SRM@SE(3)} based on the properties of the Euclidean group $SE(3)$ was proposed. This metric relies solely on the computational results of the source data and can effectively reflect the relative uncertainty between them. Building on this, the UAL-HED method, an enhanced method that improves robustness in solving the $\boldsymbol{AX=YB}$ model by correcting the iterative process. UAL-HED is an optimized version of the L-HED method. In addition, a method called SI-AH is designed to generate high-quality initial solutions, improving the overall stability and convergence speed of the algorithm. The proposed metric and calibration algorithms have been thoroughly validated through both simulation and real-world experiments. The solutions also systematically explore various factors that influence estimation accuracy. The experimental results clearly demonstrate the effectiveness of the proposed methods under both high and low uncertainty data conditions. As such, the calibration framework introduced in this work exhibits strong theoretical and practical generality, making it well-suited for solving HEC problems across a wide range of scenarios.

% if have a single appendix:
%\appendix[Proof of the Zonklar Equations]
% or
%\appendix  % for no appendix heading
% do not use \section anymore after \appendix, only \section*
% is possibly needed

% use appendices with more than one appendix
% then use \section to start each appendix
% you must declare a \section before using any
% \subsection or using \label (\appendices by itself
% starts a section numbered zero.)
%

%\appendices
%\section{Proof of the First Zonklar Equation}
%Appendix one text goes here.

% you can choose not to have a title for an appendix
% if you want by leaving the argument blank
%\section{}
%Appendix two text goes here.

% use section* for acknowledgment
%and the authors would also like to thank the anonymous reviewers and %the editors for their valuable remarks on this paper.

% Can use something like this to put references on a page
% by themselves when using endfloat and the captionsoff option.

% \section{Acknowledgements}
% We would like to acknowledge the developers of INRIA for providing the very useful ViSP library.

\section{Declaration of conflicting interests}
The authors declared no potential conflicts of interest with respect to the research, authorship, and/or publication of this article.

\section{Funding}
This work was supported by the National Natural Science Foundation of China under Grant Nos. U24A20130, 52188102, 52205521 and 52090054.

% trigger a \newpage just before the given reference
% number - used to balance the columns on the last page
% adjust value as needed - may need to be readjusted if
% the document is modified later
%\IEEEtriggeratref{8}
% The "triggered" command can be changed if desired:
%\IEEEtriggercmd{\enlargethispage{-5in}}

% This command serves to balance the column lengths
% on the last page of the document manually. It shortens
% the textheight of the last page by a suitable amount.
% This command does not take effect until the next page
% so it should come on the page before the last. Make
% sure that you do not shorten the textheight too much.

%%%%%%%%%%%%%%%%%%%%%%%%%%%%%%%%%%%%%%%%%%%%%%%%%%%%%%%%%%%%%%%%%%%%%%%%%%%%%%%%

%\twocolumn

%\onecolumn

% references section
% can use a bibliography generated by BibTeX as a .bbl file
% BibTeX documentation can be easily obtained at:
% http://mirror.ctan.org/biblio/bibtex/contrib/doc/
% The IEEEtran BibTeX style support page is at:
% http://www.michaelshell.org/tex/ieeetran/bibtex/

\bibliographystyle{SageH}
\bibliography{one}

@article{w1,
  title={A survey of welding robot intelligent path optimization},
  author={Wang, Xuewu and Zhou, Xin and Xia, Zelong and Gu, Xingsheng},
  journal={Journal of Manufacturing Processes},
  volume={63},
  pages={14--23},
  year={2021},
  publisher={Elsevier}
}

@article{w2,
  title={Robot control system for grinding of large hydro power turbines},
  author={Thomessen, Trygve and Lien, Terje K and Sann{\ae}s, Per K},
  journal={Industrial Robot: An International Journal},
  volume={28},
  number={4},
  pages={328--334},
  year={2001},
  publisher={MCB UP Ltd}
}

@article{w3,
  title={Generating optimized trajectories for robotic spray painting},
  author={Gleeson, Daniel and Jakobsson, Stefan and Salman, Raad and Ekstedt, Fredrik and Sandgren, Niklas and Edelvik, Fredrik and Carlson, Johan S and Lennartson, Bengt},
  journal={IEEE Transactions on Automation Science and Engineering},
  volume={19},
  number={3},
  pages={1380--1391},
  year={2022},
  publisher={IEEE}
}

@article{w4,
  title={Accurate modeling of material removal depth in convolutional process grinding for complex surfaces},
  author={Zhou, Haoyuan and Zhao, Huan and Li, Xiangfei and Xu, Zairan and Ding, Han},
  journal={International Journal of Mechanical Sciences},
  volume={267},
  pages={109005},
  year={2024},
  publisher={Elsevier}
}

@article{w5,
  title={High-precision assembly of electronic devices with lightweight robots through sensor-guided insertion},
  author={Metzner, Maximilian and Leurer, Sebastian and Handwerker, Andreas and Karlidag, Engin and Blank, Andreas and Hefner, Florian and Franke, J{\"o}rg},
  journal={Procedia CIRP},
  volume={97},
  pages={337--341},
  year={2021},
  publisher={Elsevier}
}

@inproceedings{w6,
  title={Design of Dual Robot Collaborative Assembly Scheme for Aerospace Products},
  author={Zhang, Lijian and Pan, Dingwen and Hu, Ruiqin},
  booktitle={International Conference On Signal And Information Processing, Networking And Computers},
  pages={850--857},
  year={2022},
  organization={Springer}
}

@article{w7,
  title={A novel method to enhance the accuracy of parameter identification in elasto-geometrical calibration for industrial robots},
  author={Yu, Shihang and Nan, Jie and Sun, Yuwen},
  journal={Robotics and Computer-Integrated Manufacturing},
  volume={90},
  pages={102809},
  year={2024},
  publisher={Elsevier}
}

@article{w8,
  title={A comparative review of hand-eye calibration techniques for vision guided robots},
  author={Enebuse, Ikenna and Foo, Mathias and Ibrahim, Babul Salam Ksm Kader and Ahmed, Hafiz and Supmak, Fhon and Eyobu, Odongo Steven},
  journal={IEEE Access},
  volume={9},
  pages={113143--113155},
  year={2021},
  publisher={IEEE}
}

@article{w9,
  title={Robot--camera calibration in tightly constrained environment using interactive perception},
  author={Zhong, Fangxun and Li, Bin and Chen, Wei and Liu, Yun-Hui},
  journal={IEEE Transactions on Robotics},
  volume={39},
  number={6},
  pages={4952--4970},
  year={2023},
  publisher={IEEE}
}

@article{w10,
  title={Real-time trajectory position error compensation technology of industrial robot},
  author={Li, Rui and Ding, Ning and Zhao, Yang and Liu, He},
  journal={Measurement},
  volume={208},
  pages={112418},
  year={2023},
  publisher={Elsevier}
}

@inproceedings{w11,
  title={An information-theoretic approach to the correspondence-free AX= XB sensor calibration problem},
  author={Ackerman, Martin Kendal and Cheng, Alexis and Chirikjian, Gregory},
  booktitle={2014 IEEE International Conference on Robotics and Automation (ICRA)},
  pages={4893--4899},
  year={2014},
  organization={IEEE}
}

@article{w12,
  title={An uncertainty analysis method for error reduction in end-effector of spatial robots with joint clearances and link dimension deviations},
  author={Hafezipour, M and Khodaygan, Saeed},
  journal={International Journal of Computer Integrated Manufacturing},
  volume={30},
  number={6},
  pages={653--663},
  year={2017},
  publisher={Taylor \& Francis}
}

@article{w13,
  title={Accuracy of videometry with CCD sensors},
  author={Lenz, Reimar and Fritsch, Dieter},
  journal={ISPRS Journal of Photogrammetry and Remote Sensing},
  volume={45},
  number={2},
  pages={90--110},
  year={1990},
  publisher={Elsevier}
}

@article{w14,
  title={Probabilistic Framework for Hand--Eye and Robot--World Calibration $ AX= YB$},
  author={Ha, Junhyoung},
  journal={IEEE Transactions on Robotics},
  volume={39},
  number={2},
  pages={1196--1211},
  year={2022},
  publisher={IEEE}
}

@article{w15,
  title={Multiscale and Uncertainty-Aware Targetless Hand-Eye Calibration Via the Gauss--Helmert Model},
  author = {{\v C}olakovi{\'c}-Benceri{\'c}, Marta and Per{\v s}i{\'c}, Juraj and Markovi{\'c}, Ivan and Petrovi{\'c}, Ivan},
  journal = {IEEE Transactions on Robotics},
  volume = {41},
  pages = {2340--2357},
  year = {2025},
  publisher = {IEEE}
}

@article{w16,
  title={Uncertainty-aware hand--eye calibration},
  author={Ulrich, Markus and Hillemann, Markus},
  journal={IEEE Transactions on Robotics},
  volume={40},
  pages={573--591},
  year={2023},
  publisher={IEEE}
}

@inproceedings{w17,
  title={Assessing the accuracy of industrial robots through metrology for the enhancement of automated non-destructive testing},
  author={Morozov, Maxim and Riise, Jonathan and Summan, Rahul and Pierce, Stephen Gareth and Mineo, Carmelo and MacLeod, Charles N and Brown, Roy Hutton},
  booktitle={2016 IEEE International Conference on Multisensor Fusion and Integration for Intelligent Systems (MFI)},
  pages={335--340},
  year={2016},
  organization={IEEE}
}

@article{w18,
  title={A local POE-based self-calibration method using position and distance constraints for collaborative robots},
  author={He, Jianhui and Gu, Lefeng and Yang, Guilin and Feng, Yiyang and Chen, Silu and Fang, Zaojun},
  journal={Robotics and Computer-Integrated Manufacturing},
  volume={86},
  pages={102685},
  year={2024},
  publisher={Elsevier}
}

@article{w19,
  title = {Optimization Method for Configuration Set for Field Calibration of Industrial Robot},
  author = {Wang, Ziyi and Qin, Lan and Liu, Jingcheng and Li, Min and Liu, Jun},
  journal = {IEEE Transactions on Industrial Electronics},
  volume = {72},
  number = {6},
  pages = {6103--6113},
  year = {2025},
  publisher={IEEE}
}

@article{w20,
  title={Efficient kinematic calibration for articulated robot based on unit dual quaternion},
  author={Luo, Jingbo and Chen, Silu and Zhang, Chi and Chen, Chin-Yin and Yang, Guilin},
  journal={IEEE Transactions on Industrial Informatics},
  volume={19},
  number={12},
  pages={11898--11909},
  year={2023},
  publisher={IEEE}
}

@article{w21,
  title={Kinematic Calibration for Serial Robots Based on a Vector Inner Product Error Model},
  author = {Liu, Fei and Gao, Guanbin and Na, Jing and Zhang, Faxiang},
  journal = {IEEE Transactions on Industrial Electronics},
  volume = {72},
  number = {3},
  pages = {2832--2841},
  year = {2025},
  publisher={IEEE}
}

@article{w22,
  title={Calibration of wrist-mounted robotic sensors by solving homogeneous transform equations of the form AX= XB},
  author={Shiu, YC and Ahmad, S},
  journal={IEEE Transactions on Robotics and Automation},
  volume={5},
  number={1},
  pages={16--29},
  year={1989},
  publisher={IEEE}
}

@article{w23,
  title={Simultaneous robot/world and tool/flange calibration by solving homogeneous transformation equations of the form AX= YB},
  author={Zhuang, Hanqi and Roth, Zvi S and Sudhakar, Raghavan},
  journal={IEEE Transactions on Robotics and Automation},
  volume={10},
  number={4},
  pages={549--554},
  year={1994},
  publisher={IEEE}
}

@article{w24,
  title={Finding the position and orientation of a sensor on a robot manipulator using quaternions},
  author={Chou, Jack CK and Kamel, M},
  journal={The International Journal of Robotics Research},
  volume={10},
  number={3},
  pages={240--254},
  year={1991},
  publisher={Sage Publications Sage CA: Thousand Oaks, CA}
}

@article{w25,
  title={Robot sensor calibration: solving AX= XB on the Euclidean group},
  author={Park, Frank C and Martin, Bryan J},
  journal={IEEE Transactions on Robotics and Automation},
  volume={10},
  number={5},
  pages={717--721},
  year={1994},
  publisher={IEEE}
}

@article{w26,
  title={A new formulation for hand--eye calibrations as point-set matching},
  author={Qiu, Shuwei and Wang, Miaomiao and Kermani, Mehrdad R},
  journal={IEEE Transactions on Instrumentation and Measurement},
  volume={69},
  number={9},
  pages={6490--6498},
  year={2020},
  publisher={IEEE}
}

@article{w27,
  title={Robot hand-eye calibration using structure-from-motion},
  author={Andreff, Nicolas and Horaud, Radu and Espiau, Bernard},
  journal={The International Journal of Robotics Research},
  volume={20},
  number={3},
  pages={228--248},
  year={2001},
  publisher={SAGE Publications}
}

@article{w28,
  title={Simultaneous robot-world and hand-eye calibration},
  author={Dornaika, Fadi and Horaud, Radu},
  journal={IEEE Transactions on Robotics and Automation},
  volume={14},
  number={4},
  pages={617--622},
  year={1998},
  publisher={IEEE}
}

@inproceedings{w29,
  title={A screw motion approach to uniqueness analysis of head-eye geometry},
  author={Chen, Homer H},
  booktitle={Proceedings. 1991 IEEE Computer Society Conference on Computer Vision and Pattern Recognition},
  pages={145--146},
  year={1991},
  organization={IEEE Computer Society}
}

@article{w30,
  title={Solving the robot-world hand-eye (s) calibration problem with iterative methods},
  author={Tabb, Amy and Ahmad Yousef, Khalil M},
  journal={Machine Vision and Applications},
  volume={28},
  number={5},
  pages={569--590},
  year={2017},
  publisher={Springer}
}

@article{w31,
  title={Simultaneous robot-world and hand-eye calibration using dual-quaternions and Kronecker product},
  author={Li, Aiguo and Wang, Lin and Wu, Defeng},
  journal={International Journal of the Physical Science},
  volume={5},
  number={10},
  pages={1530--1536},
  year={2010}
}

@article{w32,
  title={A new technique for fully autonomous and efficient 3 d robotics hand/eye calibration},
  author={Tsai, Roger Y and Lenz, Reimar K},
  journal={IEEE Transactions on Robotics and Automation},
  volume={5},
  number={3},
  pages={345--358},
  year={1989},
  publisher={IEEE}
}

@article{w33,
  title={Data selection for hand-eye calibration: A vector quantization approach},
  author={Schmidt, Jochen and Niemann, Heinrich},
  journal={The International Journal of Robotics Research},
  volume={27},
  number={9},
  pages={1027--1053},
  year={2008},
  publisher={SAGE Publications Sage UK: London, England}
}

@article{w34,
  title={Simultaneous hand-eye and robot-world calibration by solving the AX= YB problem without correspondence},
  author={Li, Haiyuan and Ma, Qianli and Wang, Tianmiao and Chirikjian, Gregory S},
  journal={IEEE Robotics and Automation Letters},
  volume={1},
  number={1},
  pages={145--152},
  year={2015},
  publisher={IEEE}
}

@article{w35,
  title={Chess-calibrating the hand-eye matrix with screw constraints and synchronization},
  author={Pachtrachai, Krittin and Vasconcelos, Francisco and Dwyer, George and Pawar, Vijay and Hailes, Stephen and Stoyanov, Danail},
  journal={IEEE Robotics and Automation Letters},
  volume={3},
  number={3},
  pages={2000--2007},
  year={2018},
  publisher={IEEE}
}

@inproceedings{w36,
  title={A novel robust approach for correspondence-free extrinsic calibration},
  author={Hu, Xiao and Olesen, Daniel and Per, Knudsen},
  booktitle={2019 IEEE/RSJ International Conference on Intelligent Robots and Systems (IROS)},
  pages={1--6},
  year={2019},
  organization={IEEE}
}

@article{w37,
  title={Correspondence matching and time delay estimation for hand-eye calibration},
  author={Wu, Jin and Liu, Ming and Zhang, Chengxi and Zhou, Zebo},
  journal={IEEE Transactions on Instrumentation and Measurement},
  volume={69},
  number={10},
  pages={8304--8313},
  year={2020},
  publisher={IEEE}
}

@article{w38,
  title={Non-orthogonal tool/flange and robot/world calibration},
  author={Ernst, Floris and Richter, Lars and Matth{\"a}us, Lars and Martens, Volker and Bruder, Ralf and Schlaefer, Alexander and Schweikard, Achim},
  journal={The International Journal of Medical Robotics and Computer Assisted Surgery},
  volume={8},
  number={4},
  pages={407--420},
  year={2012},
  publisher={Wiley Online Library}
}

@article{w39,
  title={Simultaneous robot-world and hand-eye calibration by the alternative linear programming},
  author={Zhao, Zijian},
  journal={Pattern Recognition Letters},
  volume={127},
  pages={174--180},
  year={2019},
  publisher={Elsevier}
}

@inproceedings{w40,
  title={Hand-eye calibration using convex optimization},
  author={Zhao, Zijian},
  booktitle={IEEE International Conference on Robotics and Automation},
  pages={2947--2952},
  year={2011},
  organization={IEEE}
}

@article{w41,
  title={Globally optimal hand-eye calibration using branch-and-bound},
  author={Heller, Jan and Havlena, Michal and Pajdla, Tomas},
  journal={IEEE Transactions on Pattern Analysis and Machine Intelligence},
  volume={38},
  number={5},
  pages={1027--1033},
  year={2015},
  publisher={IEEE}
}

@article{w42,
  title={A stochastic global optimization algorithm for the two-frame sensor calibration problem},
  author={Ha, Junhyoung and Kang, Donghoon and Park, Frank C},
  journal={IEEE Transactions on Industrial Electronics},
  volume={63},
  number={4},
  pages={2434--2446},
  year={2015},
  publisher={IEEE}
}

@article{w43,
  title={Globally optimal symbolic hand-eye calibration},
  author={Wu, Jin and Liu, Ming and Zhu, Yilong and Zou, Zuhao and Dai, Ming-Zhe and Zhang, Chengxi and Jiang, Yi and Li, Chong},
  journal={IEEE/ASME Transactions on Mechatronics},
  volume={26},
  number={3},
  pages={1369--1379},
  year={2020},
  publisher={IEEE}
}

@article{w44,
  title={A general approach to hand--eye calibration through the optimization of atomic transformations},
  author={Pedrosa, Eurico and Oliveira, Miguel and Lau, Nuno and Santos, V{\'\i}tor},
  journal={IEEE Transactions on Robotics},
  volume={37},
  number={5},
  pages={1619--1633},
  year={2021},
  publisher={IEEE}
}

@article{w45,
  title={Toward simultaneous coordinate calibrations of AX=YB problem by the LMI-SDP optimization},
  author={Pan, Jiabin and Fu, Zhongtao and Yue, Hengtao and Lei, Xiaoyu and Li, Miao and Chen, Xubing},
  journal={IEEE Transactions on Automation Science and Engineering},
  volume={20},
  number={4},
  pages={2445--2453},
  year={2022},
  publisher={IEEE}
}

@article{w46,
  title={Point Cloud Registration-Enabled Globally Optimal Hand--Eye Calibration},
  author={Zhu, Dahu and Wu, Hao and Ding, Tao and Hua, Lin},
  journal={IEEE/ASME Transactions on Mechatronics},
  year={2024},
  publisher={IEEE}
}

@article{w47,
  title={Two-Stage Hand-Eye Calibration based on Variance Minimization Principle},
  author={Ding, Tao and Tian, Dazhuang and Wu, Hao and Zhu, Dahu and Li, Wenlog},
  journal={IEEE Transactions on Instrumentation and Measurement},
  volume={74},
  number={3539010},
  year={2025},
  publisher={IEEE}
}

@inproceedings{w48,
  title={Hand-eye calibration algorithm based on an optimized neural network},
  author={Hua, Jiang and Zeng, Liangcai},
  booktitle={Actuators},
  volume={10},
  number={4},
  pages={85},
  year={2021},
  organization={MDPI}
}

@article{w49,
  title={Generative Adversarial Networks for Solving Hand-Eye Calibration Without Data Correspondence},
  author={Hong, Ilkwon and Ha, Junhyoung},
  journal={IEEE Robotics and Automation Letters},
  year={2025},
  publisher={IEEE}
}

@inproceedings{w50,
  title={Parameterizations for reducing camera reprojection error for robot-world hand-eye calibration},
  author={Tabb, Amy and Yousef, Khalil M Ahmad},
  booktitle={2015 IEEE/RSJ International Conference on Intelligent Robots and Systems (IROS)},
  pages={3030--3037},
  year={2015},
  organization={IEEE}
}

@article{w51,
  title={A noise-tolerant algorithm for robotic hand-eye calibration with or without sensor orientation measurement},
  author={Zuang, Hanqi and Shiu, Yiu Cheung},
  journal={IEEE Transactions on Systems, Man, and Cybernetics},
  volume={23},
  number={4},
  pages={1168--1175},
  year={1993},
  publisher={IEEE}
}

@article{w52,
  title={A two-step solution for robot-world calibration made intelligible by implementing Chasles' motion decomposition in Ad (SE (3))},
  author={Wang, Xiao and Liu, Chenglin and Sun, Haoxiang and Song, Hanwen},
  journal={Mechanism and Machine Theory},
  volume={191},
  pages={105522},
  year={2024},
  publisher={Elsevier}
}

@article{w53,
  title={Solving the robot-world/hand-eye calibration problem using the Kronecker product},
  author={Shah, Mili},
  journal={Journal of Mechanisms and Robotics},
  volume={5},
  number={3},
  pages={031007},
  year={2013},
  publisher={American Society of Mechanical Engineers}
}

\end{document}